\documentclass{article}

\usepackage{microtype}
\usepackage{graphicx}
\usepackage{booktabs} % for professional tables

\usepackage{tabularx}
\usepackage{xstring}
\usepackage{multirow}
\usepackage{xspace}
\usepackage{url}
\usepackage{subcaption}
\usepackage{xcolor}
\usepackage[hang,flushmargin]{footmisc}
\usepackage{url}
\usepackage{xcolor}
\usepackage{xspace} 
\usepackage{caption}

\usepackage{subcaption}
\usepackage{hyperref}

\usepackage[accepted]{icml2024}

\usepackage{amsmath}
\usepackage{amssymb}
\usepackage{mathtools}
\usepackage{amsthm}

\usepackage[capitalize,noabbrev]{cleveref}

\theoremstyle{plain}

\theoremstyle{definition}

\theoremstyle{remark}

\usepackage[textsize=tiny]{todonotes}

\definecolor{brown}{RGB}{160,82,45}

\icmltitlerunning{Coarse-To-Fine Tensor Trains for Compact Visual Representations}

\begin{document}

\twocolumn[
\icmltitle{Coarse-To-Fine Tensor Trains for Compact Visual Representations}

\begin{icmlauthorlist}
\icmlauthor{Sebastian Loeschcke}{KU,ITU}
\icmlauthor{Dan Wang}{KU}
\icmlauthor{Christian Leth-Espensen}{AU}
\\
\icmlauthor{Serge Belongie}{KU}
\icmlauthor{Michael J. Kastoryano}{KU}
\icmlauthor{Sagie Benaim}{HUJ}
\end{icmlauthorlist}

\icmlaffiliation{KU}{University of Copenhagen}
\icmlaffiliation{AU}{Aarhus University}
\icmlaffiliation{ITU}{IT University of Copenhagen}
\icmlaffiliation{HUJ}{Hebrew University of Jerusalem}

\icmlcorrespondingauthor{Sebastian Loeschcke}{sbl@di.ku.dk}

\icmlkeywords{Machine Learning, ICML}

\vskip 0.3in
]

\printAffiliationsAndNotice{}

\begin{abstract}
The ability to learn compact, high-quality, and easy-to-optimize representations for visual data is paramount to many applications such as novel view synthesis and 3D reconstruction. Recent work has shown substantial success in using tensor networks to design such compact and high-quality representations.
However, the ability to optimize tensor-based representations, and in particular, the highly compact tensor train representation, is still lacking. This has prevented practitioners from deploying the full potential of tensor networks for visual data. 
To this end, we propose `Prolongation Upsampling Tensor Train (PuTT)', a novel method for learning tensor train representations in a coarse-to-fine manner. Our method involves the prolonging or `upsampling' of a learned tensor train representation, creating a sequence of `coarse-to-fine' tensor trains that are incrementally refined. 
We evaluate our representation along three axes: (1). compression, (2). denoising capability, and (3). image completion capability. To assess these axes, we consider the tasks of image fitting, 3D fitting, and novel view synthesis, where our method shows an improved performance compared to state-of-the-art tensor-based methods.
\end{abstract}

\includegraphics[width=1.25em,height=1.15em]{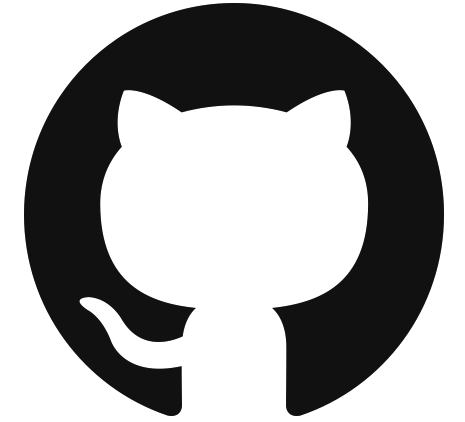}\hspace{.75em}
\parbox{\dimexpr\linewidth-7\fboxsep-7\fboxrule}{\vspace{-.5em}\url{https://github.com/sebulo/PuTT}}

\section{Introduction}
\label{sec:intro}

Building a compact, easy-to-fit, and high-quality visual data representation is paramount for many computer vision and graphics applications, such as novel view synthesis, 3D fitting, and generation~\cite{xie2022neural}.
To this end, recent research within neural field representations~\cite{tensoRF,ttnf,instantNerf}  has directed its focus towards the sparsity and efficiency of tensor networks, hinting at the potential gains in the compactness and quality of such representations. In particular, TensoRF~\cite{tensoRF} applied low-rank tensor decomposition of radiance fields to achieve substantial efficiency gains. This approach was further extended by TT-NF~\cite{ttnf}, which utilized a tensor train representation to improve compression. Indeed, tensor trains, as well as more complex tensor networks~\cite{PEPS,MERA}, are known to allow substantial
improvements in compression. 

However, despite encouraging progress, the full realization of tensor networks for compact and efficient representations is still lacking. The main obstacles of current tensor-based representations are: (1). Optimization using current gradient-based schemes often gets stuck in local minima and falls short of the ultimate compression limit of the tensor train format. 
In particular, TensoRF~\cite{tensoRF} can only handle small tensors and does not enjoy the significant parameter efficiency of tensor trains. 
TT-NF~\cite{ttnf} highlights the capacity of tensor trains to model larger tensors using significantly fewer parameters compared to TensoRF, offering, in principle, a more scalable approach. However, TT-NF lacks efficient optimization strategies aligned to TensoRF. TT-NF often reaches a local minimum in optimization and cannot model the full potential of tensor trains, resulting in poorer performance. (2). 
They often struggle with noisy and incomplete data. 

Toward resolving these obstacles, we propose a novel, coarse-to-fine, tensor-train-based representation called `Prolongation Upsampling Tensor Train (PuTT)' and an associated gradient-based optimization strategy.  
Our strategy involves the prolonging or `upsampling' of a learned tensor train representation in a coarse-to-fine manner, creating a sequence of `coarse-to-fine' tensor trains that are incrementally refined. The coarse-to-fine operations are performed directly in the tensor train format, circumventing the need to process individual data points, as is customary in linear interpolation. This allows for efficient sampling and for PuTT to be applied to extremely fine grids and to fit high-resolution data at limited additional memory and computational cost.

Central to PuTT is its unique capability to learn Quantized Tensor Trains (QTT)~\cite{qtt}, a tensor format designed for efficient representation through hierarchical structuring. It does so through a hierarchical refinement process, reminiscent of the multigrid method for solving partial differential equations~\cite{multigrid}.
Our iterative coarse-to-fine representation and optimization strategy enables significant parameter efficiency of tensor trains on the one hand and their efficient optimization on the other. The empirical findings indicate that our strategy effectively mitigates the impact of local minima during training, thereby enabling a parametrized learning model to approach the theoretical limit allowed within the TT format; as obtained by the TT-SVD decomposition~\cite{TT_decomp_oseledets}. Unlike TT-SVD, our strategy employs learning from samples, and benefits from the associated versatility.

Visual data is known to have a natural hierarchical structure, as evidenced by the success of wavelet-based methods in image and video compression~\cite{wavelets_and_quantization_vision_data}. To this end, the adoption of QTT in PuTT is motivated by its suitability for handling the multi-resolution nature and dependencies within subregions of data, akin to those in visual data. 
Compared to other tensor decompositions like CP and Tucker, which scale linearly with dimension (d) and side length (L),
QTT offers a more modest scaling in side length $\mathcal{O}(d \log(L)R^2)$, where $R$ is the rank and captures the correlation in the data across dimensions and resolution scale. This scalability positions QTT as an effective choice for large-scale tensors, surpassing CP, VM, and Tucker in space efficiency, which is especially beneficial as dimensions and resolutions increase. 
We evaluate our representation along three axes: (1). compression, (2). denoising capability, and (3). ability to learn from incomplete or missing data. To value these axes, we consider the tasks of image fitting, 3D fitting, and novel view synthesis, where our method shows improved qualitative and quantitative performance compared to state-of-the-art tensor-based baselines. 

\section{Related Work}
\label{sec:related}

\noindent \textbf{Coarse-to-Fine Representations} \quad
Coarse-to-fine, or multi-scale, visual representations were developed for several purposes, such as compression or reducing transmission time~\citep{szeliski2022computer}. Early works include multi-resolution pyramids such as Laplacian and Gaussian pyramids~\citep{adelson1984pyramid}, half-octave pyramids~\citep{crowley1984fast} and wavelets~\citep{mallat1989theory}. 
Recently, coarse-to-fine representations were used to build neural fields~\cite{yang2022polynomial, lindell2022bacon}. 
Similarly, Instant-NGP~\cite{instantNerf} uses a learned mutual-resolution grid representation as a fast and compact representation for novel view synthesis. Similarly to our method, this representation is learned using SGD. Our method, however, considers a more general and compact learned multi-scale representation, whereby the hierarchy levels are tensor trains. 

\noindent \textbf{Compact Visual Representations} \quad
Visual representations can be divided between explicit, implicit, and hybrid representations. Explicit representations, such as image grids, 3D meshes, 
point clouds
and voxel grids
are interpretable and expressive but expensive to store and scale. 
Implicit  representations~\cite{sitzmann2019deepvoxels,lombardi2019neural} were used for applications such as
novel view synthesis~\cite{instantNerf},
generative modeling~\cite{niemeyer2021giraffe,chan2021pi,chan2022efficient}, and surface reconstruction~\cite{oechsle2021unisurf}.
While compact, they are known to be slow. 
To this end, hybrid representations were developed. For instance, DVGO~\cite{DVGO} %
optimize voxel grids of features. 
This leads to significantly lower memory and capacity requirements. 
Instant-NGP~\cite{instantNerf} combines an explicit multi-resolution hashable feature grid with small MLP mapping features to color and opacity values. 
While our approach also employs a multiresolution structure, it is centered around a factorization of a quantized hierarchical structure in each layer.
Hashing is an orthogonal technique; combining it with our method is left for future work.

\noindent \textbf{Tensor network representations} \quad
Large-scale tensor networks have seen extensive applications in quantum physics~\cite{whiteDMRG}, including quantum computation~\cite{SupremSim}, and can be used to simulate strongly correlated systems~\cite{orusTN}. 
To learn visual representations, TensoRF~\cite{tensoRF} significantly improved compression by using CP and VM decomposition for NeRFs. CP decomposition was later extended by~\cite{strivec}. Instead, our method can handle a much more expressive representation of tensor trains. 
Our method extends the potential of tensor trains for larger tensors, as noted in TT-NF~\cite{ttnf}, by incorporating a coarse-to-fine approach that ensures efficient optimization similar to TensoRF and enhances denoising and image completion capabilities. This strategy helps to fully utilize the expressive strength of tensor-train representations.

\section{Notation}
\label{sec:notation}

\noindent \textbf{Graphical notation} \quad
Tensor Network Notation (TNN) provides a powerful tool for visualizing the interactions between tensors in a tensor network. In these visualizations (diagrams), each tensor is represented as a node, with a number of legs corresponding to its dimensions. 
For example: (i). A matrix $W\in\mathbb{R}^{m \times n}$ is as a node with two legs: \tikz[baseline=-0.5ex]{
    \node[draw, circle, inner sep=1pt] (tensor) {\scriptsize W};
    \draw (tensor) -- ++(-0.4,0);
    \draw (tensor) -- ++(0.4,0);
}. (ii). A vector \( x \in \mathbb{R}^n \) is a node with a single leg: \tikz[baseline=-0.5ex]{
    \node[draw, circle, inner sep=0.5pt] (tensor2) at (0.3,0) {x}; %
    \draw (tensor2) --(-0.1,0);
}. (iii). Vector-matrix multiplication $W x$ is depicted by  \textit{contracting} (summing over) legs of the connected tensors: \tikz[baseline=-0.5ex]{
    \node[draw, circle, inner sep=0.5pt] (tensor1) {\scriptsize W};
    \node[draw, circle, inner sep=0.5pt] (tensor2) at (0.6,0) {x}; %
    \draw (tensor1) -- (tensor2);
    \draw (tensor1) -- ++(-0.3,0);
}. (iv). Larger tensors $T\in \mathbb{R}^{m\times n\times r}$ have more legs: \tikz[baseline=-0.5ex]{
    \node[draw, circle, inner sep=1pt] (tensor) {\scriptsize T};
    \draw (tensor) -- ++(-0.3,0); %
    \draw (tensor) -- ++(0.3,0);  %
    \draw (tensor) -- ++(0,0.3);  %
}. 

Such diagrams, composed of nodes and their interconnections, effectively depict what is known as a ``tensor network''. We focus on four classes of tensor networks: the CP, Tucker, and tensor train decompositions. 
In 3D, we also consider the VM construction~\cite{tensoRF}. 
For discussion and details on these tensor networks, see the appendix.

\begin{figure}[]
    \centering
    \includegraphics[width=0.9\linewidth]{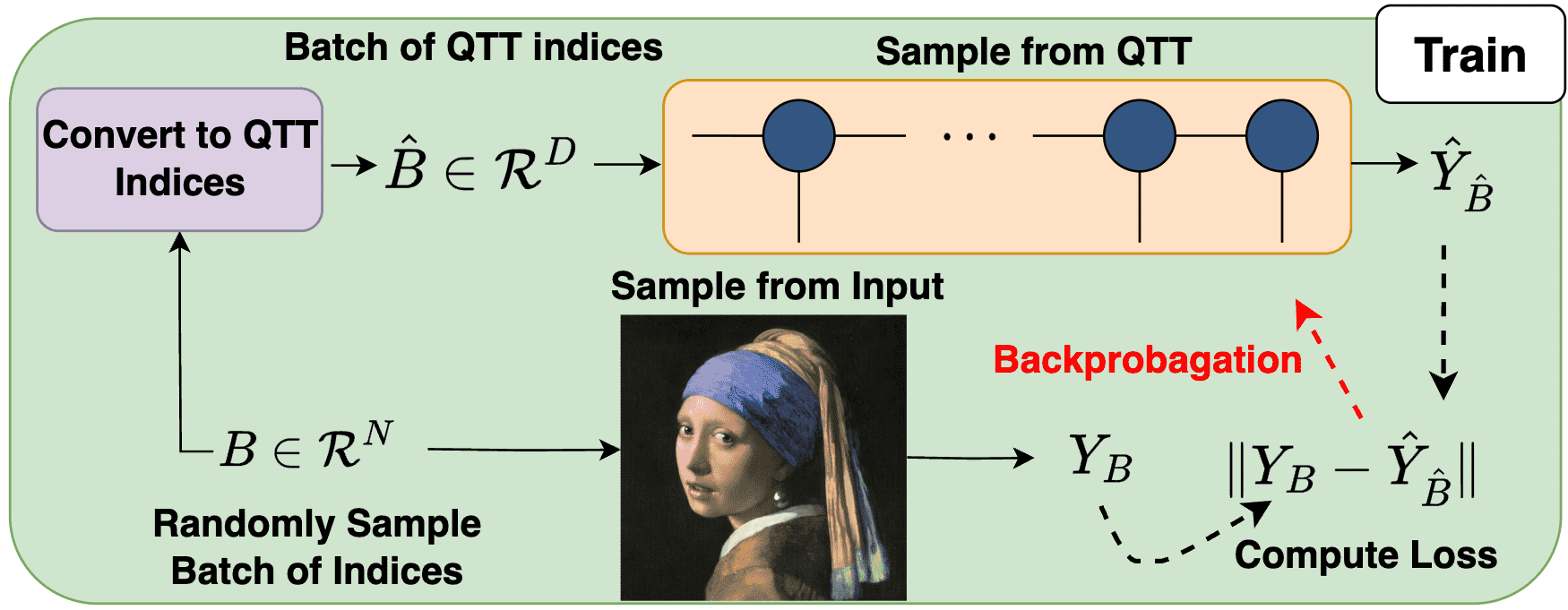}
    \caption{Training one level of hierarchy. Initially, a batch $B$ serves two purposes: (1). sample data points $Y_B$ from the target input, (2). it is 
    \textcolor{qtt_sampling_purple}{transformed into QTT indices $\hat{B}$}. Subsequently, corresponding values 
    $\hat{Y}_{\hat{B}}$ 
    are \textcolor{qtt_yellow}{sampled from the QTT}. The reconstruction loss between  
    $Y_B$ and $\hat{Y}_{\hat{B}}$ is used and \textcolor{red}{backpropagated}.} %
    \label{fig:train}
    \vspace{-0.3cm}
\end{figure}

\begin{figure}[]
    \centering
    \includegraphics[width=0.9\linewidth]{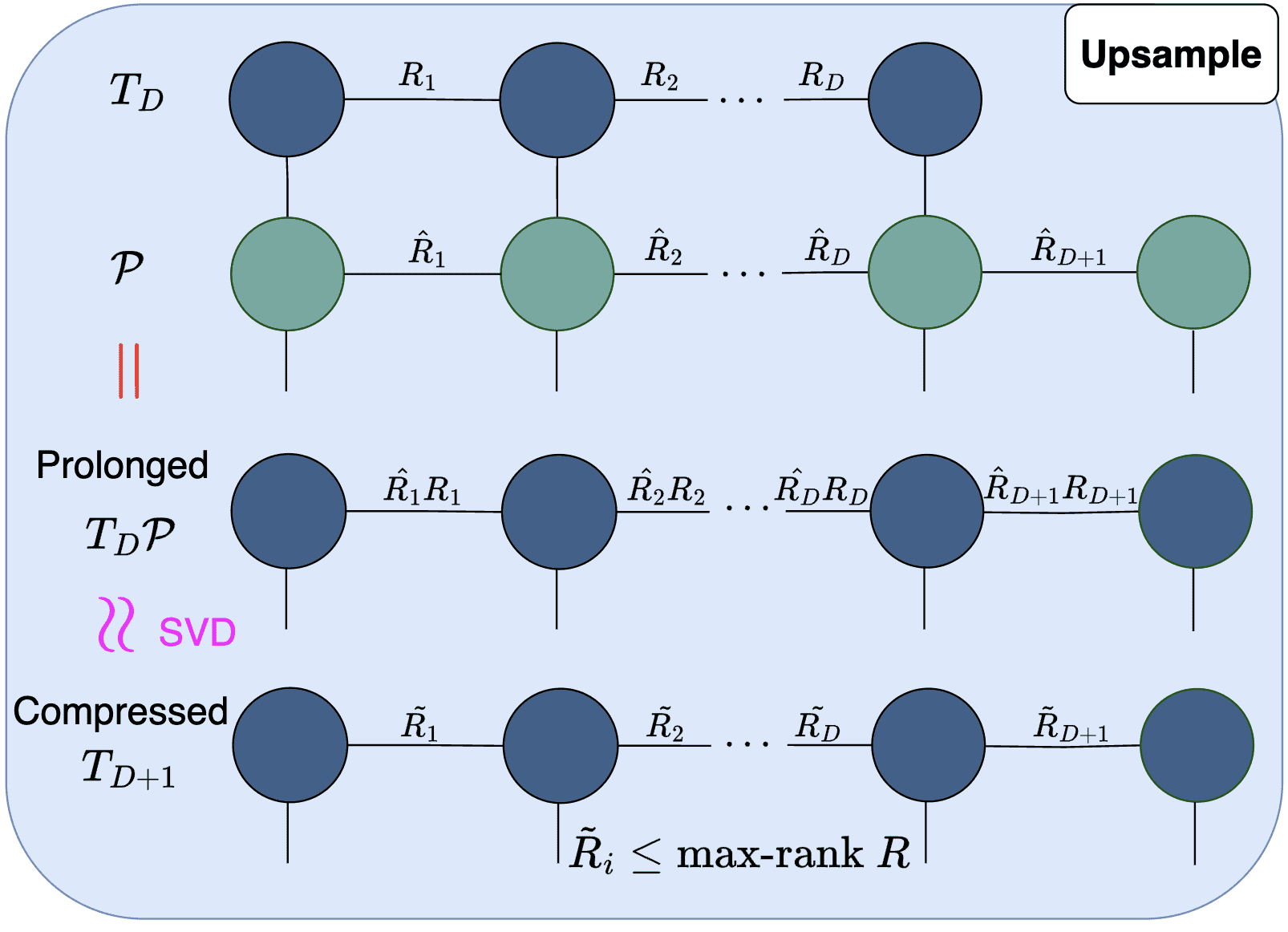}
    \caption{Illustration of QTT upsampling process. Beginning with a QTT $T_D$ of length $D$, upsampling is achieved through the prolongation MPO $\mathcal{P}$. This involves connecting the $D+1$ cores of $\mathcal{P}$ to the corresponding cores of $T_D$ and then contracting their shared indices. The result is a new QTT, $T_{D+1}$, of length $D+1$, where ranks are increased to $R_i\hat{R}_i$. To manage this rank growth, \textcolor{brightpurple}{TT-SVD}~\cite{TT_decomp_oseledets} is employed for rank truncation. This process yields $T_{D+1}$ with controlled ranks $\tilde{R}_i \leq R_{max}$. 
    }
\label{fig:prolongation_method}
    \vspace{-0.3cm}
\end{figure}

\begin{figure}[]
    \centering
    \includegraphics[width=0.9\linewidth]{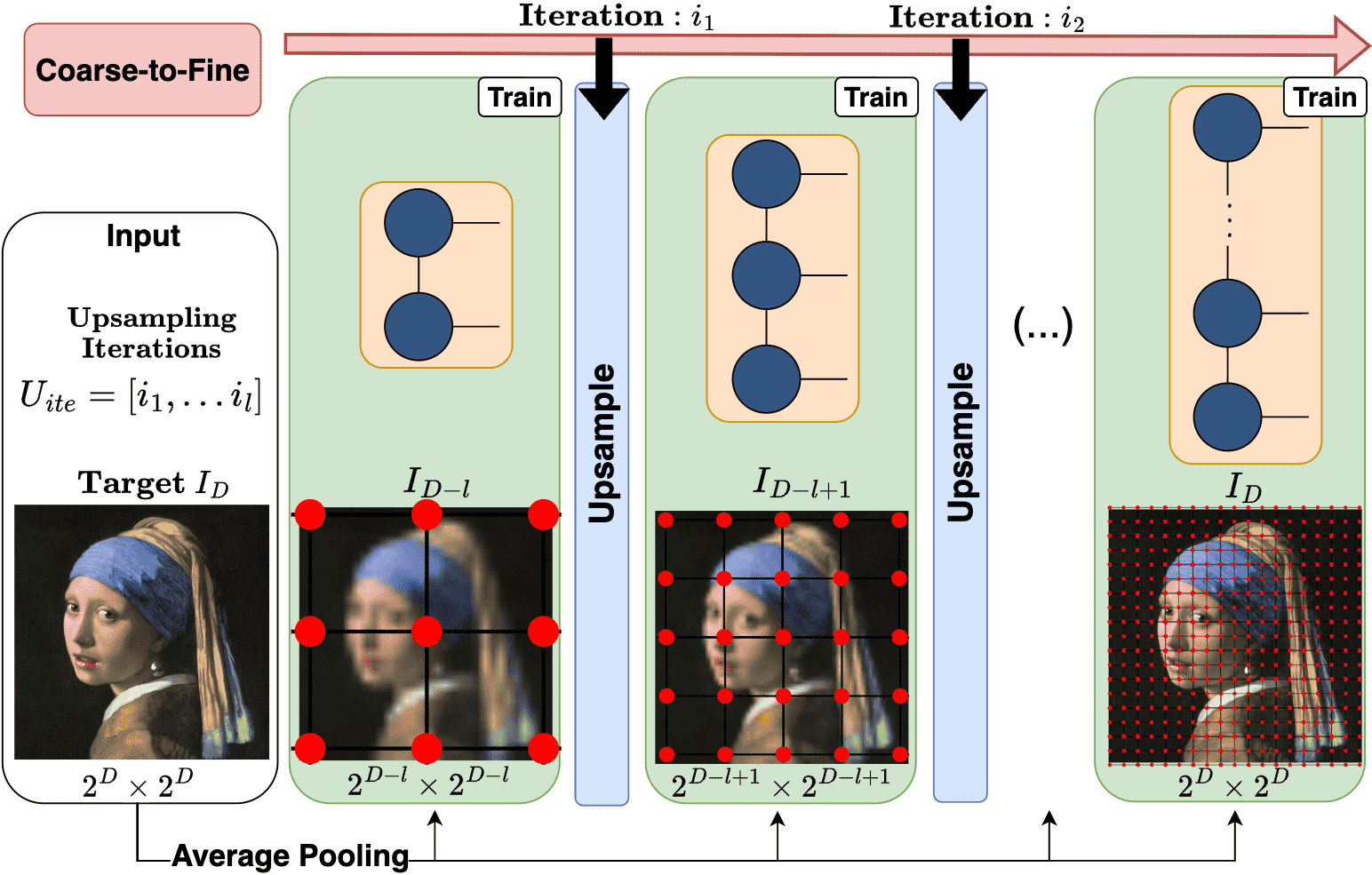}
    \caption{Full coarse-to-fine learning, integrating `Train' and `Upsample' phases of Fig.\ref{fig:train} and Fig.\ref{fig:prolongation_method}. 
    We start with input $I_D$, which is downscaled to $I_{D-l}$ (resolution $2^{d-l}\times 2^{d-l}$). %
    A \textcolor{qtt_yellow}{QTT} is then randomly initialized for $I_{D-l}$. After \textcolor{train_green}{training} this QTT (as per Fig.\ref{fig:train}) up to iteration $i_1$, we proceed with \textcolor{upsample_blue}{upsampling} (as illustrated in Fig.~\ref{fig:prolongation_method}), producing $T_{D-l+1}$ of length $D-l+1$. This represents a grid at resolution $2^{d-l+1}\times 2^{d-l+1}$ and requires sampling from a newly downsampled target $I_{D-l+1}$ of the same resolution. 
    } 
    \label{fig:coarse-to-fine}
    \vspace{-0.8cm}
\end{figure}

\textit{Tensor Train~\cite{TT_decomp_oseledets}}  %
or a Matrix Product State (MPS)~\cite{MPSold,MPS_Paper}, is a decomposition of a tensor with $N$ indices into a chain-like network of $N$ order-$3$ tensors: 
This structure can be graphically represented in Tensor Network Notation: \tikz[baseline=-0.5ex]{
    \node[draw, circle, inner sep=1pt] (tensor1) {\scriptsize A};
    \node[draw, circle, inner sep=1pt, right=0.3cm of tensor1] (tensor2) {\scriptsize A};
    \node[draw, circle, inner sep=1pt, right=0.3cm of tensor2] (tensor3) {\scriptsize A};
    \node[draw, circle, inner sep=1pt, right=0.3cm of tensor3] (tensor4) {\scriptsize A};
    \node[draw, circle, inner sep=1pt, right=0.3cm of tensor4] (tensor5) {\scriptsize A};

    \draw (tensor1) -- (tensor2);
    \draw (tensor2) -- (tensor3);
    \draw (tensor3) -- (tensor4);
    \draw (tensor4) -- (tensor5);

    \draw (tensor1) -- ++(0,0.3);
    \draw (tensor2) -- ++(0,0.3);
    \draw (tensor3) -- ++(0,0.3);
    \draw (tensor4) -- ++(0,0.3);
    \draw (tensor5) -- ++(0,0.3);
}.
Given a tensor $T\in \mathbb{R}^{n_1\times n_2\times \cdots n_d}$, the entries of the tensor are: 
\begin{equation}\label{eq:TT}
    T_{j_1\cdots j_d}
    =A^{j_1}A^{j_2}\cdots A^{j_d}, 
\end{equation}
where $\{A^{j_k}\}_{j_k=1}^{n_k}$ are $R_{j_k}\times R_{j_{k+1}}$ matrices for each $j_k$; i.e. \tikz[baseline=-0.3ex]{
    \node[draw, circle, inner sep=1pt] (tensor) {\scriptsize A};
    \draw (tensor) -- ++(-0.3,0); %
    \draw (tensor) -- ++(0.3,0);  %
    \draw (tensor) -- ++(0,0.3);  %
} are $R_k\times n_k \times R_{k+1}$ tensors. The ranks $R_k$ are referred to as the \textit{bond dimensions} of the tensor train. We distinguish the \textit{physical indices} $j_k$ from the \textit{virtual indices} $r_k$. 
The bond dimensions $R_k$ can be understood as a rank truncation along the cut $n_1,...,n_{k-1}$ and $n_k,...,n_d$, only keeping the $R$ largest singular values.\\

In the context of tensor trains, the rank measures the complexity or information content of the tensor. 
The largest rank among all virtual indices is considered the rank of the tensor train. This rank dictates the ultimate expressivity of the tensor train by determining the number of variational parameters.
A tensor $T$ can be represented exactly with a maximal bond dimension $R=\max_k \min\{n_1\cdots n_{k-1},n_k\cdots n_D\}$, which scales exponentially in $D$. Tensor trains are used as a parameterized variational class, where we upper bound the bond dimension at some maximal value $R_{\rm max}$. The maximum bond dimension governs the model's expressivity. 

\noindent \textbf{QTT} \quad
\label{subsec:QTT_notation}
Compressing visual data along physical dimensions (e.g., image height or width) is natural, as in the CP and Tucker decompositions~\cite{tensoRF}. However, significantly more compact representations are possible by compression along the `scaling dimension'.  
For that, we consider the \textit{quantized tensor train} (QTT) construction~\cite{QTT_paper}.  
The QTT format builds upon mode quantization, which decomposes the scaling dimension in powers of two. 
As an illustrative example~\cite{ttnf}, consider a 3D tensor representing a (regular) function on a $16\times16\times16$ lattice. Through mode quantization, this tensor can be recast as a 12-dimensional hypercube: $(\textcolor{red}{2_1} \times \textcolor{red}{2_2} \times \textcolor{red}{2_3} \times \textcolor{red}{2_4}) \times (\textcolor{green}{2_1} \times \textcolor{green}{2_2} \times \textcolor{green}{2_3} \times \textcolor{green}{2_4}) \times (\textcolor{blue}{2_1} \times \textcolor{blue}{2_2} \times \textcolor{blue}{2_3} \times \textcolor{blue}{2_4})$: \tikz[baseline=-0.5ex]{
    \node[draw, circle, inner sep=1pt] (tensor1) {\scriptsize Q};
    \node[draw, circle, inner sep=1pt, right=0.3cm of tensor1] (tensor2) {\scriptsize Q};
    \node[draw, circle, inner sep=1pt, right=0.3cm of tensor2] (tensor3) {\scriptsize Q};
    \node[draw, circle, inner sep=1pt, right=0.3cm of tensor3] (tensor4) {\scriptsize Q};
    \node[draw, circle, inner sep=1pt, right=0.3cm of tensor4] (tensor5) {\scriptsize Q};
    \node[draw, circle, inner sep=1pt, right=0.3cm of tensor5] (tensor6) {\scriptsize Q};
    \node[draw, circle, inner sep=1pt, right=0.3cm of tensor6] (tensor7) {\scriptsize Q};
    \node[draw, circle, inner sep=1pt, right=0.3cm of tensor7] (tensor8) {\scriptsize Q};
    \node[draw, circle, inner sep=1pt, right=0.3cm of tensor8] (tensor9) {\scriptsize Q};
    \node[draw, circle, inner sep=1pt, right=0.3cm of tensor9] (tensor10) {\scriptsize Q};
    \node[draw, circle, inner sep=1pt, right=0.3cm of tensor10] (tensor11) {\scriptsize Q};
    \node[draw, circle, inner sep=1pt, right=0.3cm of tensor11] (tensor12) {\scriptsize Q};

    \draw (tensor1) -- (tensor2);
    \draw (tensor2) -- (tensor3);
    \draw (tensor3) -- (tensor4);
    \draw (tensor4) -- (tensor5);
    \draw (tensor5) -- (tensor6);
    \draw (tensor6) -- (tensor7);
    \draw (tensor7) -- (tensor8);
    \draw (tensor8) -- (tensor9);
    \draw (tensor9) -- (tensor10);
    \draw (tensor10) -- (tensor11);
    \draw (tensor11) -- (tensor12);

    \draw (tensor5) -- ++(0,0.3);
    \draw (tensor6) -- ++(0,0.3);
    \draw (tensor7) -- ++(0,0.3);
    \draw (tensor8) -- ++(0,0.3);
    \draw (tensor1) -- ++(0.3,0.3);
    \draw (tensor2) -- ++(0.3,0.3);
    \draw (tensor3) -- ++(0.3,0.3);
    \draw (tensor4) -- ++(0.3,0.3);
    \draw (tensor9) -- ++(-0.3,0.3);
    \draw (tensor10) -- ++(-0.3,0.3);
    \draw (tensor11) -- ++(-0.3,0.3);
    \draw (tensor12) -- ++(-0.3,0.3);
}. In this notation, the subscript indicates levels of hierarchy, and parentheses are used for visual convenience.
QTT assumes a particular reordering, where each hierarchical level is grouped as: $(\textcolor{red}{2_1} \times \textcolor{green}{2_1} \times \textcolor{blue}{2_1}) \cdot (\textcolor{red}{2_2} \times \textcolor{green}{2_2} \times \textcolor{blue}{2_2}) \cdot (\textcolor{red}{2_3} \times \textcolor{green}{2_3} \times \textcolor{blue}{2_3}) \cdot (\textcolor{red}{2_4} \times \textcolor{green}{2_4} \times \textcolor{blue}{2_4})$: \tikz[baseline=-0.5ex]{
    \node[draw, circle, inner sep=1pt] (tensor1) {\scriptsize Q};
    \node[draw, circle, inner sep=1pt, right=0.3cm of tensor1] (tensor2) {\scriptsize Q};
    \node[draw, circle, inner sep=1pt, right=0.3cm of tensor2] (tensor3) {\scriptsize Q};
    \node[draw, circle, inner sep=1pt, right=0.3cm of tensor3] (tensor4) {\scriptsize Q};

    \draw (tensor1) -- (tensor2);
    \draw (tensor2) -- (tensor3);
    \draw (tensor3) -- (tensor4);

    \draw (tensor1) -- ++(0,0.3);
    \draw (tensor2) -- ++(0,0.3);
    \draw (tensor3) -- ++(0,0.3);
    \draw (tensor4) -- ++(0,0.3);
    \draw (tensor1) -- ++(0.3,0.3);
    \draw (tensor2) -- ++(0.3,0.3);
    \draw (tensor3) -- ++(0.3,0.3);
    \draw (tensor4) -- ++(0.3,0.3);
    \draw (tensor1) -- ++(-0.3,0.3);
    \draw (tensor2) -- ++(-0.3,0.3);
    \draw (tensor3) -- ++(-0.3,0.3);
    \draw (tensor4) -- ++(-0.3,0.3);
}. This ordering is far more expressive for hierarchical data, as it avoids bottlenecks connecting low-frequency data in one dimension (say $x$) to high-frequency data in another dimension (say $y$). 

\noindent \textbf{Matrix Product operators} \quad 
A matrix product operator (MPO) is a tensor network that describes a factorization of a tensor with $d$ input and $d$ output indices into a sequence of smaller tensors~\cite{MPO_link1, MPO_link2}, with each core having two physical and two virtual indices: \tikz[baseline=-0.5ex]{
    \node[draw, circle, inner sep=1pt] (tensor1) {\scriptsize M};
    \node[draw, circle, inner sep=1pt, right=0.3cm of tensor1] (tensor2) {\scriptsize M};
    \node[draw, circle, inner sep=1pt, right=0.3cm of tensor2] (tensor3) {\scriptsize M};
    \node[draw, circle, inner sep=1pt, right=0.3cm of tensor3] (tensor4) {\scriptsize M};
    \node[draw, circle, inner sep=1pt, right=0.3cm of tensor4] (tensor5) {\scriptsize M};

    \draw (tensor1) -- (tensor2);
    \draw (tensor2) -- (tensor3);
    \draw (tensor3) -- (tensor4);
    \draw (tensor4) -- (tensor5);

    \draw (tensor1) -- ++(0,0.3);
    \draw (tensor2) -- ++(0,0.3);
    \draw (tensor3) -- ++(0,0.3);
    \draw (tensor4) -- ++(0,0.3);
    \draw (tensor5) -- ++(0,0.3);
    \draw (tensor1) -- ++(0,-0.3);
    \draw (tensor2) -- ++(0,-0.3);
    \draw (tensor3) -- ++(0,-0.3);
    \draw (tensor4) -- ++(0,-0.3);
    \draw (tensor5) -- ++(0,-0.3);
}.
Tensor trains and MPOs differ fundamentally in their applications: Tensor trains serve as compressed representations of large vectors in high-dimensional spaces, while MPOs, denoting matrix-vector operations, act on tensor trains. This distinction enables MPOs to effectively handle large operators within these high-dimensional spaces in a compressed form~\cite{tensornetwork_org}.

In the appendix, we provide additional details and visualizations of tensor networks. In Sec.~\ref{app:multi_res_qtt}, Fig.~\ref{fig:coarse_to_fine_struc} visualizes the multi-resolution hierarchy of the QTT, illustrating its coarse-to-fine structure.
In Sec.~\ref{app:mpo}, we detail the structure of the specific MPO used for upsampling and the complexity of the prolongation operations.

\section{Method}
\label{sec:method}

\noindent \textbf{Prolongation operators} \quad
While QTT allows for substantial memory savings for equal representational power, it suffers from instabilities in training. Indeed, as observed in~\cite{ttnf, TTdensity}, without explicit regularization, the QTT representation often fails to improve upon the CP and Tucker representations in practice, getting easily stuck in local minima. To this end, we introduce a global upsampling strategy inspired by the multigrid method for partial differential equations~\cite{multigrid,TT-Prolongation}, illustrated in Figs.~\ref{fig:train}-\ref{fig:coarse-to-fine} and below. 

We use a specific Matrix Product Operator (MPO) -- the prolongation operator $\mathcal{P}$ -- which performs linear interpolation \textit{globally} in QTT.
Given a vector $v_n\in\mathbb{R}^{2^n}$ (e.g., a waveform), $\mathcal{P}_n$ is a  matrix that maps $v_n$ to $v_{n+1}\in\mathbb{R}^{2^{n+1}}$, where $v_{n+1}$ is a linear interpolant of $v_n$ to a lattice of twice the resolution. $\mathcal{P}_n$ is similar to the Haar wavelet transform. Crucially, the matrix $\mathcal{P}_n$ is an MPO with a fixed bond dimension of 3. Therefore, it can be applied to a QTT via local tensor operations. The resulting QTT will have (three times) larger bond dimension. If the maximal bond dimension exceeds the desired threshold ($R_{\rm max}$), then the QTT can be compressed down to $R_{\rm max}$ via a local singular value decomposition.  For two or three-dimensional visual data, we interpolate in each dimension by applying $\mathcal{P}_n\otimes\mathcal{P}_n\otimes \mathcal{P}_n$.\\
Fig.~\ref{fig:prolongation_method} demonstrates the application of $\mathcal{P}$ in upsampling a QTT $T_D$ of length $D$ to a prolonged QTT, $T_{D+1}$, of length $D+1$. This is achieved by linking $D$ input cores of $\mathcal{P}$ to the corresponding cores of $T$ and contracting the shared indices. As a result, the ranks of $\mathcal{P} T_{D}$ increase, necessitating a rank reduction through \textcolor{brightpurple}{TT-SVD}~\cite{TT_decomp_oseledets}. This reduction is critical since the ranks expand exponentially in the number of upsampling steps.
The outcome is a compressed QTT, $T_{D+1}$, with controlled ranks $\tilde{R}_i \leq R_{\rm max}$, ensuring an efficient representation of the upsampled object.

\subsection{Learning tensor trains (PuTT)}
We now outline our Prolongation Upsampling Tensor Train (\textbf{PuTT}) methodology.
Training starts at a chosen (coarse) resolution and progressively learns a QTT with finer and finer resolution. 
Consider a target vector $I_D$ representing a visual object of size $(2^d)^D$, where $d$ is the dimension (i.e. $d=2$ for images). We want to learn the best QTT approximation of $I_D$, $T_D$ with $D$ legs of dimension $2^d$ each. We start with a down-sampled object $I_{D-l}$ and learn $T_{D-l}$, where $l$ is the number of upsampling steps. For instance, for an image of resolution $512^2$ and three upsampling steps, we start learning $I_{6}$ and upsample three times to reach $I_{9}$.

In our experiments, we learn the QTTs at each resolution via back-propagation, by optimizing the mean squared error. 
After obtaining a satisfactory approximation of $I_{D-l}$ in QTT format, which we call $T_{D-l}$, we apply the upsampling MPO $\mathcal{P}_{D-l+1}$ to $T_{D-l}$, yielding a QTT with $D-l+1$ `physical' legs $\mathcal{P}_{D-l+1} T_{D-l}$. If the maximal bond dimension exceeds $R_{\rm max}$, then we perform a TT-SVD contraction to keep all ranks below the max $R_{\rm max}$. We then proceed to learn $I_{D-l+1}$ starting from $\mathcal{P}_{D-l+1} T_{D-l}$. 
Initiating learning with an upsampled prior stabilizes convergence to the correct local or global minimum.
The learning proceeds iteratively until we reach the full scale $I_D$ and a QTT approximation $T_D$.
Throughout the learning, we adopt a trapezoid structure~\cite{TT_decomp_oseledets} for the tensor train ranks, where ranks increase to a maximum (forming the trapezoid's ascending edge), remain constant (trapezoid's top), and then decrease (its descending edge). 
 The upsampling step of the algorithm is illustrated in Fig.~\ref{fig:prolongation_method}, and the full coarse-to-fine learning scheme is illustrated in Fig.~\ref{fig:coarse-to-fine}.
 
Fig.~\ref{fig:train} outlines our training process in a single hierarchy level. Initially, we sample a batch of indices $B$, which serves two purposes: 1. Conversion to a batch of QTT indices $\tilde{B}\in \mathcal{R}^D$, mapping cartesian points to QTT points for retrieving values $\hat{Y}_{\hat{B}}$ from the QTT, 2. Using $B$ to sample target input values from $I_D$, obtaining $Y_B$. These values, $Y_B$ and $\hat{Y}_{\hat{B}}$, are used to calculate a mean squared error loss, which is then backpropagated to update the QTT weights.

\subsection{Novel View Synthesis}
\label{sec:putt_nerf}
Applying PuTT to novel view synthesis involves modeling a function that maps any 3D location $x$ and viewing direction $d$ to a volume density $\sigma$ and a view-dependent color $c$, supporting differential ray marching for volume rendering~\cite{nerf}.
We use two voxel grids: $\mathcal{G}_\sigma$ for volume density and $\mathcal{G}_c$ for view-dependent color, similar to TensoRF~\cite{tensoRF}. $\mathcal{G}_\sigma$ is a single-channel grid encoding $\sigma$ values, while $\mathcal{G}_c$ is a multi-dimensional grid holding appearance features. These features are transformed into color values using a shading function $\mathcal{S}$, such as a small MLP or Spherical Harmonics. The continuous grid-based Radiance Field is expressed as:
$$ \sigma, c = \mathcal{G}_\sigma(x), \mathcal{S}(\mathcal{G}_c(x), d) $$
where $\mathcal{G}_\sigma(x)$ and $\mathcal{G}_c(x)$ are interpolated using trilinear interpolation within the 3D voxel grid.

The tensor representation, $\mathcal{G}_\sigma \in \mathbb{R}^{X \times Y \times Z}$, is a 3-order tensor for volume density, and $\mathcal{G}_c \in \mathbb{R}^{X \times Y \times Z \times P}$ is a 4-order tensor for color with an additional feature vector dimension $P$. We factorize these grids into QTT-format tensors $T_\sigma$ and $T_c$, each with $D$ cores and eight physical indices per core, using the QTT-Block structure~\cite{ttnf} for $\mathcal{G}_c$ to handle the additional feature dimension $P$.

\begin{figure*}
    \centering
    \includegraphics[width = 0.9\textwidth]{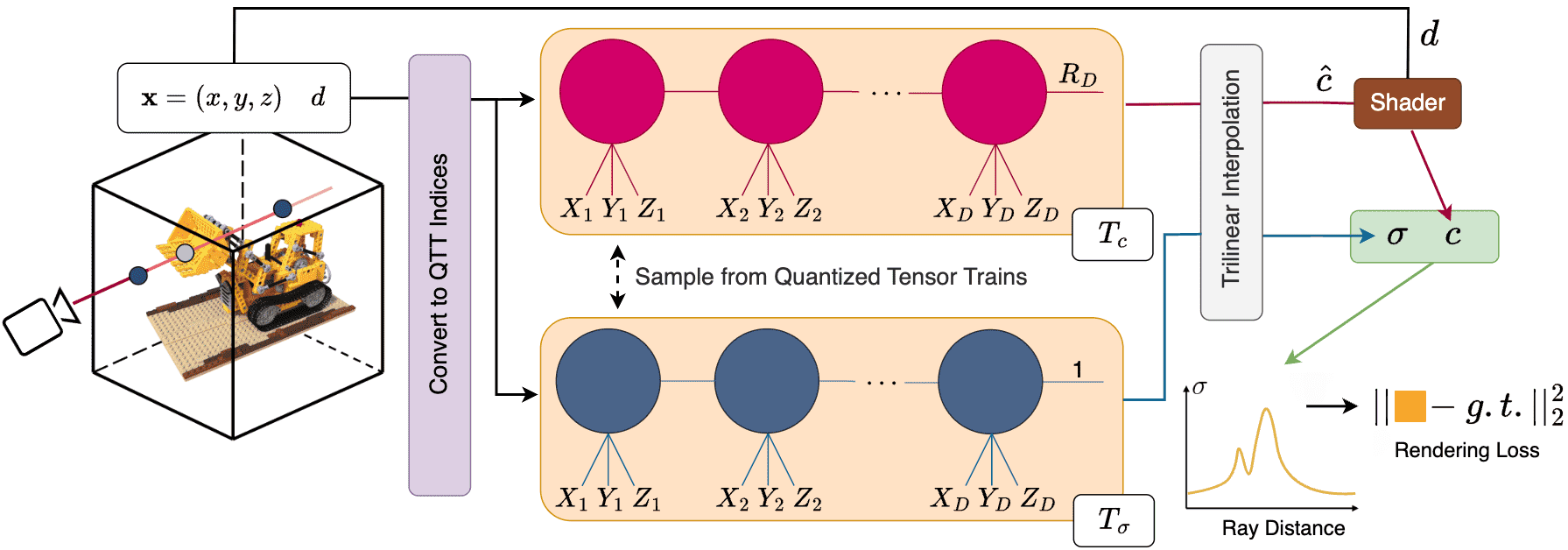}
    
    \caption{Novel view synthesis with PuTT. This illustration shows the process using two QTTs, \textcolor[rgb]{0.0,0.0,0.54}{$T_\sigma$} and \textcolor[rgb]{0.86,0.08,0.24}{$T_c$}, for computing volume density ($\sigma$) and view-dependent color ($c$) through differential ray marching. For each 3D location $x=(x,y,z)$ and viewing direction $d$, $x$ is \textcolor{qtt_sampling_purple}{converted to QTT indices}. These indices query the \textcolor[rgb]{0.0,0.0,0.54}{density voxel grid $T_\sigma$} and the \textcolor[rgb]{0.86,0.08,0.24}{color voxel grid $T_c$}. The process uses \textcolor{gray}{trilinear interpolation} to obtain continuous $\sigma$ values and appearance feature vectors $\hat{c}$. These vectors are processed through a \textcolor{brown}{shading module} to generate \textcolor[rgb]{0.6,0.87,0.54}{raw color values $c$}, which, along with \textcolor[rgb]{0.6,0.87,0.54}{$\sigma$}, are used in differential volumetric rendering. The rendering loss is computed by comparing the generated color values against ground truth (g.t.) values.}
    \vspace{-0.3cm}
    \label{fig:NeRF_model_diagram}
\end{figure*}

Fig.~\ref{fig:NeRF_model_diagram} outlines our reconstruction and rendering pipeline. Points sampled along a ray are transformed into QTT indices and used to sample from $T_\sigma$ and $T_c$. The interpolated values are then used for differentiable volume rendering, enabling training via backpropagation.

\section{Results}
\label{sec:results}

\begin{figure}
    \centering
    \begin{subfigure}[t]{0.24\linewidth}
        \captionsetup{justification=centering, font=tiny}
        \caption*{PuTT \\ PSNR 26.3, SSIM 0.72}
        \includegraphics[width=\linewidth]{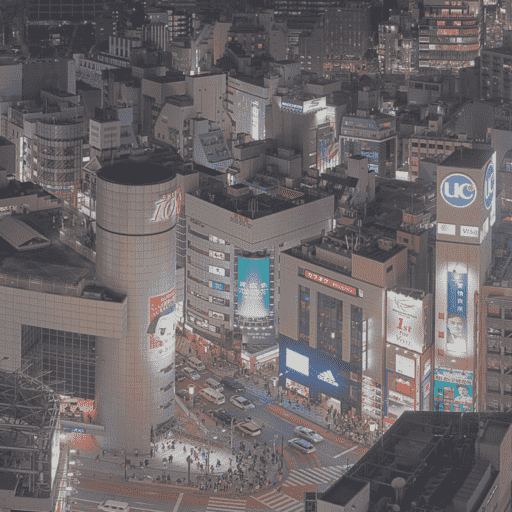}
    \end{subfigure}
    \begin{subfigure}[t]{0.24\linewidth}
        \captionsetup{justification=centering, font=tiny}
        \caption*{TT No Upsampling \\ PSNR 25.7, SSIM 0.70}
        \includegraphics[width=\linewidth]{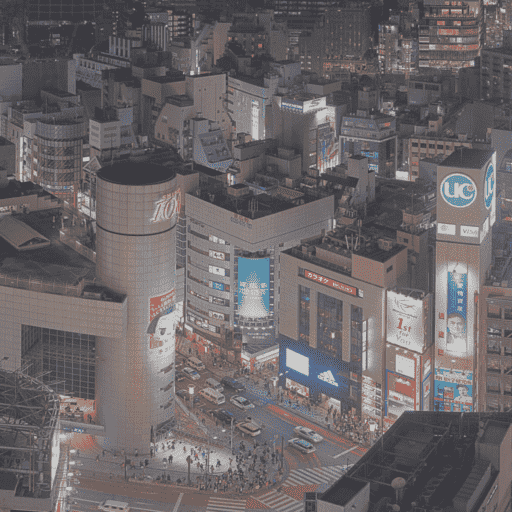}
    \end{subfigure}
    \begin{subfigure}[t]{0.24\linewidth}
        \captionsetup{justification=centering, font=tiny}
        \caption*{Tucker Upsampling \\ PSNR 21.8, SSIM 0.54}
        \includegraphics[width=\linewidth]{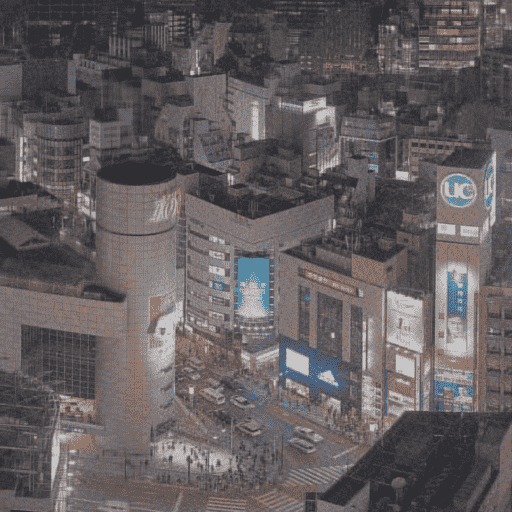}
    \end{subfigure}
    \begin{subfigure}[t]{0.24\linewidth}
        \captionsetup{justification=centering, font=tiny}
        \caption*{CP Upsampling \\ PSNR 21.6, SSIM 0.54}
        \includegraphics[width=\linewidth]{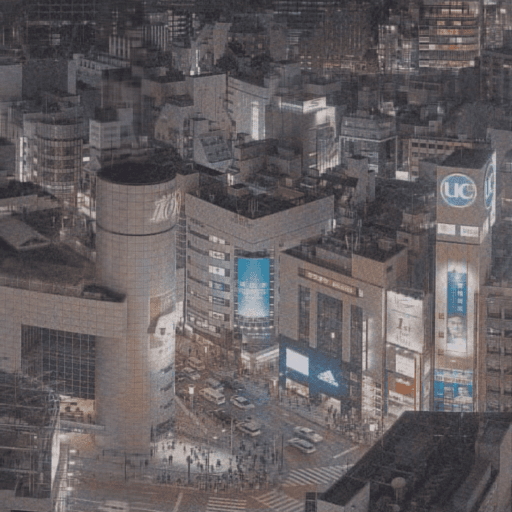}
    \end{subfigure}
    \captionsetup{justification=centering}
    \caption{Qualitative comparison to baselines on 16k images.}%
    \vspace{-0.4cm}
    \label{fig:visual_comp}
\end{figure}

\begin{figure*}
    \centering
\includegraphics[width=\linewidth]{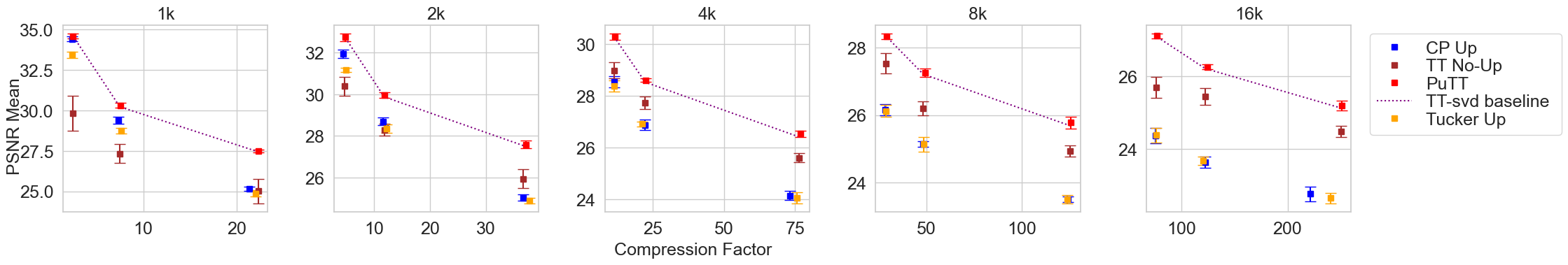} \\
    \vspace{-0.2cm}
    \caption{ %
    PSNR (y-axis) vs compression ratio (x-axis) for 2D fitting. ``up = upsampling".
    } 
    \label{fig:2d_psnr_ssim}
    \vspace{-0.4cm}
\end{figure*}

\begin{figure*}
\centering
\includegraphics[width=\linewidth]{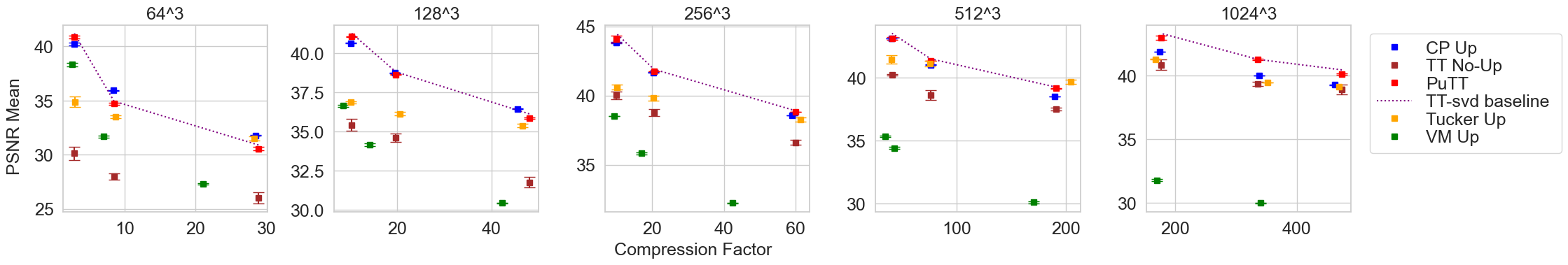} \\
\vspace{-0.2cm}
    \caption{(a) PSNR (y-axis) in comparison to compression ratio (x-axis) for 3D fitting. 
    } 
\label{fig:3d_psnr_ssim}
\vspace{-0.4cm}
\end{figure*}

\begin{figure*}
\begin{tabular}{cc}
    \includegraphics[trim={0 0 0cm 0},clip, width=0.48\linewidth] {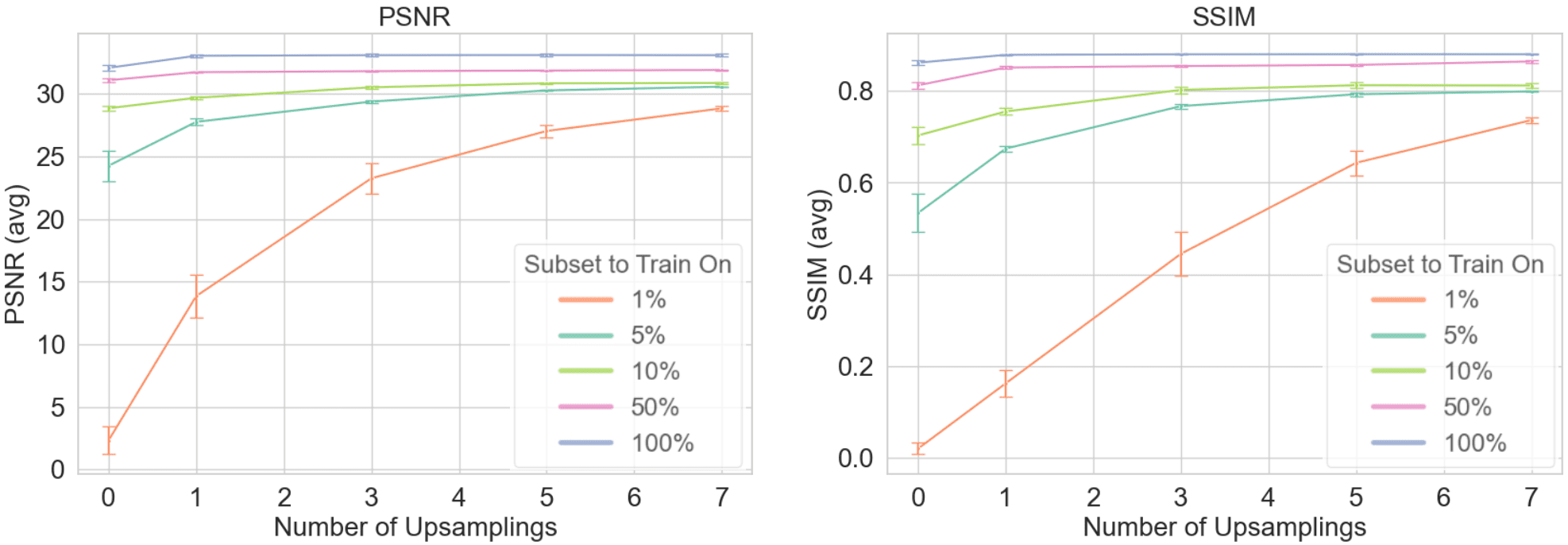} &  
    \includegraphics[trim={0 0 0cm 0},clip, width=0.48\linewidth]{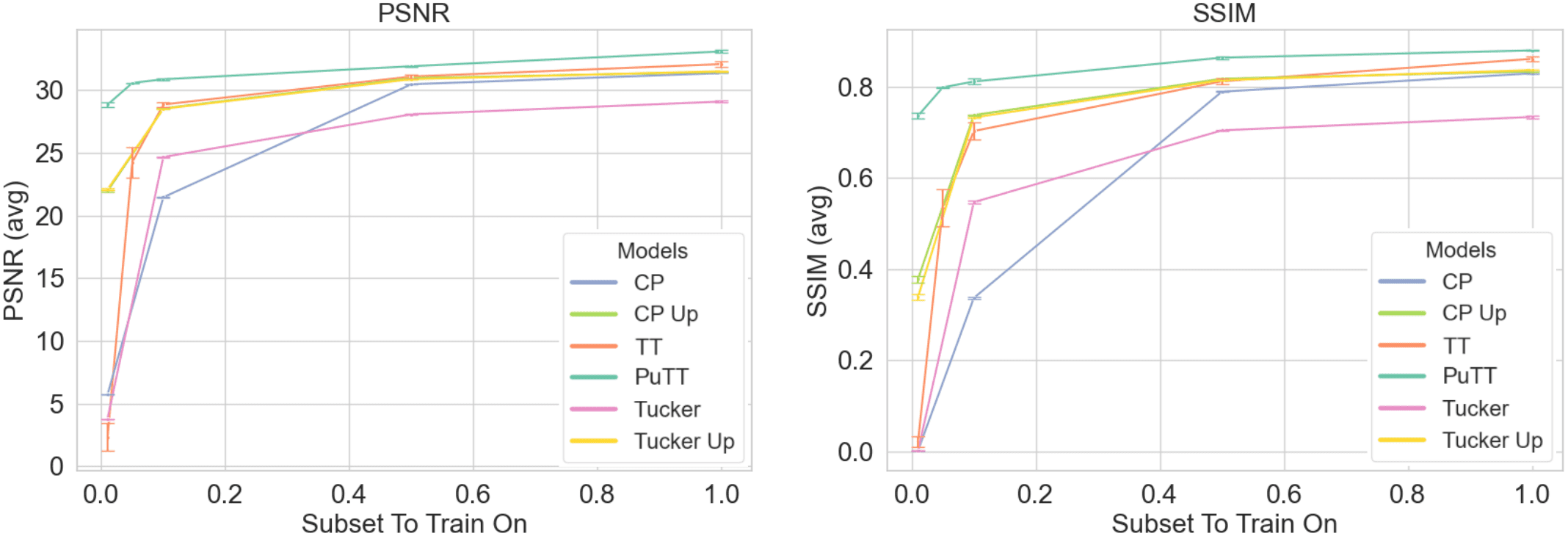} \\
    (a) & (b) 
    \vspace{-0.4cm}
\end{tabular}
    \caption{PSNR and SSIM vs number of upsampling steps for five different percentages of training data available (a) or vs different amounts of available training data available for three different models with or without upsampling (b). "Up" refers to upsampling.
    }
\label{fig:subsampling_num_upsamplings_and_subset}
\vspace{-0.2cm}
\end{figure*}

\begin{figure*} %
\centering
\begin{tabular}{c@{~}c@{~}c@{~}c@{~}c@{~}c@{~}c@{~}c}
\footnotesize{ } &
\footnotesize{Input} & 
\footnotesize{PuTT (Ours)} & 
\footnotesize{CP} & 
\footnotesize{Tucker} & 
\footnotesize{TT w/o up. } & 
\footnotesize{CP w/o up.} & 
\footnotesize{Tucker w/o up.} \\ 

\multicolumn{1}{c}{\rotatebox{90}{\parbox{2cm}{\centering 0.1 Noise}}} &
\includegraphics[width=0.12\linewidth]{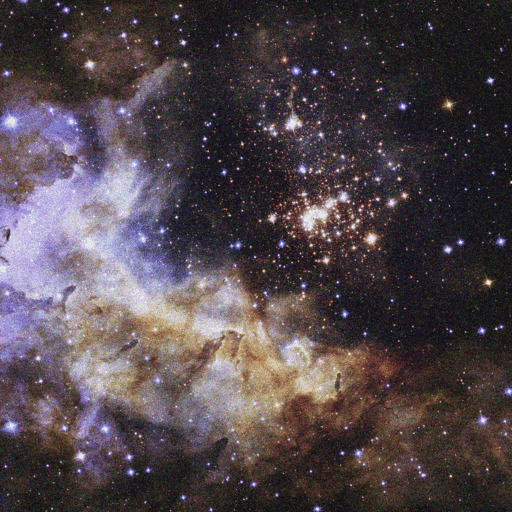} & 
\includegraphics[width=0.12\linewidth]{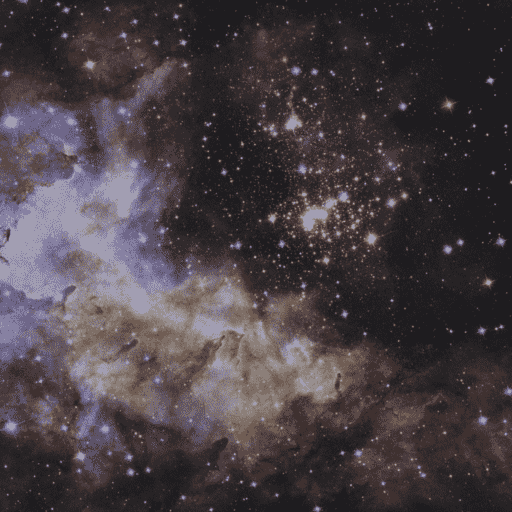} & 
\includegraphics[width=0.12\linewidth]{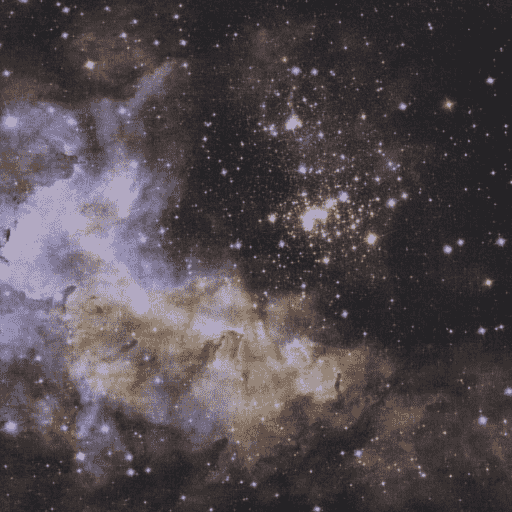} & 
\includegraphics[width=0.12\linewidth]{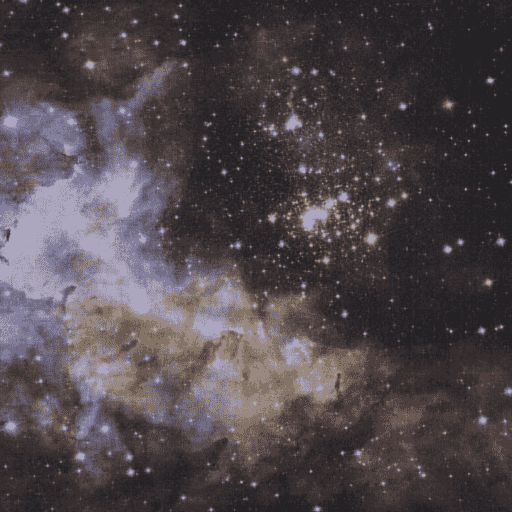} & 
\includegraphics[width=0.12\linewidth]{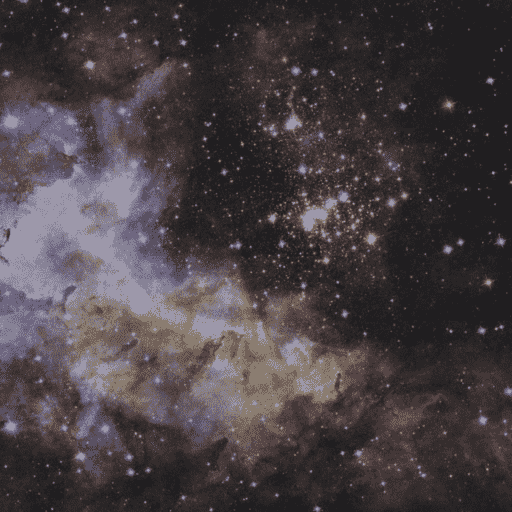} & 
\includegraphics[width=0.12\linewidth]{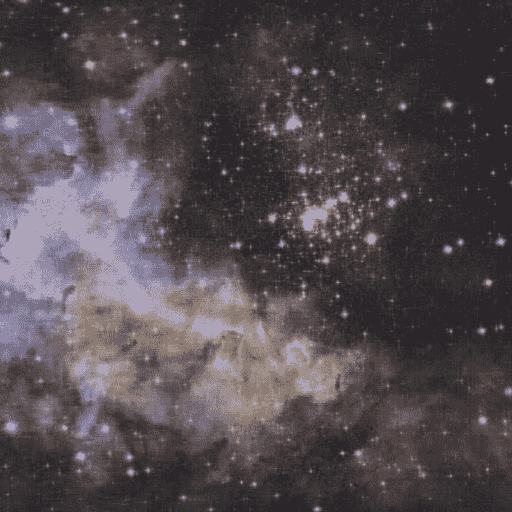} & 
\includegraphics[width=0.12\linewidth]{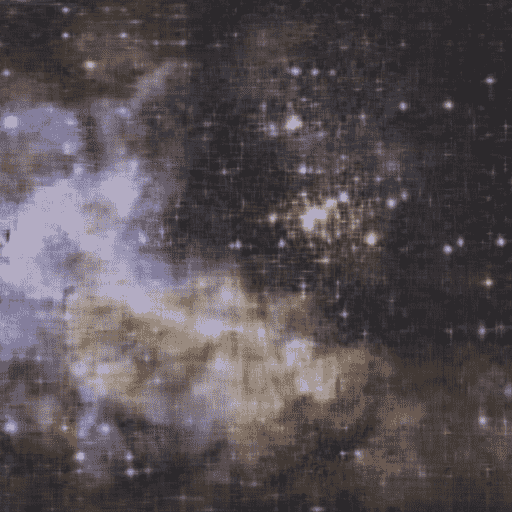} \\

\multicolumn{1}{c}{\rotatebox{90}{\parbox{2cm}{\centering 0.5 Noise}}} &
\includegraphics[width=0.12\linewidth]{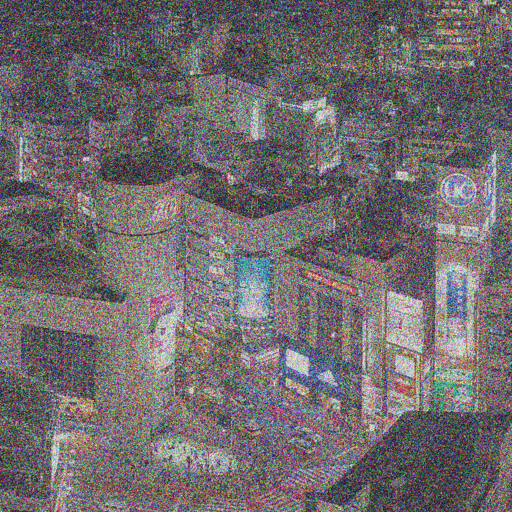} & 
\includegraphics[width=0.12\linewidth]{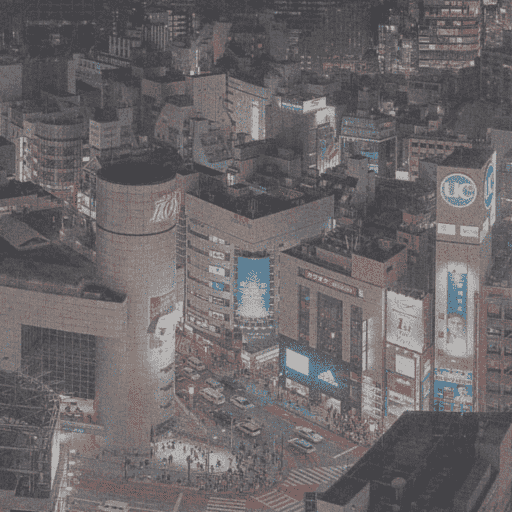} & 
\includegraphics[width=0.12\linewidth]{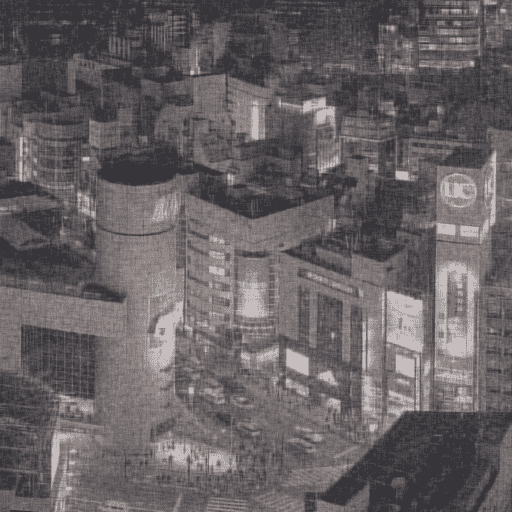} & 
\includegraphics[width=0.12\linewidth]{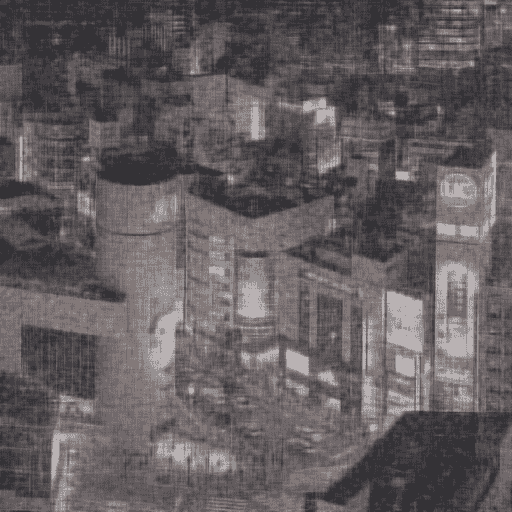} & 
\includegraphics[width=0.12\linewidth]{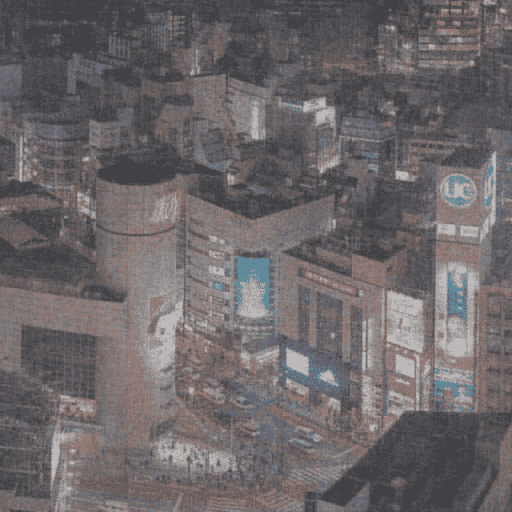} & 
\includegraphics[width=0.12\linewidth]{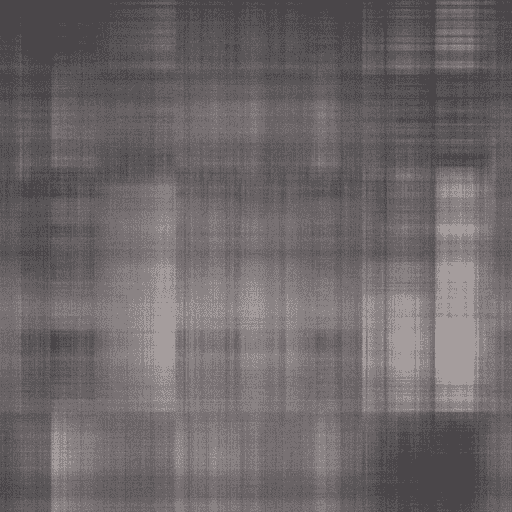} & 
\includegraphics[width=0.12\linewidth]{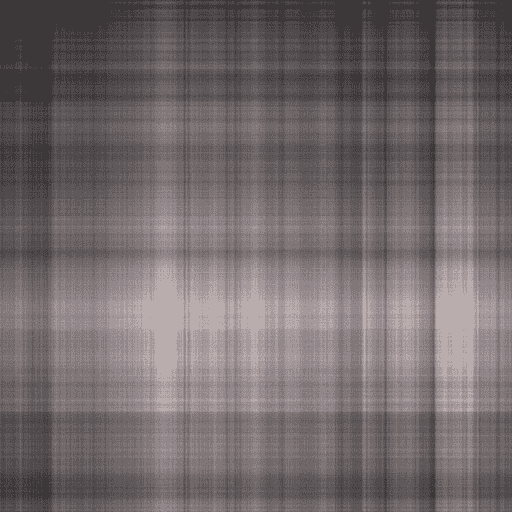}
\\
\multicolumn{1}{c}{\rotatebox{90}{\parbox{2cm}{\centering 1.0 Noise}}} &
\includegraphics[width=0.12\linewidth]{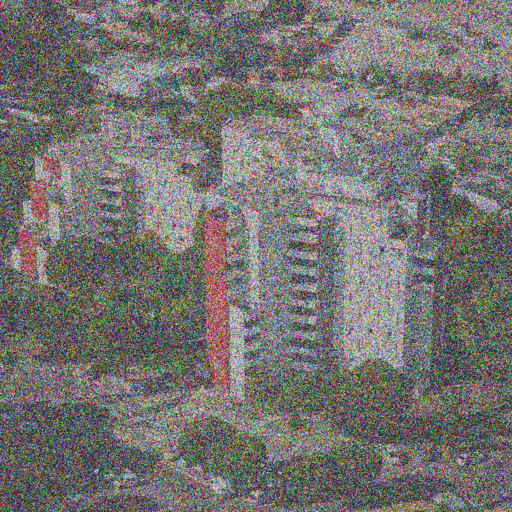} & 
\includegraphics[width=0.12\linewidth]{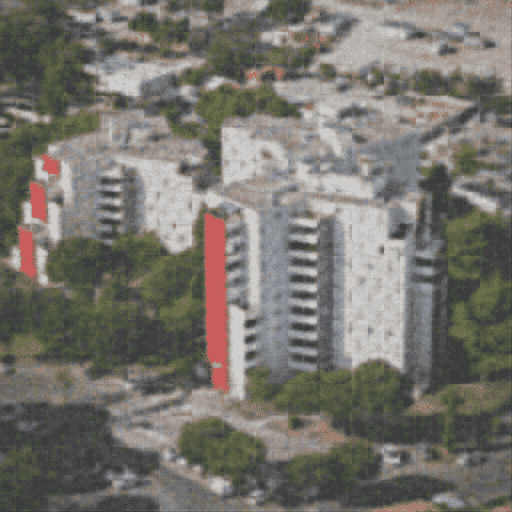} & 
\includegraphics[width=0.12\linewidth]{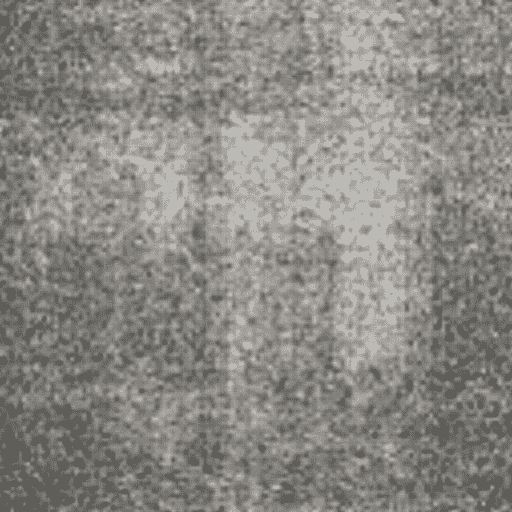} & 
\includegraphics[width=0.12\linewidth]{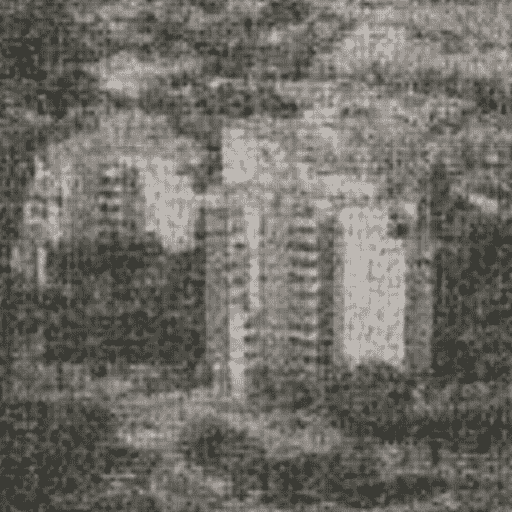} & 
\includegraphics[width=0.12\linewidth]{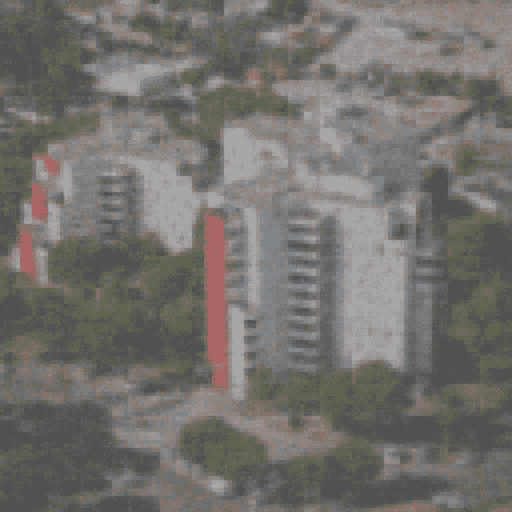} & 
\includegraphics[width=0.12\linewidth]{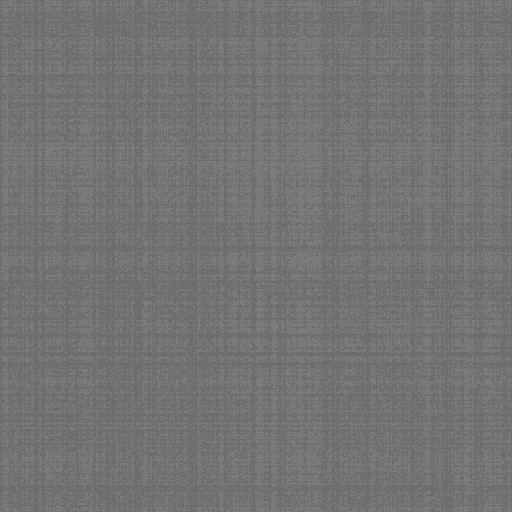} & 
\includegraphics[width=0.12\linewidth]{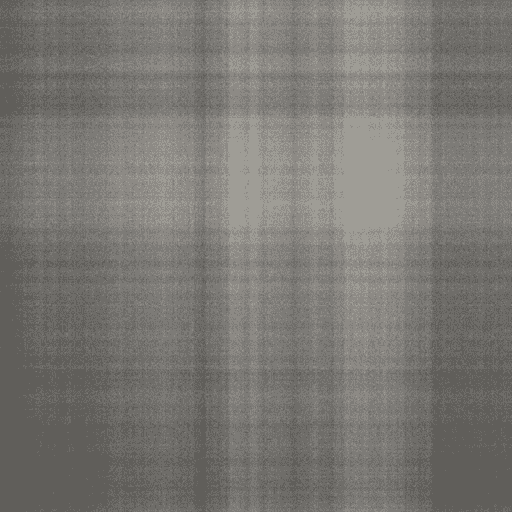} \\

\end{tabular}
\caption{Visual comparison of training with 
    noise levels $0.1$ (Top), $0.5$ (Middle),, and $1.0$ (Bottom). ``W/o up = without upsampling". }
    \label{fig:3_noise_levels_all_models}
    \vspace{-0.4cm}
\end{figure*}

\begin{figure*}
\begin{tabular}{cc}
\includegraphics[clip,width=0.48\linewidth]{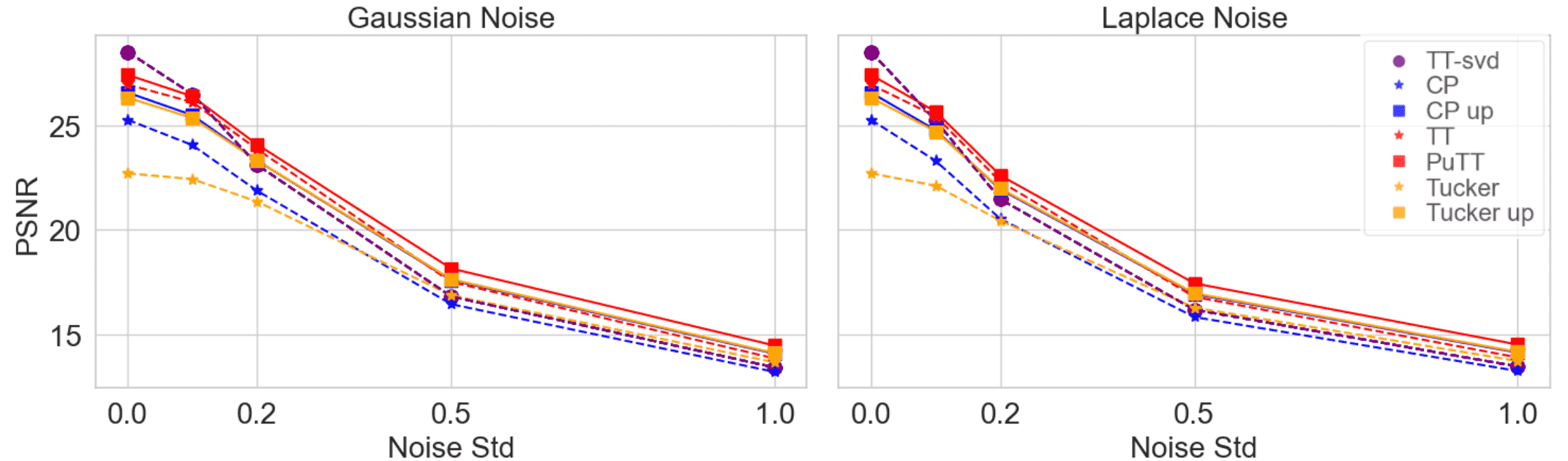} & 
\includegraphics[clip,width=0.48\linewidth]{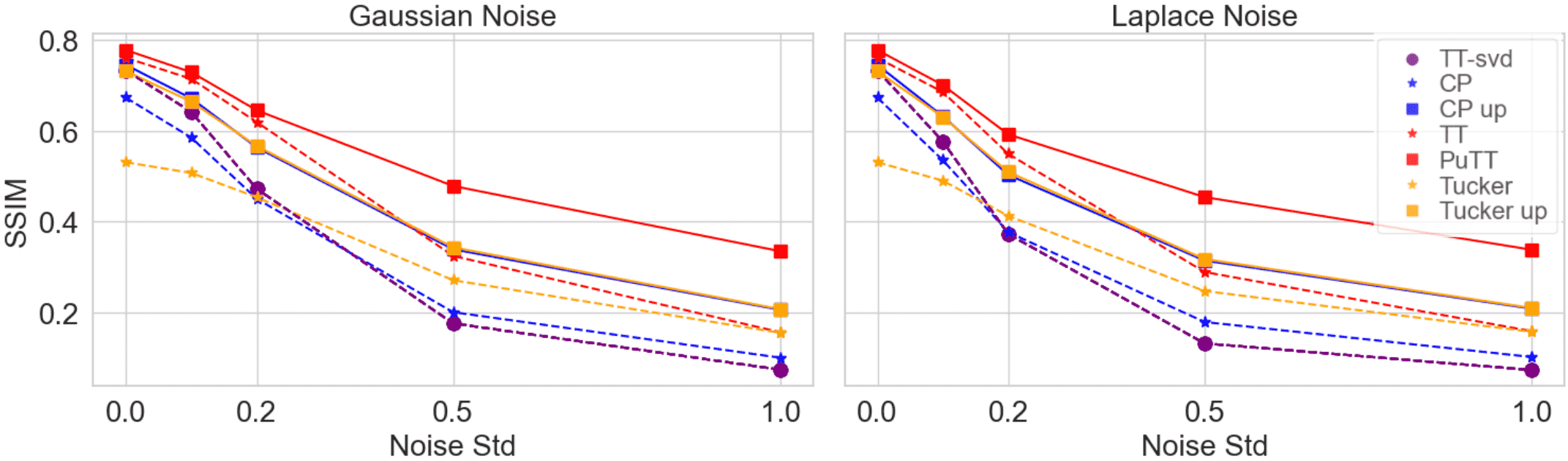} \\
(a) & (b) 
\vspace{-0.2cm}
\end{tabular}
\label{fig:noise_experiments_psnr_and_ssim_2d}
\caption{
PSNR (a) and SSIM (b) when varying the amount of Gaussian or Laplacian noise for PuTT and baselines. ``up = upsampling".}
\label{fig:noise_exps}
\vspace{-0.3cm}

\end{figure*}

We demonstrate the applicability of PuTT along three axes: (1). \textit{Compression}, showing that PuTT can be optimized to achieve the expressive power and compactness of tensor trains, (2). \textit{Learning from missing data}, showing that PuTT can learn to interpolate missing data effectively, (3). \textit{Denoising}, showing that PuTT can better recover the original signal given noisy inputs. 
We evaluate these axes with respect to the applications of 2D fitting, 3D fitting, and novel view synthesis. Lastly, we conduct an ablation study to underscore the impact of individual components. In the appendix and supplementary webpage, we provide additional results accompanying existing figures and discuss the limitations.

\noindent \textbf{Baseline Methods} \quad
\label{subsec:baselines_metrics}
We compare our method to state-of-the-art tensor-based methods:
CP, Tucker, and the VM decomposition~\cite{tensoRF} (see details in Sec.~\ref{sec:notation} and appendix). Furthermore, we include a comparison with TT-SVD~\cite{TT_decomp_oseledets} (adapted to QTT, see appendix for details). TT-SVD is a deterministic approach,
that starts from the full uncompressed tensor and compresses the tensor into tensor train form with a single pass over the input. TT-SVD 
and can only be applied when the direct signal (e.g., 2D image or 3D voxel grid) is provided, as it is applied analytically. 
It does not apply when optimization is required, such as in learning from missing data, or novel view synthesis, and does not perform well on noisy data. When referring to a baseline using upsampling, we train in a similar coarse-to-fine approach where the tensor factors are linearly interpolated to a finer grid using the approach developed by TensoRF~\cite{tensoRF}. The term ``TT" in our comparisons refers to training a QTT without adopting a coarse-to-fine learning approach. %

\noindent \textbf{Datasets} \quad
\label{subsec:datasets}
For 2D, we utilize two high-resolution images: ``Girl With a Pearl Earring''~\cite{Vermeer1665GirlPearlEarring} photograph, and ``Tokyo gigapixel"~\cite{Dobson2018Tokyo}, which are center-cropped to a 16k resolution. We also include three 4k images: ``Marseille''~\cite{marseille_img}, ``Pluto''~\cite{pluto_img}, and ``Westerlund''~\cite{westerlund_img} for noise and missing data experiments.
For 3D, we utilize the ``Flower'' data~\cite{ETH_FLOWER_3D_DATA_2023}, and 
John Hopkins Turbulence dataset~\cite{John_Hopkins_Tubelence_dataset}, which consists of a set of 3D voxel grids at $1024^3$ resolution, providing a diverse range of high-resolution structures by downsampling to different resolutions. 
For novel view synthesis, we employ the Blender~\cite{nerf} and NSVF~\cite{NSVF} datasets, comprising eight synthetic 3D scenes at $800 \times 800$ resolution, alongside the TanksTemples~\cite{tanks_and_temples} dataset ($1920\times 1080$).

\subsection{Compression}
\label{subsec:compression_quality}

\noindent \textbf{2D Compression} \quad
Fig.~\ref{fig:visual_comp} provides a visual comparison to baselines while Fig.~\ref{fig:2d_psnr_ssim} provides a numerical one, showcasing PSNR  results as a function of the compression ratio, defined as the ratio of uncompressed parameters to compressed parameters (Appendix shows corresponding SSIM results). 
Theoretically, QTTs have a logarithmic dependency on the image side length, while CP and Tucker have linear dependence. However, the tensor rank governs expressivity, and the tensor ranks of CP, Tucker, and QTT are not directly comparable; i.e. a QTT with max rank $50$ might be far more expressive than a CP with rank $50$. However, we clearly see the improved efficiency of QTT across different resolutions, which 
becomes more pronounced as the resolution increases across all compression ratios. At 16k resolution, PuTT significantly outperforms baselines, showing an advantage of over $2.5$ in PSNR and $0.1$ in SSIM.
PuTT performs better than the analytical TT-SVD in all 2D scenarios in terms of PSNR. For SSIM, PuTT consistently outperforms TT-SVD up to 8k resolution, with a minimum improvement of $0.05$ SSIM across all compression ratios. At 16k resolution, TT-SVD exhibits better SSIM, indicating the challenges in capturing structural image properties as the resolution and the ratio between batch size and image size increase. The CP, Tucker, and QTT ranks were adapted to match the target compression ratios shown in the figure. %

\setlength{\tabcolsep}{4pt}
\begin{table*}[t]
\centering
\resizebox{\textwidth}{!}{%
\begin{tabular}{l|cc|cc|cc|cc}
\toprule
 & & & \multicolumn{2}{c|}{Synthetic-NeRF} & \multicolumn{2}{c|}{NSVF} & \multicolumn{2}{c}{TanksTemples} \\
Method & Steps & Size(MB)$\downarrow$ & PSNR$\uparrow$ & SSIM$\uparrow$ & PSNR$\uparrow$ & SSIM$\uparrow$ & PSNR$\uparrow$ & SSIM$\uparrow$ \\
\midrule
    PlenOctrees*~\cite{PlenOctrees}-\textbf{L}  & 200k  & 1976.3 & 31.71 & 0.958 &   -     & -       &  27.99 & 0.917 \\
    Plenoxels*~\cite{plenoxels}-\textbf{L}      & 128k  & 778.1 & 31.71 & 0.958 &   -     & -       &  27.43 & 0.906 \\
    DVGO*~\cite{DVGO}-\textbf{L}             & 30k   & 612.1 & 31.95 & 0.957 &   35.08 & 0.975  &  28.41 & 0.911 \\
    K-planes*~\cite{fridovichkeil2023kplanes}-\textbf{L}             & 30k   & 122.1 & 32.36 & 0.962 &   - & -  &  - & - \\
    Instant-NGP*~\cite{instantNerf}\textbf-{L}                             & 35k   &   45.5 & \textbf{33.18} & 0.963  & -  & - & -  &  \\
    TensoRF~\cite{tensoRF} VM-192*-\textbf{L}                          & 30k  & 71.8   & 33.14 & \textbf{0.963} &   36.52 & 0.982   &  \textbf{28.56} & \textbf{0.920} \\
    Our PuTT-600-\textbf{L}                       & 80k  & 60.3   & 32.79 & 0.958 & \textbf{36.57} & \textbf{0.982}   &  28.37 & 0.917 \\
    \midrule
    NeRF*~\cite{nerf}-\textbf{S}             & 300k  &  5.0    & 31.01 & 0.947 &   30.81 & 0.952   &  25.78 & 0.864 \\
    TT-NF * \cite{ttnf}-\textbf{S}           & 80k  & 7.9  & 31.09  & 0.945 &   - & -   &  - & - \\
    TensoRF CP-384*-\textbf{S}                            & 30k  & 3.9    & 31.56 & 0.949 &   34.48 & 0.971   &  27.59 & 0.897 \\
    TensoRF VM-30 (no shrinking)-\textbf{S}                & 80k  & 6.9   & 31.10  & 0.948 &   34.57 & 0.974   &  27.67 & 0.898 \\
    Our PuTT-200-\textbf{S}                       & 80k  & 6.7   & \textbf{31.66} & \textbf{0.956} & \textbf{35.58} & \textbf{0.976}   &  \textbf{27.99} & \textbf{0.901} \\
    \midrule
    TensoRF VM-48 (no shrinking)-\textbf{M}                & 80k  & 12.0   & 31.62 & 0.952 &   35.44	 & 0.976   &  28.03 & 0.901 \\
    Our PuTT-280-\textbf{M}                       & 80k  & 12.0  & \textbf{31.95} & \textbf{0.957}  & \textbf{36.04} & \textbf{0.977}   &  \textbf{28.15} & \textbf{0.905} \\
    \bottomrule
\end{tabular}
} 
\caption{Comparison to baselines for the task of novel view synthesis for the Synthetic-NeRF~\cite{nerf}, NSVF~\cite{NSVF}, and TanksTemples~\cite{tanks_and_temples} datasets. Scores of the baseline methods with a $*$ are taken directly from corresponding papers whenever possible. \textbf{L}, \textbf{M} and \textbf{S} indicate a Large ($>60$ MB), Medium ($12$ MB) or small ($<8$ MB) model size. We compare our method on Medium and Small-sized models but also include Large-sized models for reference. } \vspace{-2mm}
\label{table:results_nerf}  
\end{table*}

\begin{figure*}%

\centering
\begin{tabular}{cc}
\includegraphics[trim={0 0 0cm 0},clip,width=0.48\linewidth,height=3cm]{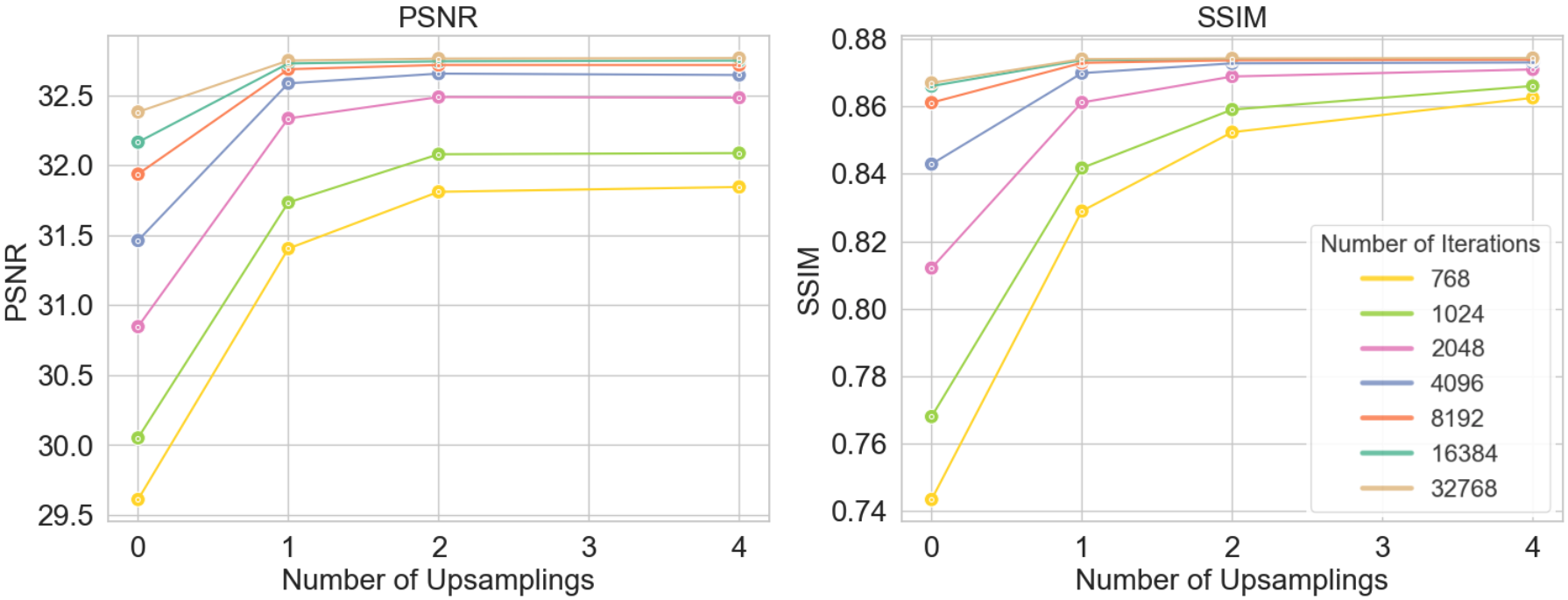} & 
\includegraphics[trim={0 0 0cm 0},clip,width=0.48\linewidth,height=3cm]{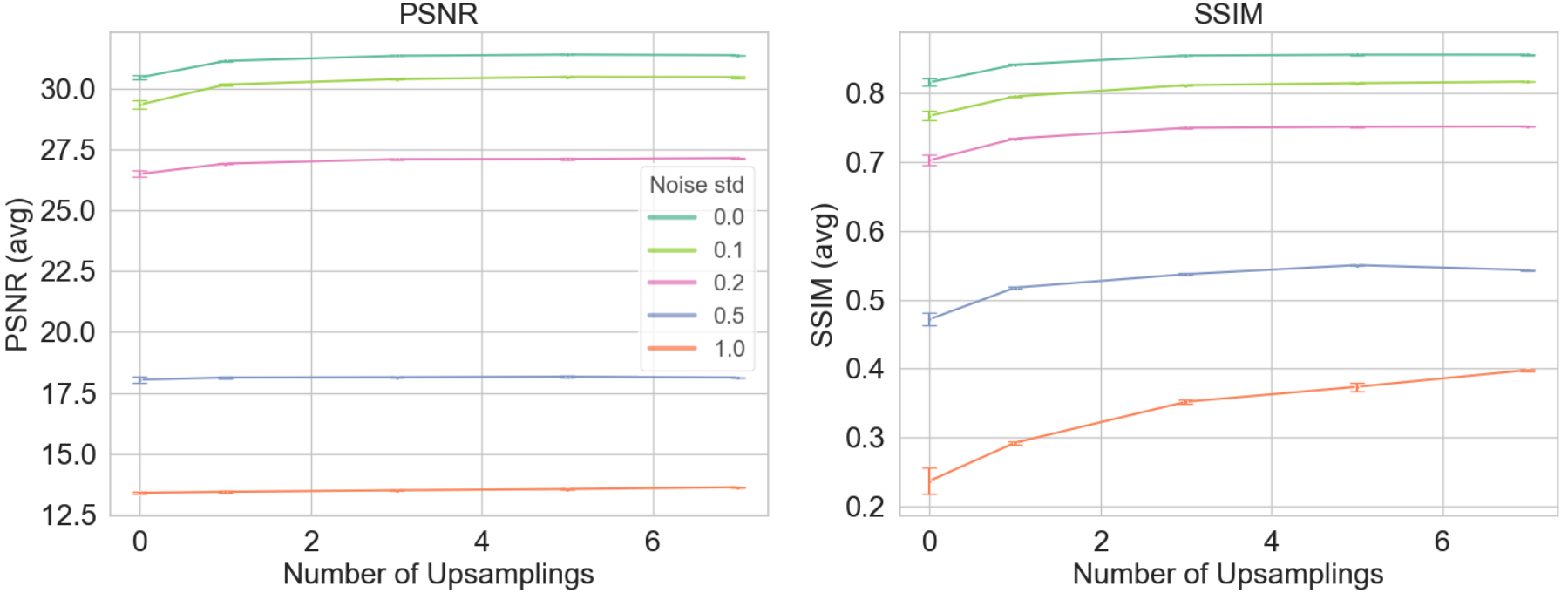} \\ 
(a) & (b) 
\vspace{-0.2cm}
\end{tabular}
\caption{PSNR and SSIM for PuTT, under varying iteration counts (a) and 
varying levels of Gaussian noise ($\sigma$) (b).}
\label{fig:num_upsamplings_vs_iterations_a_b}
\vspace{-0.4cm}
\end{figure*}

\begin{figure}
    \centering
    \includegraphics[width=\linewidth]{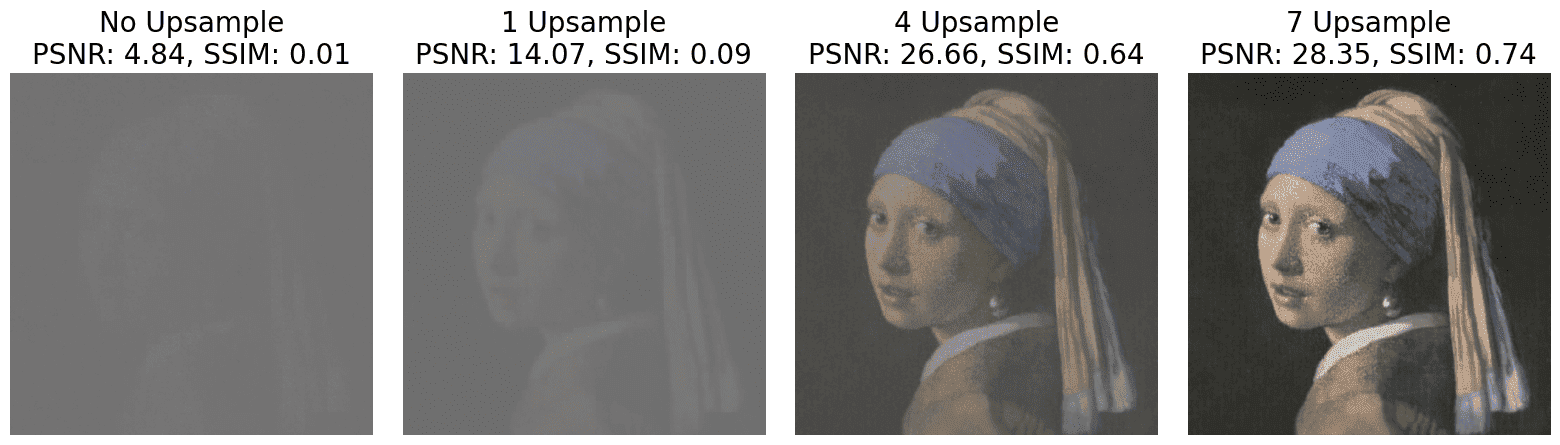}
    \caption{Varying the number of upsampling steps when training $99\%$ missing data. The first image displays the training data, followed by PuTT results with $0, 1, 4$, and $7$ upsampling steps. 
    }
    \label{fig:ablation_visual_upsampling_incomplete_data}
    \vspace{-0.4cm}
\end{figure}

\begin{figure}[tp]
    \centering
    \includegraphics[width=\linewidth]{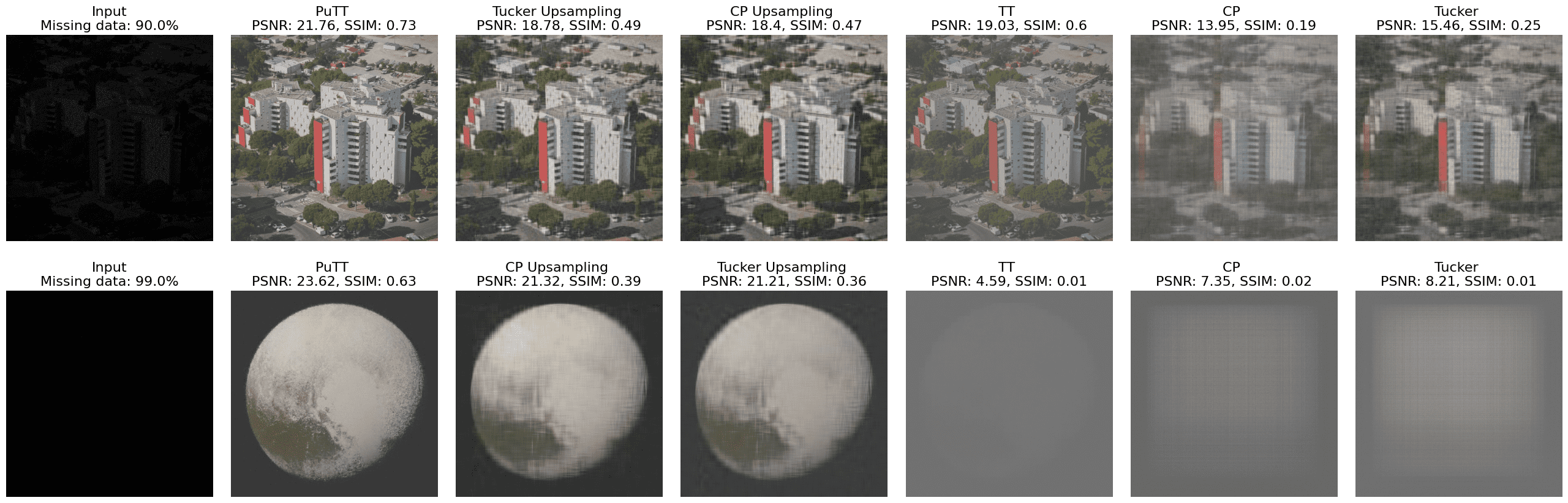} 
    \caption{Visual comparison of PuTT to baselines when training 
    with $90\%$ (Top) and $99\%$ (Bottom) missing data. 
    }
    \label{fig:visual_missing}
    \vspace{-0.6cm}
\end{figure}

\begin{table}[ht]
\centering

\small 
\renewcommand{\arraystretch}{1}
\begin{tabular}{c@{~~~}c@{~~~}c@{~~~}c@{~~~}c@{~~~}c@{~~~}c}
\toprule
  & \multicolumn{2}{c}{Chair} & \multicolumn{2}{c}{Ficus} & \multicolumn{2}{c}{All Scenes} \\
  \midrule
 & TensoRF      & PuTT       & TensoRF      & PuTT       & TensoRF          & PuTT          \\
\midrule
Near   & 35.59        & \textbf{37.12}      & 32.53        & \textbf{33.95}      & 31.93            & \textbf{32.93}         \\
Far    & 31.74        & \textbf{32.71}      & 30.16        & \textbf{31.06}      & 28.97            & \textbf{29.41}         \\
All    & 32.22        & \textbf{33.50}      & 30.71        & \textbf{31.69}      & 30.43            & \textbf{31.66}         \\
\bottomrule
\end{tabular}
\caption{Novel view synthesis comparison to TensoRF (PSNR) on the Chair, Ficus, and all scenes (Blender dataset~\cite{nerf}). We test on views near training (Near), far from training (Far), and all test views (All). See the appendix for all details.}
\vspace{-0.3cm}
\label{tab:nerf_NearFar}
\end{table}

\begin{figure}[h]
\vspace{-0.2cm}
\centering
\includegraphics[width = 0.9\linewidth]{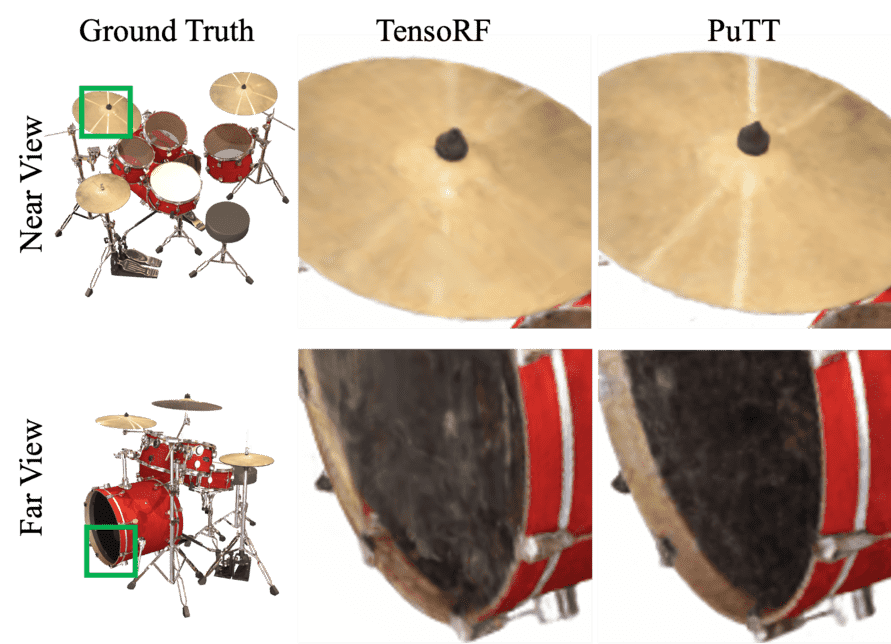} 
\vspace{-0.4cm}
\caption{Visual comparison of a single NeRF for ``drums". 
}
\label{fig:drums-visual}
\vspace{-0.4cm}
\end{figure}

\noindent \textbf{3D Compression} \quad
In 3D, we observe similar trends, as shown in the Fig.~\ref{fig:3d_psnr_ssim} (PSNR) and appendix (SSIM). 
However, the dynamics differ notably among the PuTT, CP, and Tucker at various resolutions.
Up to $256^3$, CP and Tucker perform closely to PuTT. Yet, for $1024^3$, PuTT significantly outshines the other methods, leading to more than $1$ PSNR improvement and nearly $0.005$ improvement in SSIM. VM's performance is less substantial due to its quadratic dependency on the side length, unlike the linear dependency of CP and Tucker and the logarithmic dependency of QTT.
VM shows relatively poor performance. This is attributable to the high compression setting of our experiment, coupled with VM's quadratic dependency on side length. 
PuTT is comparable to deterministic TT-SVD, but achieving high SSIM is exacerbated at larger resolutions, underscoring the increased difficulty in maintaining image quality at higher compression settings. 
One might speculate that the crossover in efficiency, where it pays off to use the QTT representation, occurs around sidelength $L=512$.

\noindent \textbf{Novel View Synthesis} \quad
As seen in Tab.~\ref{table:results_nerf}, PuTT outperforms TensoRF when using high compressions (7MB and 12MB) for Blender, NSVF, and TanksTemples datasets and has comparable performance on the large model size.
We evaluate models in three sizes: Large ($>$60 MB), Medium (12 MB), and Small ($<$8 MB). Our 12MB PuTT model, despite having six times fewer parameters, matches the performance of the Large baselines on the NSVF and TanksTemples datasets. Moreover, our 7MB PuTT model outperforms Small baselines by over one PSNR on the NSVF dataset. For a detailed overview of the implementation, please consult the appendix. 
In Tab.~\ref{tab:nerf_NearFar}, for the 7MB size, we consider a more detailed comparison, distinguishing between near and far views at test time. Our improvement is more pronounced at nearby views, indicating our ability to capture fine details better than TensoRF. Fig.~\ref{fig:drums-visual}
demonstrates visually our ability to better capture fine-grained details. In Appendix \ref{app:Putt_nerf} we provide per-scene results.%

\subsection{Learning from Incomplete Data}
\label{subsec:subset_sampling}

We evaluate the capability of PuTT to learn 2D or 3D representations by training only on a subset of the available data, thus assessing its generalizability. We randomly select a percentage $p\%$ of the full input data $I_D$. This yields a subset of indices, $\mathcal{I}_{D,p}$, which we use exclusively for training. We evaluate the performance of the full input data, including untrained indices. 
To train with a down-sampled target at lower resolutions, we create a modified target image $I_{D-l,p}$. We build $I_{D-1,p}$ from $I_{D,p}$,
we use a custom masked average pooling method, averaging only the non-zero values in each window, sized according to the downsampling factor. 
The resulting downsampled image, $I_{D-l,p}$, thus contains aggregated information within each patch of size $2^l \times 2^l$ of $I_{D,p}$. Any non-zero value in $I_{D-l,p}$ represents aggregated data from the original data within the corresponding patch. This forms a new set of indices $\mathcal{I}_{l,p}$, from which we sample during training on the downsampled input $I_{D-l,p}$. 
Fig.~\ref{fig:subsampling_num_upsamplings_and_subset}(a) considers the effect of varying the number of upsampling steps,
when training with different training data percentages (see visual examples in Fig.~\ref{fig:ablation_visual_upsampling_incomplete_data}). 
It illustrates the significance of upsampling when learning from limited data samples, highlighting its growing importance as training data size decreases. 
Without upsampling, training on just $1\%$ of the input, PuTT gets a PSNR of $1.87$ and SSIM of $0.0018$. However, applying seven upsampling steps enhances the results to a PSNR of $28.73$ and SSIM of $0.7349$. Similar improvements are observed with $10\%$ training data. 
Fig.~\ref{fig:subsampling_num_upsamplings_and_subset}(b) shows PuTT consistently surpasses CP and Tucker, especially at lower resolutions. Fig.~\ref{fig:visual_missing}
gives a visual example when $90\%$ and $99\%$ of the data is missing for PuTT and baselines, showing a better recovery of the input image.

\subsection{Noise Removal}
\label{subsec:noise_removal}

We aim to learn a tensor representation of a target, denoted as $\mathcal{X}_I$, using samples from its noisy counterpart, $\hat{\mathcal{X}}_I$, obtained by 
adding noise $Z$. In our experiments, $Z$ is sampled from either a Normal or Laplacian distribution (see details in appendix). 
At training, we minimize 
$||\hat{\mathcal{X}}_I-R_{T}||^2_2$. At inference, we compare the learned $R_{T}$ with the original clean target $\mathcal{X}_I$. 
As shown in Fig.~\ref{fig:noise_exps}(a), PuTT consistently achieves higher PSNR than other tensor network methods across various levels of noise sigma. Notably, the application of upsampling not only avoids overfitting to noisy samples but also consistently outperforms the non-upsampling approach. Fig.~\ref{fig:noise_exps}(b) underscores the clear benefits of employing upsampling strategies. This is particularly evident in PuTT, where upsampling yields an SSIM score improvement of more than $0.1$ for noise levels exceeding $0.2 \sigma$.

Fig.~\ref{fig:3_noise_levels_all_models} 
shows recovered images, both with and without upsampling at Gaussian noise levels of $\sigma = 0.05, 0.5, 1.0$. These examples highlight the efficacy of upsampling in mitigating noise, demonstrating visibly superior results compared to models not utilizing upsampling. Moreover, PuTT showcases enhanced details and a more effective noise-filtering capability because of its hierarchical structure.

\subsection{Ablation Study}

Fig.~\ref{fig:num_upsamplings_vs_iterations_a_b}(a) illustrates the impact of varying the number of upsamplings and the number of iterations on efficiency.
The results highlight the efficiency of our method: using just $1024$ iterations with four upsampling steps, we achieve comparable PSNR and SSIM to training without upsampling for $16$k iterations. 
Beyond $4$k iterations, there is no significant improvement in quality.
Without upsampling this plateau is not reached even after $32$k iterations.
More upsampling steps notably enhance SSIM because SSIM is sensitive to global features like overall mean, standard deviation, and luminance. Training on downsampled inputs using tensor trains, which involves gradient updates based on batches covering larger portions of the input, seems to facilitate learning these global characteristics more effectively. 

Fig.~\ref{fig:num_upsamplings_vs_iterations_a_b}(b) 
presents an ablation study on the "Girl with a Pearl Earring" image at 4k resolution and rank $200$, with a fixed iteration count of $8192$. We examine the impact of varying the number of upsampling steps on the model's performance, for varying degrees of Gaussian noise applied on the input. 
The LHS of Fig.~\ref{fig:num_upsamplings_vs_iterations_a_b}(b) illustrates the influence of the first upsampling step on PSNR, showing a notable improvement. Subsequent upsampling steps result in incremental gains. 
The SSIM plot, on the RHS, reveals a more pronounced improvement with each upsampling step, indicating the effectiveness of upsampling in enhancing the model's ability to learn global features. 

These findings are supported by additional experiments and analyses in the appendix. Sec.~\ref{appendix:seed_variability} highlights the statistical robustness of our experiments, showing minimal variation with upsampling compared to without. Sec.~\ref{app:init_influence} examines the influence of initialization, demonstrating PuTT's consistent performance across different initialization values. Without upsampling, QTT training is sensitive to initialization standard deviation, with PSNR fluctuating between $34.958$ (std $0.05$) and $12.320$ (std $0.5$) for a $512 \times 512$ RGB image. In contrast, PuTT's PSNR remains stable between $36.100$ and $36.116$ for std values from $0.001$ to $0.5$. Sec.~\ref{app:slimmerf} shows how rank incrementation strategy inspired by SlimmeRF~\cite{yuan2023slimmerf} can further improve performance.

\section{Conclusion}
\label{sec:conclusion}

We proposed a novel coarse-to-fine, Tensor-Train-based representation called `Prolongation Upsampling Tensor Train (PuTT)' and an associated optimization strategy. This approach involves prolonging or “upsampling” a learned Tensor Train representation, creating a sequence of tensor trains that are incrementally refined in a coarse-to-fine manner.
We demonstrated the applicability of our method along the axes of compression, learning from missing data, and denoising. 
We evaluated these axes with respect to the applications of 2D fitting, 3D fitting, and novel view synthesis and demonstrated state-of-the-art performance compared to other tensor-based methods, especially in high-compression settings. 
Our learning scheme leverages the improved compression ability and multi-resolution nature of the Quantized Tensor Train (QTT) for improved denoising and image completion capabilities. In future work, we hope to apply PuTT to large-scale Neural Radiance Fields (NeRFs) and dynamic neural fields, utilizing the logarithmic dimensionality advantages of QTTs to represent large and finely detailed scenes.

\noindent \textbf{Acknowledgements} \quad 
Sebastian Loeschcke is supported by the Danish Data Science Academy, which is funded by the Novo Nordisk Foundation (NNF21SA0069429) and VILLUM FONDEN (40516). Serge Belongie and Dan Wang are supported by the Pioneer Centre for AI, DNRF grant number P1. MJK acknowledges support from the Carlsberg Foundation and the Novo Nordisk Foundation.

\noindent \textbf{Impact Statement} \quad This paper contributes to the Machine Learning field by enhancing visual representation learning using tensor train optimization. Given the current stage of our research, we identify no specific ethical issues warranting exploration. Nonetheless, we acknowledge the dynamic nature of technological impacts and pledge to continually assess the ethical implications as our research advances.

\bibliography{11_references}

\begin{thebibliography}{59}
\providecommand{\natexlab}[1]{#1}
\providecommand{\url}[1]{\texttt{#1}}
\expandafter\ifx\csname urlstyle\endcsname\relax
  \providecommand{\doi}[1]{doi: #1}\else
  \providecommand{\doi}{doi: \begingroup \urlstyle{rm}\Url}\fi

\bibitem[Adelson et~al.(1984)Adelson, Anderson, Bergen, Burt, and Ogden]{adelson1984pyramid}
Adelson, E.~H., Anderson, C.~H., Bergen, J.~R., Burt, P.~J., and Ogden, J.~M.
\newblock Pyramid methods in image processing.
\newblock \emph{RCA engineer}, 29\penalty0 (6):\penalty0 33--41, 1984.

\bibitem[Antonini et~al.(1990)Antonini, Barlaud, Mathieu, and Daubechies]{wavelets_and_quantization_vision_data}
Antonini, M., Barlaud, M., Mathieu, P., and Daubechies, I.
\newblock Image coding using vector quantization in the wavelet transform domain.
\newblock In \emph{International Conference on Acoustics, Speech, and Signal Processing}, pp.\  2297--2300 vol.4, 1990.
\newblock \doi{10.1109/ICASSP.1990.116036}.

\bibitem[Bramble(2019)]{multigrid}
Bramble, J.~H.
\newblock \emph{Multigrid methods}.
\newblock Chapman and Hall/CRC, 2019.

\bibitem[Carroll \& Chang(1970)Carroll and Chang]{CP1}
Carroll, J.~D. and Chang, J.~J.
\newblock Analysis of individual differences in multidimensional scaling via an n-way generalization of “eckart-young” decomposition.
\newblock \emph{Psychometrika}, 35:\penalty0 283--319, 1970.
\newblock URL \url{https://api.semanticscholar.org/CorpusID:50364581}.

\bibitem[Chan et~al.(2021)Chan, Monteiro, Kellnhofer, Wu, and Wetzstein]{chan2021pi}
Chan, E.~R., Monteiro, M., Kellnhofer, P., Wu, J., and Wetzstein, G.
\newblock pi-gan: Periodic implicit generative adversarial networks for 3d-aware image synthesis.
\newblock In \emph{Proceedings of the IEEE/CVF conference on computer vision and pattern recognition}, pp.\  5799--5809, 2021.

\bibitem[Chan et~al.(2022)Chan, Lin, Chan, Nagano, Pan, De~Mello, Gallo, Guibas, Tremblay, Khamis, et~al.]{chan2022efficient}
Chan, E.~R., Lin, C.~Z., Chan, M.~A., Nagano, K., Pan, B., De~Mello, S., Gallo, O., Guibas, L.~J., Tremblay, J., Khamis, S., et~al.
\newblock Efficient geometry-aware 3d generative adversarial networks.
\newblock In \emph{Proceedings of the IEEE/CVF Conference on Computer Vision and Pattern Recognition}, pp.\  16123--16133, 2022.

\bibitem[Chen et~al.(2022)Chen, Xu, Geiger, Yu, and Su]{tensoRF}
Chen, A., Xu, Z., Geiger, A., Yu, J., and Su, H.
\newblock Tensorf: Tensorial radiance fields, 2022.
\newblock URL \url{https://arxiv.org/abs/2203.09517}.

\bibitem[Crowley \& Stern(1984)Crowley and Stern]{crowley1984fast}
Crowley, J.~L. and Stern, R.~M.
\newblock Fast computation of the difference of low-pass transform.
\newblock \emph{IEEE transactions on pattern analysis and machine intelligence}, 1\penalty0 (2):\penalty0 212--222, 1984.

\bibitem[Dobson(2018)]{Dobson2018Tokyo}
Dobson, T.
\newblock 1.2 gigapixel panorama of shibuya in tokyo, japan.
\newblock \url{https://www.flickr.com/photos/trevor_dobson_inefekt69/29314390837}, 2018.
\newblock Accessed: 17-11-2023.

\bibitem[Fridovich-Keil et~al.(2023)Fridovich-Keil, Meanti, Warburg, Recht, and Kanazawa]{fridovichkeil2023kplanes}
Fridovich-Keil, S., Meanti, G., Warburg, F., Recht, B., and Kanazawa, A.
\newblock K-planes: Explicit radiance fields in space, time, and appearance, 2023.

\bibitem[Harshman(1970)]{CP2}
Harshman, R.~A.
\newblock {F}oundations of the {P}{A}{R}{A}{F}{A}{C} procedure: {M}odels and conditions for an "explanatory" multi-modal factor analysis.
\newblock \emph{UCLA Working Papers in Phonetics}, 16:\penalty0 1--84, 1970.

\bibitem[Holtz et~al.(2012)Holtz, Rohwedder, and Schneider]{ALS}
Holtz, S., Rohwedder, T., and Schneider, R.
\newblock The alternating linear scheme for tensor optimization in the tensor train format.
\newblock \emph{SIAM Journal on Scientific Computing}, 34\penalty0 (2):\penalty0 A683--A713, 2012.

\bibitem[Hubig et~al.(2017)Hubig, McCulloch, and Schollwoeck]{MPO_link1}
Hubig, C., McCulloch, I.~P., and Schollwoeck, U.
\newblock Generic construction of efficient matrix product operators.
\newblock \emph{Phys. Rev. B}, 95:\penalty0 035129, Jan 2017.
\newblock \doi{10.1103/PhysRevB.95.035129}.
\newblock URL \url{https://link.aps.org/doi/10.1103/PhysRevB.95.035129}.

\bibitem[Khoromskij(2011)]{QTT_paper}
Khoromskij, B.~N.
\newblock $o(d\log n)$-quantics approximation of n-d tensors in high-dimensional numerical modeling.
\newblock \emph{Constructive Approximation}, 34:\penalty0 257--280, 2011.

\bibitem[Kingma \& Ba(2014)Kingma and Ba]{kingma2014adam}
Kingma, D.~P. and Ba, J.
\newblock Adam: A method for stochastic optimization.
\newblock \emph{arXiv preprint arXiv:1412.6980}, 2014.

\bibitem[Kl{\"u}mper et~al.(1993)Kl{\"u}mper, Schadschneider, and Zittartz]{MPSold}
Kl{\"u}mper, A., Schadschneider, A., and Zittartz, J.
\newblock Matrix product ground states for one-dimensional spin-1 quantum antiferromagnets.
\newblock \emph{Europhysics Letters}, 24\penalty0 (4):\penalty0 293, 1993.

\bibitem[Knapitsch et~al.(2017)Knapitsch, Park, Zhou, and Koltun]{tanks_and_temples}
Knapitsch, A., Park, J., Zhou, Q.-Y., and Koltun, V.
\newblock Tanks and temples: benchmarking large-scale scene reconstruction.
\newblock \emph{ACM Trans. Graph.}, 36\penalty0 (4), jul 2017.
\newblock ISSN 0730-0301.
\newblock \doi{10.1145/3072959.3073599}.
\newblock URL \url{https://doi.org/10.1145/3072959.3073599}.

\bibitem[Kolda \& Bader(2009)Kolda and Bader]{tensordecompositions}
Kolda, T.~G. and Bader, B.~W.
\newblock Tensor decompositions and applications.
\newblock \emph{SIAM review}, 51\penalty0 (3):\penalty0 455--500, 2009.

\bibitem[Li et~al.(2008)Li, Perlman, Wan, Yang, Meneveau, Burns, Chen, Szalay, and Eyink]{John_Hopkins_Tubelence_dataset}
Li, Y., Perlman, E., Wan, M., Yang, Y., Meneveau, C., Burns, R., Chen, S., Szalay, A., and Eyink, G.
\newblock A public turbulence database cluster and applications to study lagrangian evolution of velocity increments in turbulence.
\newblock \emph{Journal of Turbulence}, 9:\penalty0 N31, jan 2008.
\newblock \doi{10.1080/14685240802376389}.
\newblock URL \url{https://doi.org/10.1080%2F14685240802376389}.

\bibitem[Lindell et~al.(2022)Lindell, Van~Veen, Park, and Wetzstein]{lindell2022bacon}
Lindell, D.~B., Van~Veen, D., Park, J.~J., and Wetzstein, G.
\newblock Bacon: Band-limited coordinate networks for multiscale scene representation.
\newblock In \emph{Proceedings of the IEEE/CVF conference on computer vision and pattern recognition}, pp.\  16252--16262, 2022.

\bibitem[Liu et~al.(2020{\natexlab{a}})Liu, Gu, Lin, Chua, and Theobalt]{NSVF}
Liu, L., Gu, J., Lin, K.~Z., Chua, T.-S., and Theobalt, C.
\newblock Neural sparse voxel fields.
\newblock \emph{NeurIPS}, 2020{\natexlab{a}}.

\bibitem[Liu et~al.(2020{\natexlab{b}})Liu, Gu, Lin, Chua, and Theobalt]{strivec}
Liu, L., Gu, J., Lin, K.~Z., Chua, T.-S., and Theobalt, C.
\newblock Neural sparse voxel fields.
\newblock \emph{NeurIPS}, 2020{\natexlab{b}}.

\bibitem[Lombardi et~al.(2019)Lombardi, Simon, Saragih, Schwartz, Lehrmann, and Sheikh]{lombardi2019neural}
Lombardi, S., Simon, T., Saragih, J., Schwartz, G., Lehrmann, A., and Sheikh, Y.
\newblock Neural volumes: Learning dynamic renderable volumes from images.
\newblock \emph{arXiv preprint arXiv:1906.07751}, 2019.

\bibitem[Lubasch et~al.(2018)Lubasch, Moinier, and Jaksch]{TT-Prolongation}
Lubasch, M., Moinier, P., and Jaksch, D.
\newblock Multigrid renormalization.
\newblock \emph{Journal of Computational Physics}, 372:\penalty0 587--602, nov 2018.
\newblock \doi{10.1016/j.jcp.2018.06.065}.
\newblock URL \url{https://doi.org/10.1016\%2Fj.jcp.2018.06.065}.

\bibitem[Mallat(1989)]{mallat1989theory}
Mallat, S.~G.
\newblock A theory for multiresolution signal decomposition: the wavelet representation.
\newblock \emph{IEEE transactions on pattern analysis and machine intelligence}, 11\penalty0 (7):\penalty0 674--693, 1989.

\bibitem[McCulloch(2008)]{MPO_link2}
McCulloch, I.
\newblock Infinite size density matrix renormalization group, revisited.
\newblock \emph{arxiv:0804.2509}, 2008.
\newblock URL \url{https://arxiv.org/abs/0804.2509}.

\bibitem[Mildenhall et~al.(2020)Mildenhall, Srinivasan, Tancik, Barron, Ramamoorthi, and Ng]{nerf}
Mildenhall, B., Srinivasan, P.~P., Tancik, M., Barron, J.~T., Ramamoorthi, R., and Ng, R.
\newblock Nerf: Representing scenes as neural radiance fields for view synthesis.
\newblock \emph{CoRR}, abs/2003.08934, 2020.
\newblock URL \url{https://arxiv.org/abs/2003.08934}.

\bibitem[M\"uller et~al.(2022)M\"uller, Evans, Schied, and Keller]{instantNerf}
M\"uller, T., Evans, A., Schied, C., and Keller, A.
\newblock Instant neural graphics primitives with a multiresolution hash encoding.
\newblock \emph{ACM Trans. Graph.}, 41\penalty0 (4):\penalty0 102:1--102:15, July 2022.
\newblock \doi{10.1145/3528223.3530127}.
\newblock URL \url{https://doi.org/10.1145/3528223.3530127}.

\bibitem[{NASA and ESA}(2023)]{westerlund_img}
{NASA and ESA}.
\newblock Westerlund 2 - nasa and esa, a. nota (esa/stsci), and the westerlund 2 science team.
\newblock \url{https://hubblesite.org/contents/news-releases/2020/news-2020-15}, 2023.
\newblock Accessed: 17-11-2023.

\bibitem[{NASA/Johns Hopkins University}(2023)]{pluto_img}
{NASA/Johns Hopkins University}.
\newblock True colors of pluto - from applied physics laboratory/southwest research institute/alex parker.
\newblock \url{https://science.nasa.gov/resource/true-colors-of-pluto/?category=planets/dwarf-planets_pluto}, 2023.
\newblock Accessed: 17-11-2023.

\bibitem[Niemeyer \& Geiger(2021)Niemeyer and Geiger]{niemeyer2021giraffe}
Niemeyer, M. and Geiger, A.
\newblock Giraffe: Representing scenes as compositional generative neural feature fields.
\newblock In \emph{Proceedings of the IEEE/CVF Conference on Computer Vision and Pattern Recognition}, pp.\  11453--11464, 2021.

\bibitem[Novikov et~al.(2021)Novikov, Panov, and Oseledets]{TTdensity}
Novikov, G.~S., Panov, M.~E., and Oseledets, I.~V.
\newblock Tensor-train density estimation.
\newblock In \emph{Uncertainty in artificial intelligence}, pp.\  1321--1331. PMLR, 2021.

\bibitem[Obukhov et~al.(2022)Obukhov, Usvyatsov, Sakaridis, Schindler, and Van~Gool]{ttnf}
Obukhov, A., Usvyatsov, M., Sakaridis, C., Schindler, K., and Van~Gool, L.
\newblock Tt-nf: Tensor train neural fields, 2022.
\newblock URL \url{https://arxiv.org/abs/2209.15529}.

\bibitem[Oechsle et~al.(2021)Oechsle, Peng, and Geiger]{oechsle2021unisurf}
Oechsle, M., Peng, S., and Geiger, A.
\newblock Unisurf: Unifying neural implicit surfaces and radiance fields for multi-view reconstruction.
\newblock In \emph{Proceedings of the IEEE/CVF International Conference on Computer Vision}, pp.\  5589--5599, 2021.

\bibitem[of~Zurich(2023)]{ETH_FLOWER_3D_DATA_2023}
of~Zurich, U.
\newblock Computer-assisted paleoanthropology group and visualization and multimedia lab.
\newblock University of Zurich, 2023.
\newblock URL \url{https://www.ifi.uzh.ch/en/vmml/research/datasets.html}.
\newblock We acknowledge the Computer-Assisted Paleoanthropology group and the Visualization and MultiMedia Lab at University of Zurich (UZH) for the acquisition of the CT datasets.

\bibitem[Or{\'u}s(2019)]{orusTN}
Or{\'u}s, R.
\newblock Tensor networks for complex quantum systems.
\newblock \emph{Nature Reviews Physics}, 1\penalty0 (9):\penalty0 538--550, 2019.

\bibitem[Orús(2014)]{PEPS}
Orús, R.
\newblock A practical introduction to tensor networks: Matrix product states and projected entangled pair states.
\newblock \emph{Annals of Physics}, 349:\penalty0 117--158, 2014.
\newblock ISSN 0003-4916.
\newblock \doi{https://doi.org/10.1016/j.aop.2014.06.013}.
\newblock URL \url{https://www.sciencedirect.com/science/article/pii/S0003491614001596}.

\bibitem[Oseledets(2009)]{qtt}
Oseledets, I.
\newblock Approximation of matrices with logarithmic number of parameters.
\newblock \emph{Doklady Mathematics}, 80:\penalty0 653--654, 10 2009.
\newblock \doi{10.1134/S1064562409050056}.

\bibitem[Oseledets \& Tyrtyshnikov(2010)Oseledets and Tyrtyshnikov]{TT_cross}
Oseledets, I. and Tyrtyshnikov, E.
\newblock Tt-cross approximation for multidimensional arrays.
\newblock \emph{Linear Algebra and its Applications}, 432\penalty0 (1):\penalty0 70--88, 2010.
\newblock ISSN 0024-3795.
\newblock \doi{https://doi.org/10.1016/j.laa.2009.07.024}.
\newblock URL \url{https://www.sciencedirect.com/science/article/pii/S0024379509003747}.

\bibitem[Oseledets(2011)]{TT_decomp_oseledets}
Oseledets, I.~V.
\newblock Tensor-train decomposition.
\newblock \emph{SIAM Journal on Scientific Computing}, 33\penalty0 (5):\penalty0 2295--2317, 2011.
\newblock \doi{10.1137/090752286}.
\newblock URL \url{https://doi.org/10.1137/090752286}.

\bibitem[Pan \& Zhang(2022)Pan and Zhang]{SupremSim}
Pan, F. and Zhang, P.
\newblock Simulation of quantum circuits using the big-batch tensor network method.
\newblock \emph{Physical Review Letters}, 128\penalty0 (3):\penalty0 030501, 2022.

\bibitem[Paszke et~al.(2019)Paszke, Gross, Massa, Lerer, Bradbury, Chanan, Killeen, Lin, Gimelshein, Antiga, Desmaison, Kopf, Yang, DeVito, Raison, Tejani, Chilamkurthy, Steiner, Fang, Bai, and Chintala]{pytorch}
Paszke, A., Gross, S., Massa, F., Lerer, A., Bradbury, J., Chanan, G., Killeen, T., Lin, Z., Gimelshein, N., Antiga, L., Desmaison, A., Kopf, A., Yang, E., DeVito, Z., Raison, M., Tejani, A., Chilamkurthy, S., Steiner, B., Fang, L., Bai, J., and Chintala, S.
\newblock Pytorch: An imperative style, high-performance deep learning library.
\newblock In \emph{Advances in Neural Information Processing Systems 32}, pp.\  8024--8035. Curran Associates, Inc., 2019.

\bibitem[Perez-Garcia et~al.(2007)Perez-Garcia, Verstraete, Wolf, and Cirac]{MPS_Paper}
Perez-Garcia, D., Verstraete, F., Wolf, M.~M., and Cirac, J.~I.
\newblock Matrix product state representations, 2007.

\bibitem[Sitzmann et~al.(2019)Sitzmann, Thies, Heide, Nie{\ss}ner, Wetzstein, and Zollhofer]{sitzmann2019deepvoxels}
Sitzmann, V., Thies, J., Heide, F., Nie{\ss}ner, M., Wetzstein, G., and Zollhofer, M.
\newblock Deepvoxels: Learning persistent 3d feature embeddings.
\newblock In \emph{Proceedings of the IEEE/CVF Conference on Computer Vision and Pattern Recognition}, pp.\  2437--2446, 2019.

\bibitem[Stoudenmire(2021)]{tensornetwork_org}
Stoudenmire, E.~M.
\newblock The tensor network, 2021.
\newblock URL \url{https://tensornetwork.org/mpo}.
\newblock Support for the Tensor Network is provided by the Flatiron Institute, part of the Simons Foundation.

\bibitem[Studio(2023)]{marseille_img}
Studio, O.~.
\newblock Aerial drone view of urban buildings from top.
\newblock \url{https://www.pexels.com/@orbital101studio/}, 2023.
\newblock Accessed: 17-11-2023.

\bibitem[Sun et~al.(2022)Sun, Sun, and Chen]{DVGO}
Sun, C., Sun, M., and Chen, H.
\newblock Direct voxel grid optimization: Super-fast convergence for radiance fields reconstruction.
\newblock In \emph{CVPR}, 2022.

\bibitem[Szeliski(2022)]{szeliski2022computer}
Szeliski, R.
\newblock \emph{Computer vision: algorithms and applications}.
\newblock Springer Nature, 2022.

\bibitem[Tucker(1966)]{Tucker1966}
Tucker, L.~R.
\newblock Some mathematical notes on three-mode factor analysis.
\newblock \emph{Psychometrika}, 31\penalty0 (3):\penalty0 279--311, Sep 1966.
\newblock ISSN 1860-0980.
\newblock \doi{10.1007/BF02289464}.
\newblock URL \url{https://doi.org/10.1007/BF02289464}.

\bibitem[Usvyatsov et~al.(2022)Usvyatsov, Ballester-Ripoll, and Schindler]{usvyatsov2022tntorch}
Usvyatsov, M., Ballester-Ripoll, R., and Schindler, K.
\newblock tntorch: Tensor network learning with pytorch, 2022.

\bibitem[Vermeer(1665)]{Vermeer1665GirlPearlEarring}
Vermeer, J.
\newblock {Girl with a Pearl Earring}.
\newblock \url{https://commons.wikimedia.org/wiki/Category:Girl_with_a_Pearl_Earring_by_Johannes_Vermeer}, 1665.
\newblock Accessed: 17-11-2023.

\bibitem[Vidal(2007)]{MERA}
Vidal, G.
\newblock Entanglement renormalization.
\newblock \emph{Physical Review Letters}, 99\penalty0 (22), nov 2007.
\newblock \doi{10.1103/physrevlett.99.220405}.
\newblock URL \url{https://doi.org/10.1103\%2Fphysrevlett.99.220405}.

\bibitem[White(1992{\natexlab{a}})]{DMRG}
White, S.~R.
\newblock Density matrix formulation for quantum renormalization groups.
\newblock \emph{Physical review letters}, 69\penalty0 (19):\penalty0 2863, 1992{\natexlab{a}}.

\bibitem[White(1992{\natexlab{b}})]{whiteDMRG}
White, S.~R.
\newblock Density matrix formulation for quantum renormalization groups.
\newblock \emph{Physical review letters}, 69\penalty0 (19):\penalty0 2863, 1992{\natexlab{b}}.

\bibitem[Xie et~al.(2022)Xie, Takikawa, Saito, Litany, Yan, Khan, Tombari, Tompkin, Sitzmann, and Sridhar]{xie2022neural}
Xie, Y., Takikawa, T., Saito, S., Litany, O., Yan, S., Khan, N., Tombari, F., Tompkin, J., Sitzmann, V., and Sridhar, S.
\newblock Neural fields in visual computing and beyond.
\newblock In \emph{Computer Graphics Forum}, volume~41, pp.\  641--676. Wiley Online Library, 2022.

\bibitem[Yang et~al.(2022)Yang, Benaim, Jampani, Genova, Barron, Funkhouser, Hariharan, and Belongie]{yang2022polynomial}
Yang, G., Benaim, S., Jampani, V., Genova, K., Barron, J., Funkhouser, T., Hariharan, B., and Belongie, S.
\newblock Polynomial neural fields for subband decomposition and manipulation.
\newblock \emph{Advances in Neural Information Processing Systems}, 35:\penalty0 4401--4415, 2022.

\bibitem[Yu et~al.(2021{\natexlab{a}})Yu, Fridovich{-}Keil, Tancik, Chen, Recht, and Kanazawa]{plenoxels}
Yu, A., Fridovich{-}Keil, S., Tancik, M., Chen, Q., Recht, B., and Kanazawa, A.
\newblock Plenoxels: Radiance fields without neural networks.
\newblock \emph{CoRR}, abs/2112.05131, 2021{\natexlab{a}}.
\newblock URL \url{https://arxiv.org/abs/2112.05131}.

\bibitem[Yu et~al.(2021{\natexlab{b}})Yu, Li, Tancik, Li, Ng, and Kanazawa]{PlenOctrees}
Yu, A., Li, R., Tancik, M., Li, H., Ng, R., and Kanazawa, A.
\newblock Plenoctrees for real-time rendering of neural radiance fields.
\newblock \emph{CoRR}, abs/2103.14024, 2021{\natexlab{b}}.
\newblock URL \url{https://arxiv.org/abs/2103.14024}.

\bibitem[Yuan \& Zhao(2023)Yuan and Zhao]{yuan2023slimmerf}
Yuan, S. and Zhao, H.
\newblock Slimmerf: Slimmable radiance fields, 2023.

\end{thebibliography}
\bibliographystyle{icml2024}

%%%%%%%%%%%%%%%%%%%%%%%%%%%%%%%%%%%%%%%%%%%%%%%%%%%%%%%%%%%%%%%%%%%%%%%%%%%%%%%
%%%%%%%%%%%%%%%%%%%%%%%%%%%%%%%%%%%%%%%%%%%%%%%%%%%%%%%%%%%%%%%%%%%%%%%%%%%%%%%
% APPENDIX
%%%%%%%%%%%%%%%%%%%%%%%%%%%%%%%%%%%%%%%%%%%%%%%%%%%%%%%%%%%%%%%%%%%%%%%%%%%%%%%
%%%%%%%%%%%%%%%%%%%%%%%%%%%%%%%%%%%%%%%%%%%%%%%%%%%%%%%%%%%%%%%%%%%%%%%%%%%%%%%
\newpage
\appendix
\onecolumn
This appendix material complements the main paper by providing detailed explanations and extended analyses. 

Sec.~\ref{sec:sup_html} discusses the supplementary webpage. In Sec.~\ref{sec:limitations}, we discuss the limitations of our approach and its constraints and boundaries. In Sec.~\ref{appendix:seed_variability}, we discuss the statistical robustness of our experiments. In Sec.~\ref{app:TNs}, we give detailed descriptions of various tensor network methods, their complexities, and specific adaptations for our study. 
Sec.~\ref{app:init_influence}, we discuss the influence of initialization on model performance.
In Sec.~\ref{sec:experiment_setup}, we provide insights into our experimental framework, including batch sizes, learning rate strategies, and handling of noise and incomplete data. In Sec.~\ref{app:Putt_nerf}, we provide additional novel view synthesis results. In Sec.~\ref{app:no_shrinkage}, we explain the compatibility of TensoRF's "shrinkage" with QTT. In Sec.~\ref{app:impainting}, we explore the inpainting capabilities with PuTT. In Sec.~\ref{app:slimmerf}, we explore rank incrementation with SlimmeRF~\cite{yuan2023slimmerf}.

\section{Supplementary Webpage}
\label{sec:sup_html}

Additional supplementary materials and the associated code can be accessed at \url{https://sebulo.github.io/PuTT_website/}.

\section{Limitations}
\label{sec:limitations}
In our compression results in Sec.~\ref{subsec:compression_quality} of the main text, we observe certain constraints associated with our PuTT method. While learning QTT representations using PuTT show a marked advantage in scenarios requiring high compression, as evident in our results for 2D and 3D data (Figures \ref{fig:2d_psnr_ssim}, \ref{fig:3d_psnr_ssim}) and in Novel View Synthesis (Tab.~\ref{table:results_nerf} of the main text), their effectiveness tends to converge with other methods like CP, Tucker, and VM in less compressed settings with smaller tensor sizes.
For example, in the case of 1k resolution images (referenced in Sec.~\ref{subsec:compression_quality}, Fig.~\ref{fig:2d_psnr_ssim} of the main text), we noted that the performance benefits of using QTTs were less pronounced. This suggests that while QTTs are highly effective in handling large-scale, high-compression tasks, their advantages are less significant in simpler scenarios with lower compression.

\section{Sensitivity and Statistical Analysis}
\label{appendix:seed_variability}

In our experimental framework, we tested each configuration using three distinct seeds to ensure the reliability and consistency of our results. 
We use error bars representing $\pm 1$ standard deviation from the mean in all plots.
In the context of our 2D and 3D experiments, it's noteworthy that the error bars are often so minimal (see Sec.~\ref{subsec:compression_quality} of the main text, Fig.~\ref{fig:2d_psnr_ssim}) that they might be challenging to discern in the plots. This observation points to a low degree of variation, suggesting a high level of reproducibility in our results. Such consistency is particularly evident in experiments involving upsampling, where a consistent trend towards similar endpoints in training was observed. This pattern can be attributed to the progressive nature of our approach, which systematically builds the representation on increasingly finer grids.
Furthermore, when comparing the performance of our PuTT method with other established methods like Tucker, CP, and VM, we observed less variation in PuTT's performance across different experimental setups. This reduced variability, especially visible in the compression plots, indicates that PuTT not only ensures a more stable and predictable performance but also underscores its effectiveness in different experimental conditions and settings. %\seb{Should we make a new plot where the variability is easier to see?}

\subsection{Enhancing Training Stability through Upsampling}

\begin{figure}[h]
    \centering
    \begin{subfigure}[b]{0.45\textwidth}
        \centering
        \includegraphics[width=\textwidth]{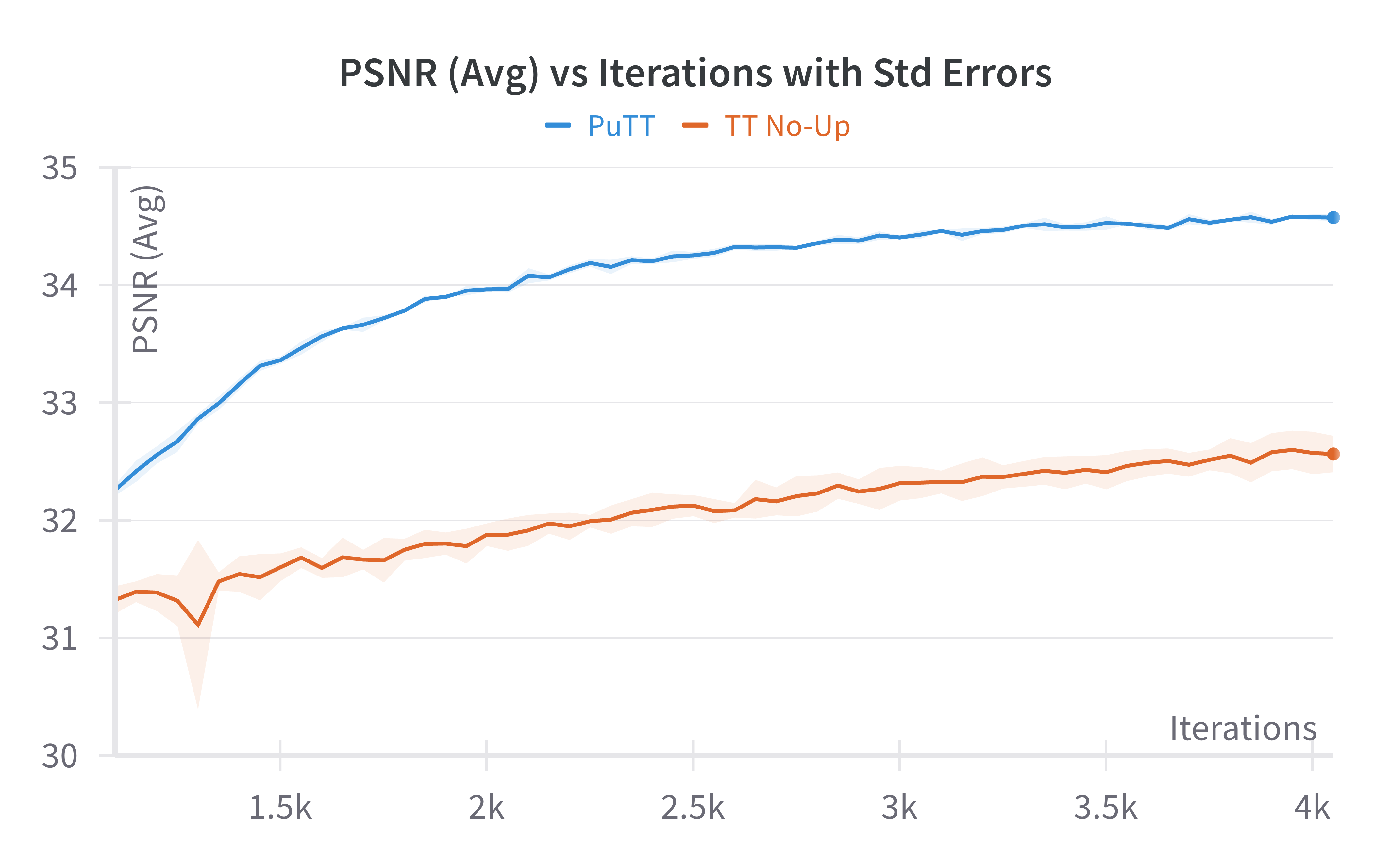}
        \caption{Girl 4K}
        \label{subfig:sub_a}
    \end{subfigure}
    \hfill 
    \begin{subfigure}[b]{0.45\textwidth}
        \centering
        \includegraphics[width=\textwidth]{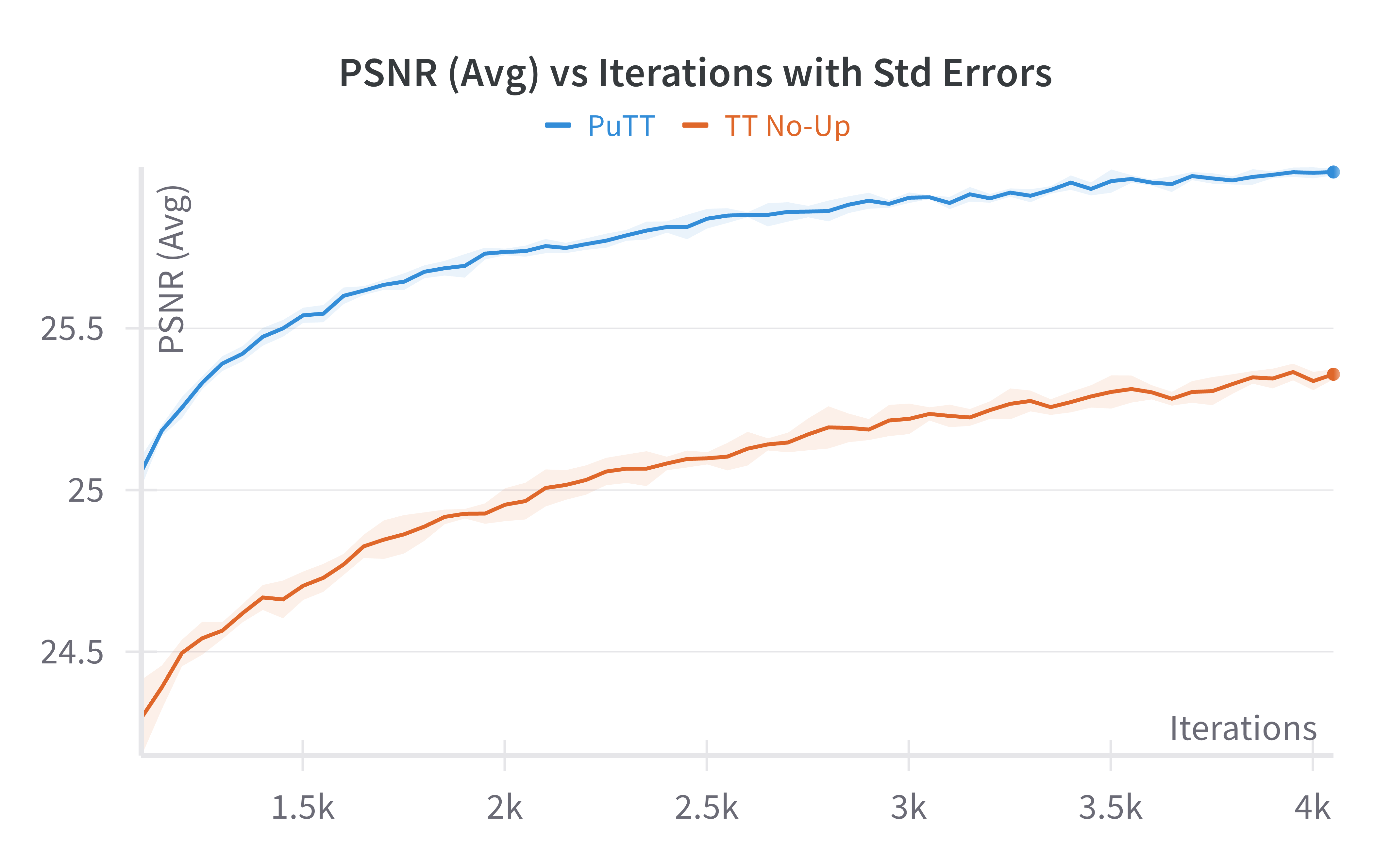} 
        \caption{Tokyo 4K}
        \label{subfig:sub_b}
    \end{subfigure}
    \caption{The two figures present plots corresponding to the "Girl with Pearl Earring" (\ref{subfig:sub_a}) and "Tokyo" 4K images (\ref{subfig:sub_b}), showcasing the performance of our PuTT method with upsampling versus a baseline QTT model without upsampling. For both images, we plot the average PSNR over $4000$ iterations, averaged across three seeds. The shaded regions around each curve represent the variability within $\pm 1$ standard deviation from the mean PSNR, serving as an indicator of training stability. In both plots, it is evident that the upsampling model exhibits lower fluctuation in standard deviation and achieves higher PSNR levels by the end of the training period. This highlights the improved stability and effectiveness of our upsampling approach in 2D compression tasks.}
    \label{fig:stabalize_exp}
\end{figure}

Figure \ref{fig:stabalize_exp} shows an experiment further illustrating the stabilizing effect of our upsampling strategy within the context of 2D compression tasks. The experiment analyses the "Girl with Pearl Earring" and "Tokyo" images, both at 4K resolution, to draw comparisons between our upsampling approach and a baseline QTT model with no upsampling.
The findings, visualized in our plot, reveal the average PSNR against the number of iterations, with the shaded area depicting the variance within $\pm 1$ standard deviation across three separate runs. This comparison underscores the enhanced stability of training with our upsampling method, evidenced by reduced performance fluctuations and consistently higher PSNR levels compared to models without upsampling.

\section{Tensor Networks}
\label{app:TNs}

We describe the three main families of tensor decompositions that are considered in this work: CP/VM, and Tucker. Tensor Trains and the specific QTT format are defined in the main article. We provide additional information on MPOs and the general complexity of manipulating tensor networks.  
%\sbe{Only describe the ones used directly relevant to our method. The rest should go in supplementary}

\paragraph{The canonical polyadic (CP) decompositions~\cite{CP1, CP2}} factorizes a tensor $\mathcal{T}\in \mathbb{R}^{n_1\times n_2\times \cdots n_d}$ into a sum of rank-$1$ components:
\begin{equation}
    \mathcal{T}=\sum^R_{r=1}v_r^1 \otimes v_r^2 \otimes \cdots \otimes v_r^d
\end{equation}
where $v^1_r, v_r^2, \cdots, v_r^d$ are rank-$1$ vectors in $\mathbb{R}^{n_1}, \cdots, \mathbb{R}^{n_d}$. Each tensor element is then a sum of scalar products:
\begin{equation} \label{eq:cp_sample}
    \mathcal{T}_{j_1\cdots j_d}=\sum^R_{r=1}\sum_{k=1}^d v_{r,j_k}^k,
\end{equation}
where $j_k= 1,..., n_k$ for each $k=1,...,d$. We will mostly consider $d=2,3$ in this work, as we care about visual data. 

\paragraph{The Vector Matrix (VM) decomposition~\cite{tensoRF}} is a specific 3D tensor decomposition, introduced to remedy a "too high" compactness of the CP decomposition. Instead of purely using vector factors, the VM decomposition factorizes a tensor into vectors and matrices. For $\mathcal{T}\in\mathbb{R}^{n_1\times n_2\times n_3}$,
\begin{equation} \label{eq:vm}
\mathcal{T}=\sum^{R_1}_{r=1}v_r^1 \otimes M_r^{2,3} + \sum^{R_2}_r v_r^2 \otimes M_r^{1,3} + \sum_{r=1}^{R_3} v_r^3 \otimes M_r^{1,2}
\end{equation}
where the superscript on the vectors $v$ and matrices $M$ indicate the space on which they act. The VM decomposition can be understood as a half-way between the dense (bare) representation of $\mathcal{T}$ and the CP decomposition. As such, VM has reduced compression as compared with CP.
The overall rank of the VM decomposition is equal to the total number of components $(R_1+R_2+R_3)$.

\paragraph{The Tucker decomposition~\cite{Tucker1966}} provides only a partial compression of a tensor. Given a tensor $\mathcal{T}\in \mathbb{R}^{n_1\times n_2\times \cdots n_d}$, we define a reduced tensor $\mathcal{K}\in \mathbb{R}^{m_1\times m_2\times\cdots m_d}$ and matrices $U^k\in \mathbb{R}^{m_k\times n_k}$, such that 
\begin{equation}
    \mathcal{T}_{j_1,...,j_d} = \sum_{i_1=1}^{m_1}\cdots\sum_{i_d}^{m_d}\mathcal{K}_{i_1,...,i_d}U^1_{i_1,j_1}\cdots U^d_{i_d,j_d}. 
\end{equation}
The Tucker decomposition is mostly useful in settings where $d$ is small, and $n_k$ are large.

\paragraph{Matrix Product Operators}
A matrix product operator (MPO) is a tensor network that describes a factorization of a tensor with $d$ input and $d$ output indices into a linear network of smaller tensors~\cite{MPO_link1, MPO_link2}. Each core (except for the first and last) has two physical indices and two virtual indices.

The difference between TTs and MPOs is the nature of the objects they are meant to represent. A TT can be seen as a parameterization of a large vector in a high-dimensional space, providing a compressed yet expressive representation of such vectors. The "operator" part of the term "Matrix Product Operator" refers to the mathematical operation of a matrix acting on a vector~\cite{tensornetwork_org}, i.e., the MPO operates on a TT. This essentially allows one to efficiently represent and work with large operators in high-dimensional spaces, working directly in the compressed TT representation. 

Using the same index notation as for TTs, an MPO is a tensor network of the form:
\begin{equation}\label{eq:MPO}
\mathcal{M}_{i_1 ...i_d}^{j_1,...,j_d} = A^{i_1,j_1}A^{i_2,j_2}...A^{i_d,j_d},
\end{equation}
where $A^{i_k,j_k}$ are $R_k\times R_{k+1}$ matrices for each pair $i_k,j_k$. 
MPOs have important applications in representing large, sparse matrices in a form that is convenient for TT algorithms~\cite{DMRG,ALS,TT-Prolongation}. They play a central role in TT algorithms in Physics, yet have been under-exploited in the data science setting. Here we show how to leverage MPOs in the visual data setting in connection with multi-resolution analysis. 

\paragraph{Complexity of the CP, VM, Tucker, and TT decompositions}

Given a tensor with $d$ indices of maximal dimension $n$, CP decomposition stores $dR_{CP}$ rank-$1$ tensors of size $n$, resulting in a space complexity of $O(dnR_{CP})$. Here, $R_{CP}$ is the  CP rank. In 3D, the VM decomposition stores $R_{VM}$ times a size $n$ vector and a $n\times n$ matrix, resulting in a space complexity $O(R_{VM}(n+n^2))$. The Tucker decomposition stores a tensor $\mathcal{K}$ with maximal index range $m$ and $d$ matrices of size $n\times m$ resulting in a space complexity $O(dm^{d+1}n)$. Finally, the tensor train stores $n d$ matrices of size $R_{TT}^2$, resulting in a space complexity $O(n d R_{TT}^2)$.% Recall that the $o(\cdot)$ notation indicates an upper bound on the complexity. 
In practice, the complexity is lower, as one can further exploit the tensor network structures to reduce the individual ranks in the network.

If $d$ is large - typically $d> 3$ - then we expect the memory cost to scale as Tucker, CP $\gg$ TT. While the separation between Tucker and TT is obvious, since Tucker scales exponentially in $d$, the separation between CP and TT is contained in the rank dependence. For fixed accuracy, we expect TT to require a constant rank $R_{TT}$, independent of $d$. While the CP rank $R_{CP}$ will grow up to exponentially in $d$~\cite{tensordecompositions}. 

However, with visual data, we have $d\leq 4$, hence the relationships are more subtle and depend crucially on the representational power of each decomposition; i.e. what ranks are necessary to represent visual data up to a given precision, and whether these representations can be learned in an efficient and reliable manner. In the QTT representation, the complexity is $O(2^d \log(n)R^2)$.

In summary, one should expect that for $d=2,3,4$ and large $n$ (i.e. high resolution), the QTT format should outperform CP, VM, and Tucker, with the gap growing up to exponentially as $d$ and $n$ grow. 

\paragraph{Computational Complexity of Prolongation:} For a tensor $T$ of dimensions $D^N$, where $N=2$ typically represents the dimensions of an image (height and width), the upper bound for the computational complexity involved in multiplying a Matrix Product Operator (MPO) with a Quantized Tensor Train (QTT) is denoted by $O(\log(D) \cdot \max(R, S)^3 \cdot N^D)$. In this expression, $(\log(D))$ denotes the number of cores in the QTT, which corresponds to the level of quantization. $(R)$ and $(S)$ are the maximum ranks within the QTT and MPO, respectively. This operation is efficiently executed and does not significantly influence the overall runtime.

\paragraph{Implementation}
When learning tensor networks such as Tucker, CP, TT, QTT, and VM, especially in challenging scenarios with noisy, incomplete data or in applications like NeRF where the full scene tensor is not accessible, it becomes essential to adopt a sample-based training approach. Unlike traditional analytic techniques such as Higher-Order Singular Value Decomposition (HOSVD)~\cite{Tucker1966} for Tucker decomposition, or Singular Value Decomposition (SVD) and QR decompositions for TT \cite{TT_decomp_oseledets} and QTT~\cite{QTT_paper}, our methodology does not rely on these direct decomposition methods. Instead, we commence with an initial skeleton tensor that mirrors the structure required by the chosen decomposition scheme. This skeleton serves as a starting point for an iterative optimization process, where the tensor values are refined using stochastic gradient descent. This approach allows for a flexible and efficient adaptation of the tensor factorization methods to the intricacies and limitations of the data and application context, such as those encountered in NeRF scenarios.

\subsection{Multiresolution Format of Quantized Tensor Train}
\label{app:multi_res_qtt}

\begin{figure}[h]
    \centering
    \includegraphics[width=0.7\textwidth]{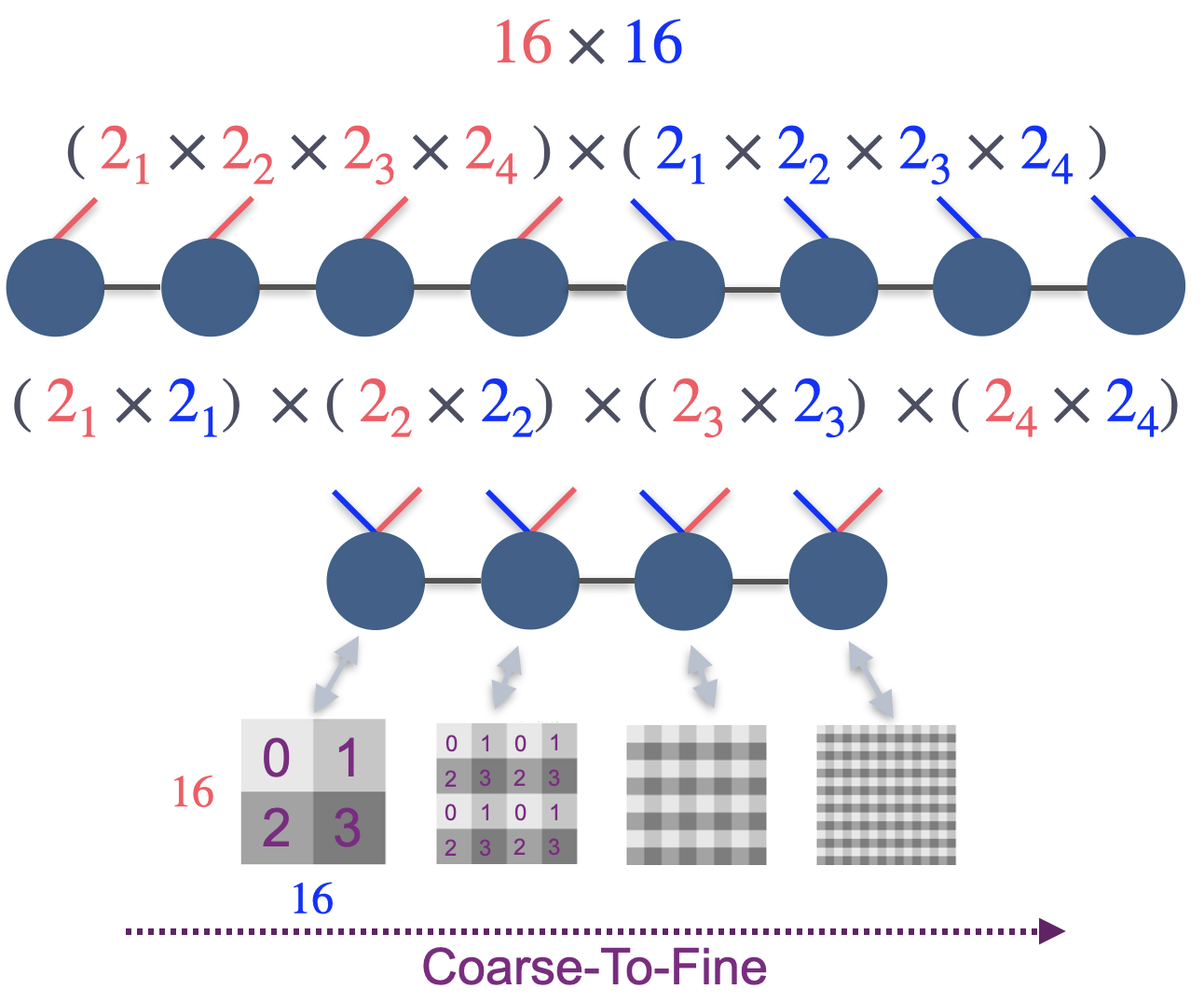}
    \caption{
    This figure illustrates the transformation of a 16x16 matrix into QTT format, showcasing a transition from a standard grid representation to a mixed factor order. Initially presented as a simple $16_x$ by $16_y$ matrix, it is restructured into factors of two: $\left(2_{1 x} \times\right.$ $\left.2_{2 x} \times 2_{3 x} \times 2_{4 x}\right)$ combined with $\left(2_{1 y} \times 2_{2 y} \times 2_{3 y} \times 2_{4 y}\right)$. Subsequently, the factors are permuted into a mixed factor order $\left(2_{1 x} \times 2_{1 y}\right) \times$ $\left(2_{2 x} \times 2_{2 y}\right) \times\left(2_{3 x} \times 2_{3 y}\right) \times\left(2_{4 x} \times 2_{4 y}\right)$. This results in a multiresolution format where the initial core's indices correspond to indexing into four quadrants of the input matrix. Specifically, index 0 in the first core is associated with the top-left quadrant of the grid. Subsequent cores progressively examine increasingly detailed segments of the matrix. This hierarchical breakdown allows the QTT to capture and represent data at various levels of granularity, starting from broad, overarching structures down to more intricate details within the grid, which is reminiscent of a wavelet.
}
    \label{fig:coarse_to_fine_struc}
\end{figure}

The multiresolution format, as explained in Sec.\ref{sec:notation}, is illustrated in Fig.\ref{fig:coarse_to_fine_struc}.

\subsection{The prolongation operators}
\label{app:mpo}
We employ a Matrix Product Operator (MPO) known as a prolongation operator $\mathcal{P}$, suggested by Lubach et al.~\cite{TT-Prolongation}, to upsample coarse solutions in the Quantized Tensor Train (QTT) format.

Consider a one-dimensional vector $v_d\in \mathbb{R}^{2^d}$ representing data like a continuous function. To achieve a finer representation $v_{d+1}\in\mathbb{R}^{2^{d+1}}$, we linearly interpolate between adjacent points in $v_d$ using the matrix $P_{2^d\rightarrow 2^{d+1}}$. This matrix resembles the Haar transform matrix. For instance for $d=2$, we can express $P_{2^d\rightarrow 2^{d+1}}$ as:
\begin{equation}
P_{4 \rightarrow 8}=\left(\begin{matrix} .5&0&0&0\\
1&0&0&0\\
.5&.5&0&0\\
0&1&0&0\\
0&.5&.5&0\\
0&0&1&0\\
0&0&.5&.5\\
0&0&0&1\\
\end{matrix}\right),
\end{equation}
This matrix is closely related to the Haar transform matrix. 
The matrix $P_{2^d\rightarrow 2^{d+1}}$ can be written as an MPO $\mathcal{P}$ with bond dimension two as 
\begin{equation} \label{eq:MPO_structure_propagation}
\mathcal{P}_{j_1, ..., j_n}^{i_1, ..., i_n, i_{n+1}} =  P[1]^{j_1, i_1} \cdots P[d]^{j_d, i_d} P[d+1]^{i_{n+1}}. 
\end{equation}
The middle $d-1$ tensors in the MPO, $P[k]^{j_k,i_k}_{\alpha_k,\alpha_{k+1}}$ have two physical indices $j_k,i_k$ and two virtual indices $\alpha_k,\alpha_{k+1}$. Each index $i_k,j_k,\alpha_k\in [0,1]$. The first tensor $P[1]^{j_1,i_1}_{\alpha_1}$ has one virtual index $\alpha_1$ and two physical indices $j_1,i_1$, while the last tensor $P[d+1]^{i_{d+1}}_{\alpha_d}$ has one virtual index $\alpha_d$ and one physical index $i_{d+1}$. The entries are given explicitly as: 
\begin{eqnarray}
    P[1]_0^{0,0} &=& P[1]_0^{1,1} = P[1]_1^{1,0} = 1 \nonumber\\ 
    P[k]_{0,0}^{0,0} &=& P[k]_{1,0}^{0,1} = P[k]_{0,1}^{1,1} = P[i]_{1,1}^{1,0} = 1, \nonumber \\
    && \text{for } 1 < k < d+1 \nonumber\\
    P[d+1]_{0}^0 &=& 1, \quad \text{and } P[d+1]_{0}^1 = P[d+1]_{1}^1 = \frac{1}{2}, \nonumber
\end{eqnarray}
with all other entries being zero~\cite{TT-Prolongation}\footnote{Note the difference in convention in index notation as compared to~\cite{TT-Prolongation}}. 

The previously described prolongation operator $\mathcal{P}$ applies to tensor train representations of one-dimensional tensors in QTT format. In higher dimensions, the prolongation operator is simply the tensor product of the one-dimensional operators on each dimension:  $\mathcal{P} \bigotimes \mathcal{P}$ for $2$-dimensions and $\mathcal{P} \bigotimes \mathcal{P} \bigotimes \mathcal{P}$ for $3$-dimensions. 

\subsection{Quantized Tensor Train Initialization}
\label{app:qtt_initialziation}
We initialize the TT-ranks between each core to follow a trapezoid structure like first introduced by Oseledets et al.~\cite{TT_decomp_oseledets}.
In a trapezoid structure, the TT-ranks may increase up to a certain point (creating the rising edge of the trapezoid), then stay constant, and then decrease (creating the falling edge of the trapezoid). The top (flat part) of the trapezoid corresponds to the maximum rank among all cores, which is what we call the TT-rank. 

This structure can be beneficial because it allows representing a high-dimensional tensor in a compressed form which still maintains a good approximation of the original tensor~\cite{TT_decomp_oseledets}.\\

To compute the TT-ranks in the context of the trapezoid structure, consider an array containing the $D$ physical indices of a TT $n = [n_1, n_2, \ldots, n_D]$. The TT-ranks are derived as follows:
\begin{enumerate}
    \item Compute the cumulative product of the tensor shape from left to right and from right to left, denoted as \texttt{ranks\_left} and \texttt{ranks\_right} respectively:
    \begin{align*}
        \texttt{ranks\_left}[i] &= \prod_{j=1}^{i} n_j, \quad \forall i = 1, \ldots, D \\
        \texttt{ranks\_right}[i] &= \prod_{j=i}^{D} n_j, \quad \forall i = 1, \ldots, D
    \end{align*}
    \item Then, calculate the TT-ranks by taking the minimum of the \texttt{ranks\_left} and \texttt{ranks\_right} at each position:
   \begin{align*}
    \texttt{ranks\_tt}[i] &= \min(\texttt{ranks\_left}[i], \texttt{ranks\_right}[i]), \\
    &\quad \forall i = 1, \ldots, D
    \end{align*}

    \item Finally, if a rank, \texttt{max\_rank}, is specified, ensure that no index rank exceeds this value:
    \begin{align*}
        \texttt{ranks\_tt}[i] &= \min(\texttt{ranks\_tt}[i], \texttt{max\_rank}), \\
        &\quad \forall i = 1, \ldots, D
    \end{align*}

\end{enumerate}
From this, it follows that if the payload dimension is greater than $1$, then we have to scale the dimensions of each core from the left to the right until we reach the rank top of the trapezoid structure of the TT. \\

\begin{equation}
\label{Eq:TT_init_sigma}
\hat{\sigma} = \exp \left( \frac{1}{2D} \left(  2 \log \sigma - \Sigma_{i=1}^D \log R_i \right) \right) 
\end{equation}

\subsection{TT-SVD Baseline}
The Tensor Train - Singular Value Decomposition (TT-SVD) algorithm, as detailed by Oseledets et al.~\cite{TT_decomp_oseledets}, offers a method for decomposing high-dimensional tensors into tensor trains. It sequentially reshapes and applies matrix SVD to each tensor mode, transforming a high-dimensional tensor into a series of interconnected lower-dimensional tensors, effectively forming a tensor train. For an in-depth understanding of TT-SVD, Oseledets et al.'s work provides comprehensive details.

As also pointed out in Sec.~\ref{subsec:baselines_metrics} of the main text, TT-SVD has limitations, particularly in its applicability only when the direct signal (e.g., a 2D image or 3D voxel grid) is provided for an analytical application. It falls short in scenarios requiring optimization, such as learning from incomplete data, novel view synthesis, or handling noisy data.
Another limitation of TT-SVD is its infeasibility when the input tensor is too large for memory. 

In our work, TT-SVD, implemented via the TNTorch framework~\cite{usvyatsov2022tntorch}, serves as a baseline to determine whether our Prolongation Upsampling Tensor Train (PuTT) method can escape local minima without specific hyperparameters. This benchmarking aids in assessing PuTT's effectiveness in handling large-scale tensor data and its ability to navigate the challenges of optimization.
In our implementation, we adapt TT-SVD for the QTT format. This adaptation involves initially reshaping the input tensor into $\log(N)$ factors of $2$, rearranging these factors to enforce Z-order correlation between spatial dimensions, and then applying TT-SVD. This process aligns with how we typically construct the specialized QTT structure.

\subsection{Differences Between Vector Quantization and QTT}
VQ compresses data by mapping input vectors to a set of discrete symbols from a codebook. In VQ, the codebook is typically generated by clustering a large set of input vectors. Each input vector is then replaced by the index of the closest codebook entry. During decompression, the original vectors are reconstructed using the codebook entries. 
representation through a limited set of prototype vectors.

QTT, on the other hand, pertains to a tensor decomposition approach where a high-dimensional tensor is factorized into a sequence of lower-dimensional tensors, referred to as "cores". These cores are connected in a train-like structure, which gives the method its name. The term "Quantized" in this context refers to the decomposition of the tensor into these smaller, discrete units or "quants", enabling a compact and efficient representation of the tensor's multidimensional structure.
Such distinction was also discussed in the previous work of TT-NF by Obukhov et al\cite{ttnf}. (Discussion in Appendix page 19), which mentions: “The term “quantization” in machine learning often refers to the reduction of the dynamic range of values of learned parameters in a parametric model”. 

It is also noteworthy that VQ and our QTT factorization approach are orthogonal techniques. There exists a potential to merge these methodologies, where each tensor core in a QTT might be quantized using a learned dictionary, presenting an intriguing avenue for future research.

\section{Influence of Initialization on QTT Performance}
\label{app:init_influence}
To examine the influence of initialization standard deviations on model performance, we conducted an experiment utilizing the Lena $512\times512\times3$ image across 3000 iterations. This experiment aims to analyze how varying the initialization standard deviation ($\sigma$) of tensor cores impacts the efficacy of our models, particularly comparing our PuTT method against a QTT model without upsampling. The experiment spanned a range of $\sigma$ values from $0.001$ to $0.5$, with the TT-NF initialization approach also evaluated for comparison and is shown in Table \ref{tab:init_exp}. 
Notably, PuTT exhibited remarkable consistency in PSNR values, averaging around $36.10$ regardless of the selected $\sigma$ value. This stability highlights the effectiveness of PuTT's progressive learning strategy in mitigating the impacts of initialization variability. In contrast, the QTT model without upsampling displayed considerable sensitivity to $\sigma$ variations, with pronounced fluctuations in PSNR values across the tested spectrum. At a higher $\sigma$ of $0.5$, the QTT model's performance significantly deteriorated, underscoring the model's challenges in learning effectively under such initialization conditions. The highest PSNR observed for the QTT model without upsampling was $34.96$, attained at an intermediate $\sigma$ value, illustrating the nuanced role of initialization in model learning.

\begin{table}[ht]
\centering
\begin{tabular}{l|c|c}
\hline
\textbf{$\sigma$} & \textbf{PuTT} & \textbf{QTT w/o Upsampling} \\
\hline
0.001 & 36.103 $\pm$ 0.001 & 32.676 $\pm$ 0.082 \\
0.005 & 36.101 $\pm$ 0.001 & 32.873 $\pm$ 0.107 \\
0.01 & 36.116 $\pm$ 0.013 & 34.845 $\pm$ 0.102 \\
0.05 & 36.100 $\pm$ 0.002 & 34.958 $\pm$ 0.083 \\
0.1 & 36.116 $\pm$ 0.001 & 34.946 $\pm$ 0.092 \\
0.5 & 36.103 $\pm$ 0.006 & 12.320 $\pm$ 1.610 \\
TT-NF init. & 36.116 $\pm$ 0.001 & 34.941 $\pm$ 0.102 \\
\hline
\end{tabular}
\caption{Comparison of average PSNR ($\pm$1 std) values for Lena 512x512x3 image over 3000 iterations with various initialization standard deviations ($\sigma$) per core, mean of zero, for PuTT and QTT without upsampling. TT-NF init. is the TT-NF initialization as detailed in Appendix D.2 Line 774. PuTT employs four upsampling steps at iterations $[50, 100, 200, 400]$, with a learning rate of $5e$-$3$ and batch size of $128^2$. Results averaged over three seeds. }
\label{tab:init_exp}
\end{table}

\section{Experiment Setup}
\label{sec:experiment_setup}

The core of our experiments involve comparing the effectiveness of PuTT against established tensor-based methods (CP, VM, and Tucker) across both 2D and 3D grayscale data, as well as in the setting of incomplete or noisy data. Due to VM's quadratic dependency on the side length of the input tensor, its application to 2D images is impractical, as it fails to provide a compressed representation. 
The quantitative experiments for 2D, Noise, and incomplete data are performed on grayscale images to reduce the amount of computation required. These results are consistent with color image examples.

We test these methods at five distinct resolutions from 1k to 16k in doubling steps for images and from $64^3$ to $1024^3$ for 3D, examining three compression ratios for each resolution to ensure a comprehensive analysis. For each specific combination of resolution and compression ratio, we compute the average values of the PSNR and SSIM metrics as well as $\pm 1$ standard deviation.

The PuTT implementation starts by initializing the QTT in PyTorch~\cite{pytorch}, as outlined in Sec.~\ref{app:qtt_initialziation}. The parameter optimization utilizes the Adam algorithm~\cite{kingma2014adam}. PuTT's upsampling steps and iteration counts are tailored to the desired resolution for each task. For instance, to attain a 2D resolution of $1024^2$, the model undergoes three upsampling steps from an initial $128^2$ resolution at the $64^{th}$, $128^{th}$, and $256^{th}$ iterations, culminating in $1024$ iterations in total. A resolution of $2048^2$ involves four upsampling steps and $2048$ iterations, starting from $128^2$. This strategy allows substantial initial training at a lower resolution, enhancing efficiency and mitigating the risk of overfitting to low-resolution aspects, before transitioning to full resolution for the remaining training to capture more complex details. For Noise and Incomplete Data scenarios, we limit our focus to median resolutions ($4k$ and $256^3$) and double the iteration count due to the increased difficulty these present.

For detailed configurations of each resolution and task, refer to Tab.~\ref{tab:upsampling_configs}.
% \sbe{make sure to add all citations and address any errors}

\begin{table*}[ht]
\centering
\begin{tabular}{l r r l r}
\toprule
Experiment & Full Resolution & Initial Resolution & Iterations for Upsampling & Total Iterations \\
\midrule
2D & $1024^2$ & $128^2$ & [64, 128, 256] & 1024 \\
   & $2048^2$ & $128^2$ & [64, 128, 256, 512] & 2048 \\
   & $4096^2$ & $128^2$ & [64, 128, 256, 512, 1024] & 4096 \\
   & $8192^2$ & $128^2$ & [64, 128, 256, 512, 1024, 2048] & 8192 \\
   & $16384^2$& $128^2$ & [64, 128, 256, 512, 1024, 2048, 4096] & 16384 \\
\midrule
3D & $64^3$ & $8^3$ & [16, 48, 144] & 512 \\
   & $128^3$ & $8^3$ & [16, 48, 144, 432] & 1536 \\
   & $256^3$ & $8^3$ & [16, 48, 144, 432, 1296] & 4608 \\
   & $512^3$ & $8^3$ & [16, 48, 144, 432, 1296, 3888] & 13824 \\
   & $1024^3$ & $8^3$ & [48, 144, 432, 1296, 3888, 11664, 34992] & 69984 \\
\midrule
Noise 2D & $4096^2$ & $128^2$ & [64, 128, 256, 512, 1024, 2048] & 8192 \\
\midrule
Noise 3D & $256^3$ & $8^3$ & [16, 48, 144, 432, 1296, 3888] & 13824 \\
\midrule
Incomplete Data 2D & $4096^2$ & $128^2$ & [64, 128, 256, 512, 1024, 2048] & 8192 \\
\midrule
PuTT-NeRF & $256^2$ & $32^2$ & [500, 1000, 4000] & 80000 \\
\bottomrule
\end{tabular}
\caption{Configurations for upsampling iterations of all experiments. Resolution details the specific resolution in pixels/voxels, with the initial starting resolutions, iterations for upsampling, and iteration counts for different resolutions and experiments.}
\label{tab:upsampling_configs}
\end{table*}

\subsection{Batch Size and Learning Rate}
\label{appendix:batchsize_and_learning_rate}
% \sbe{makes sense to put this near the experimental setup}

In our compression experiments with PuTT, CP, Tucker, and VM, we did a comprehensive analysis to determine the ideal learning rate and batch size combinations. This entailed evaluating batch sizes ranging from $32^2$ to $1024^2$ and learning rates from $10^{-1}$ to $10^{-4}$. Our findings indicated that larger batch sizes generally led to improved results but with a trade-off in increased training time. Striking a balance between efficiency and accuracy, we opted for a batch size of $512^2$ with a base learning rate of $5\cdot10^{-3}$.
For the high-resolution case of $1024^3$, we had to adjust the batch size to $128^2$ since Tucker and CP methods due to computational limits. Instead, we doubled the number of iterations. 
 We found that PuTT was more sensitive to the learning rate and was configured at $0.005$. In contrast, the CP, VM, and Tucker methods utilize a slightly higher learning rate of $0.01$, as these configurations were found to be optimal for each respective model. These settings balance the computational demands with the need for accuracy and efficiency in our experiments.

\begin{figure*}
\centering
\includegraphics[width=\textwidth]{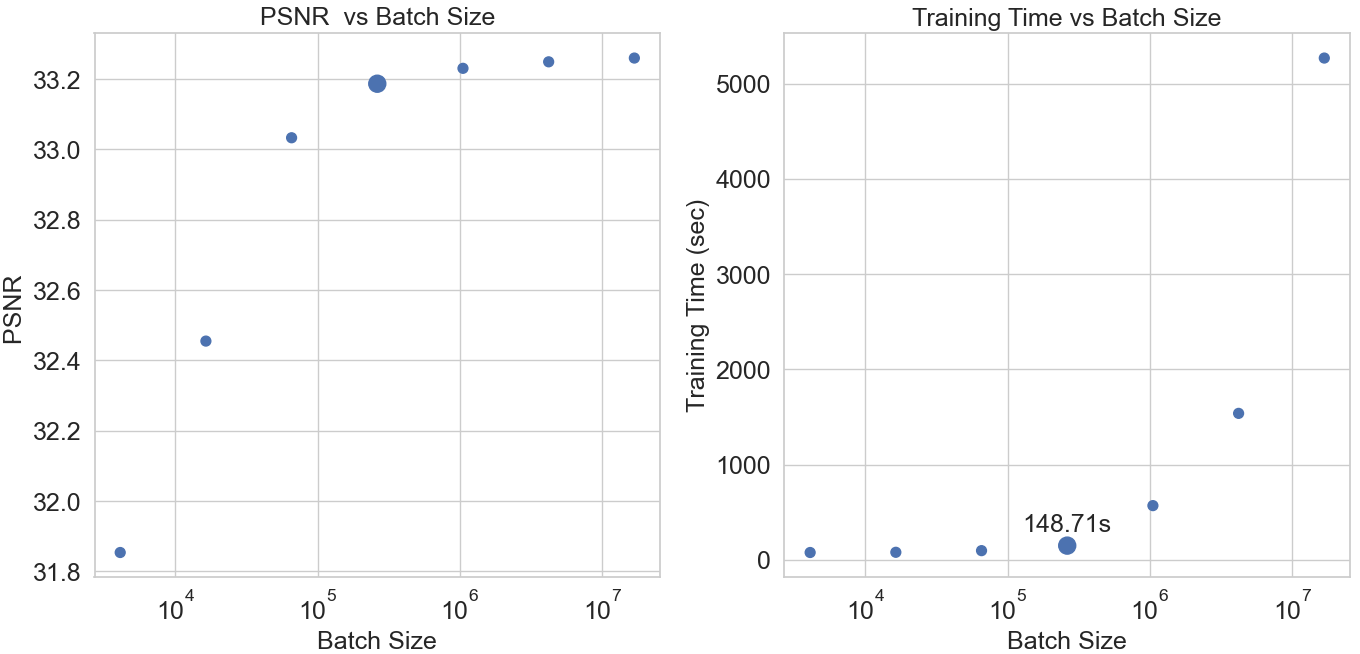}
\caption{The graph showcases the impact of batch size on PSNR (LHS) and training time (RHS). Various learning rates from $[0.0001, 0.1]$ were tested, and the optimal rate for each batch size was identified. The training spanned over 4096 iterations with upsampling steps at intervals $[64, 128, 256, 512, 1024]$. The batch size of $512^2$, highlighted in the figure, was selected for its optimal balance between high accuracy and time efficiency, as evidenced in the training of the 4k "Girl with a Pearl Earring" image with a rank of 200.}
\label{fig:batch_size_vs_acc_appendix}
\end{figure*}

Fig.~\ref{fig:batch_size_vs_acc_appendix} illustrates these relationships, displaying how different batch sizes influenced both the PSNR and the training time in our experiments with a 4k "Girl with a Pearl Earring" image using a PuTT.

\subsection{Learning Rate Strategy in Training Process}
\label{appendix:learning_rate_strategy}

In our training setup, we implement an exponential learning rate decay strategy with a decay factor of $\alpha=0.1$. This decay reduces the learning rate exponentially by $\alpha$ between upsampling iterations. For example, if upsampling occurs at iteration $1000$, the learning rate declines from $5\cdot10^{-3}$ to  $\alpha \cdot 5\cdot10^{-3} = 5\cdot10^{-4}$ from step $0$ to step $1000$. 

We observe a characteristic pattern in the loss behavior post-upsampling: an initial spike in loss due to the introduction of higher-resolution interpolations. As the number of parameters in the representation increases, a new optimizer with an adjusted learning rate is necessary. Simply maintaining the pre-upsampling learning rate leads to minimal loss reduction, potentially trapping the model in local minima. Conversely, resetting to the base learning rate causes significant loss fluctuations.

To address this, we employ a $\beta$ upsampling learning rate decay, where $\beta=0.9$. Post-upsampling, the learning rate is recalibrated to $0.9^l \cdot 5\cdot10^{-3}$ (where $l$ is the number of upsampling steps completed), ensuring a balanced approach between loss stability and effective learning. Following this, we introduce a 50-iteration warm-up phase for the learning rate after each upsampling, enhancing the stability of the training process. Finally, post the last upsampling step, the learning rate undergoes another exponential decay of $\alpha$ until the final iteration, ensuring a smooth convergence towards the end of the training.

\subsection{Compression}
\begin{figure*}
    \centering
\includegraphics[width=\linewidth]{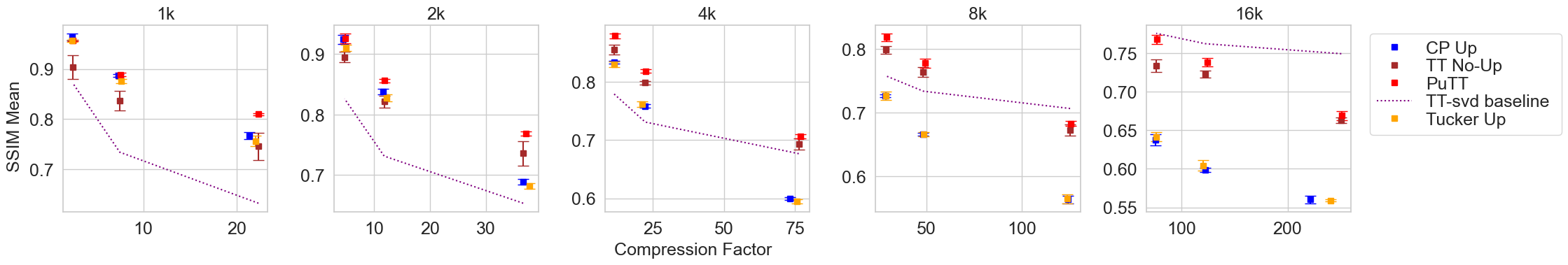} \\
    \vspace{-0.2cm}
    \caption{ % (a) 
    SSIM (y-axis) vs compression ratio (x-axis) for 2D fitting. ``up = upsampling".
    } 
    \label{fig:2d_psnr_ssim_app}
    \vspace{-0.4cm}
\end{figure*}

\begin{figure*}
\centering
\includegraphics[width=\linewidth]{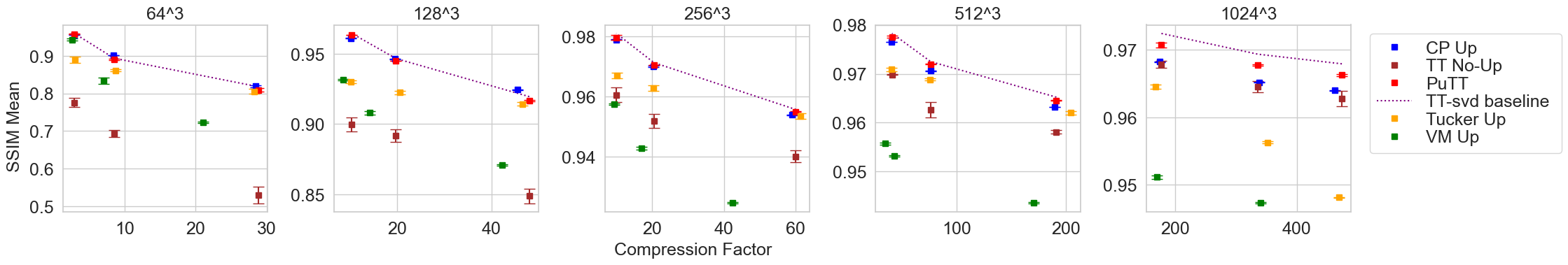} \\
\vspace{-0.2cm}
    \caption{(a) SSIM (y-axis) in comparison to compression ratio (x-axis) for 3D fitting. 
    } 
\label{fig:3d_psnr_ssim_app}
\vspace{-0.4cm}
\end{figure*}

In Fig. \ref{fig:2d_psnr_ssim_app},\ref{fig:3d_psnr_ssim_app} we see the SSIM results for our 2D and 3D compression experiments that complements Fig. \ref{fig:2d_psnr_ssim} and \ref{fig:3d_psnr_ssim} (PSNR results) of the main text. Similarly to the PSNR results, we see that PuTT outperform baselines for both 2D and 3D.

\subsection{Noise Removal}
\label{appendix:noise}

In Sec.~\ref{subsec:noise_removal} of the main text, we evaluate PuTT's proficiency in learning from noisy samples, comparing it with various tensor network methods and QTTs without upsampling. 
We aim to learn a tensor representation of a target, denoted as $\mathcal{X}_I$, using samples from its noisy counterpart, $\hat{\mathcal{X}}_I$, obtained by 
This noisy version has an identical structure to $\mathcal{X}_I$ and is generated by adding noise $Z$. In our experiments, $Z$ is sampled from either a Normal or Laplacian distribution, $Z \sim \mathcal{N}(0, \sigma)$, or a Laplacian distribution, $Z \sim \text{Laplace}(0, b)$. 
Thus, the noisy target is formulated as $\hat{\mathcal{X}}_I = \mathcal{X}_I + Z$.

\begin{figure*}[t!]
\centering
\includegraphics[width=\linewidth]{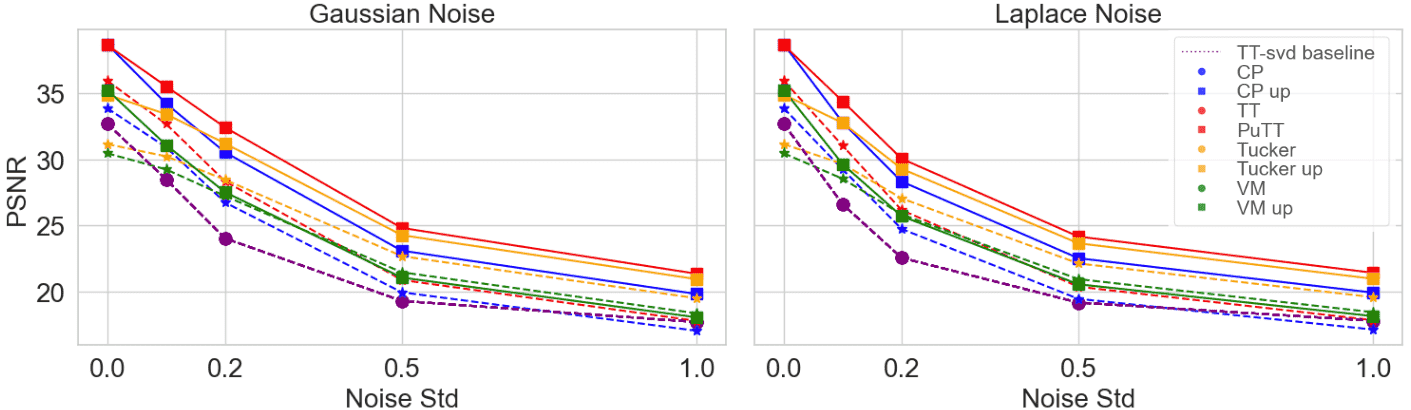}
\caption{PSNR when varying the amount of Gaussian or Laplacian noise for PuTT and baselines for 3D experiments. ``up = upsampling".}
\label{fig:noise3d_psnr_appendix}
\end{figure*}

\begin{figure*}[t!]
\centering
\includegraphics[width=\linewidth]{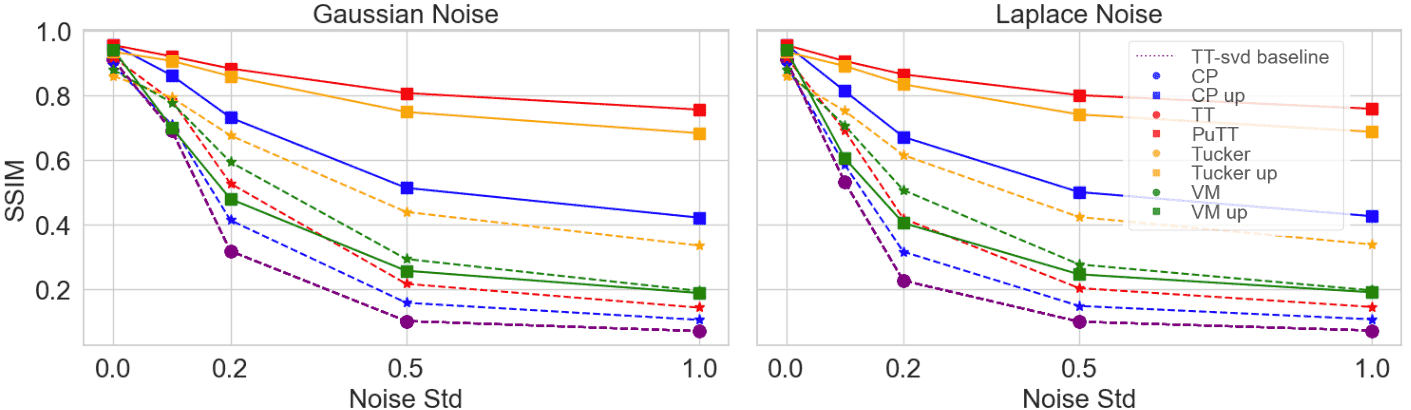}
\caption{ Same as Fig.~\ref{fig:noise3d_psnr_appendix} but for SSIM}
\label{fig:noise3d_ssim_appendix}
\end{figure*}

In Sec.~\ref{subsec:noise_removal} of the main text, Figures~\ref{fig:noise_exps} (a) and (b) present quantitative results for noise resilience in 2D focusing on PSNR and SSIM metrics.
In the 3D noise resilience analysis (Figures~\ref{fig:noise3d_psnr_appendix} and \ref{fig:noise3d_ssim_appendix}), PuTT mirrors its 2D success, maintaining higher PSNR against other methods across various noise levels. Upsampling significantly boosts performance, especially in SSIM metrics for noise above $0.2 \sigma$, demonstrating PuTT's strong noise handling capabilities in both 2D and 3D scenarios.

For our Noise Removal experiments, we adjust the learning rate to accommodate varying noise levels. Starting with an initial learning rate of $lr_{init} = 0.005$ at zero noise, we modify this rate for both PuTT and baseline methods according to the formula $lr_{\sigma} = 0.005 \times 0.1^{\sigma}$, where $0.1$ is our chosen adaptation factor. For instance, at a noise level of $\sigma = 0.5$, the learning rate becomes $lr_{\sigma} = 0.005 \times 0.1^{0.5} \approx 0.00158$. This approach helps in achieving optimal results across different noise intensities.

\subsection{Incomplete Data Experiments}
\label{app:subsampled_targets}

\begin{figure*}[ht!]
    \centering
    \begin{subfigure}[b]{\textwidth}
        \includegraphics[width=\textwidth]{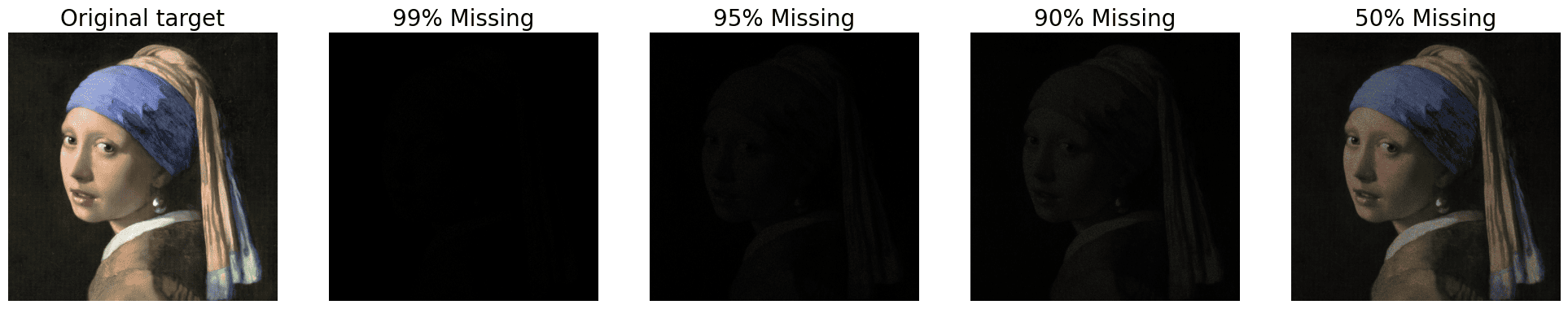}
        \caption{Original and Incomplete Images }
    \end{subfigure}
    \caption{Incomplete data training inputs for various Percentages: The sequence starts with the original target image (leftmost), followed by the varying percentages of missing data from 99\% to 50\%.}
    \label{fig:subsamples_dif_percentages_appendix}
\end{figure*}

In Sec.~\ref{subsec:subset_sampling} of the main text, we presented the results for training on limited data for various percentages. Fig.~\ref{fig:subsamples_dif_percentages_appendix} presents the training data used in training at various percentages. The sequence begins with the original target image, followed by its version with a percentage of the original data sampled. These versions are created by randomly setting a specified percentage of the pixels in the original image to zero, representing scenarios with limited data availability. The subsampling percentages, ranging from 1\% to 50\%, are specified above each corresponding image.

\begin{figure*}
    \centering
    \includegraphics[width=\linewidth]{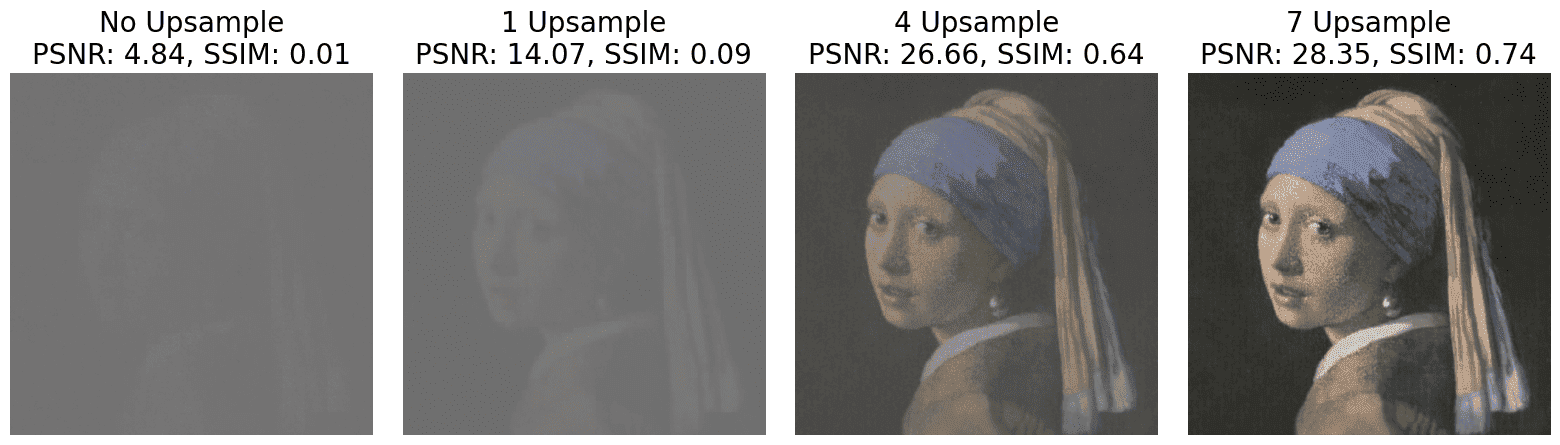} 
    \caption{Varying the number of upsampling steps when training $99\%$ missing data. The first image displays the training data, followed by PuTT results with $0, 1, 4$, and $7$ upsampling steps. 
    }
    \label{fig:ablation_visual_upsampling_incomplete_data_appendix}
    \vspace{-0.4cm}
\end{figure*}

In Sec.~\ref{subsec:subset_sampling} of the main text, Fig.~\ref{fig:ablation_visual_upsampling_incomplete_data} shows how varying the number of upsampling steps when training $99\%$ missing data affects PSNR and SSIM for "Girl with Pearl Earrings" 4k. The first image displays the training data, followed by PuTT results with $0, 1, 4$, and $7$ upsampling steps.
Fig.~\ref{fig:ablation_visual_upsampling_incomplete_data_appendix} shows the larger visual examples.

In our Incomplete Data experiments, similar to the Noise Removal experiments, we modify the learning rate based on the proportion of available training data. We observed that optimal results were achieved with lower learning rates when less training data was accessible. Initially, we set the learning rate to $lr_{init} = 0.005$ when all data points are available. As the amount of training data decreases, we adjust the learning rate for PuTT and baseline methods using the formula $lr_{p} = 0.005 \times f^{1-p}$. Here, $f$ represents the adaptation factor, and $p$ denotes the proportion of available training data, ranging between 0 and 1. This strategy helps us tailor the learning process effectively to different data availability scenarios.

\begin{figure*}[ht!]
    \centering
    \begin{subfigure}[b]{\textwidth}
        \includegraphics[width=\textwidth]{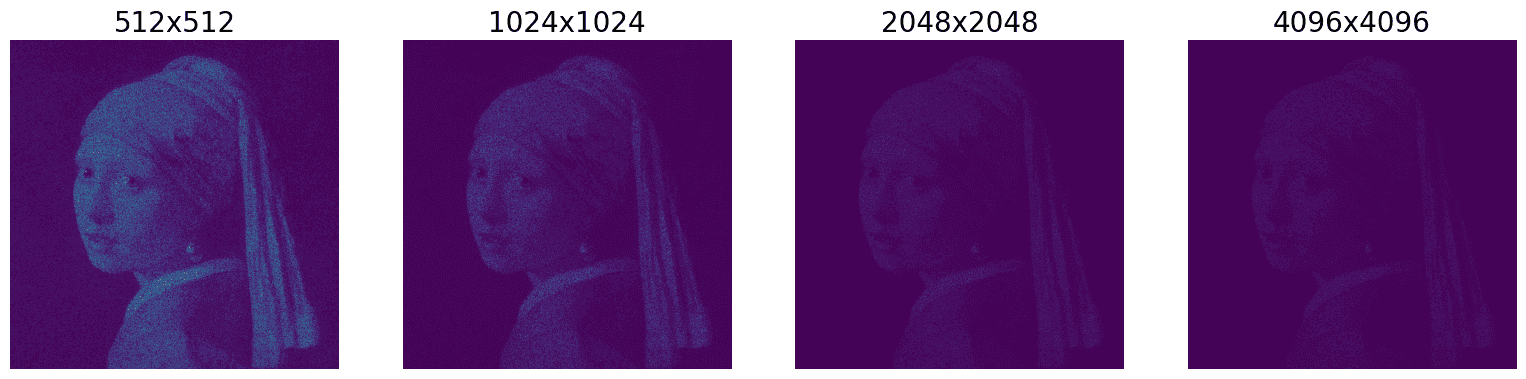}
    \end{subfigure}
    \caption{Illustration of the effects of downsampling a 4k image where $95\%$ of the data at the target resolution is missing. The progression from left to right shows images at lower resolutions containing aggregated information about the pixels available at higher resolutions. The three channels have been aggregated to more clearly show this pattern}
    \label{fig:downsampled_targets}
\end{figure*}

\noindent \textbf{Downsampled images} \quad 
Fig.~\ref{fig:downsampled_targets} shows the downsampled targets used during training for $95\%$ missing data of "Girl with Pearl Earring". The three RGB channels are averaged to visualize how aggregated values form a blurred version of the actual target at lower downsampled resolutions.

\noindent \textbf{Masked average pooling} \quad 
As explained in Sec.~\ref{subsec:subset_sampling} of the main text, to train with a downsampled target at lower resolutions, we create a modified target image $I_{D-l,p}$. We build $I_{D-1,p}$ from $I_{D,p}$, we use a custom masked average pooling method, averaging only the non-zero values in each window, sized according to the downsampling factor. 
The masked average pooling can be illustrated through a simple example. Consider a $2 \times 2$ window containing values $[2, 2, 0, 0]$. In standard average pooling, the output would be the average of all values, $(2 + 2 + 0 + 0) / 4 = 1$. However, in our masked average pooling, where zero values are excluded from the calculation, the output is the average of the non-zero values, $(2 + 2) / 2 = 2$. This method ensures that only relevant data contributes to the pooling process.
Fig.~\ref{fig:downsampled_targets} provides a visual representation of this process. It showcases how a 4k image, with 95\% of its data missing at the target resolution, is progressively downsampled. Each image in the sequence, moving from left to right, represents a lower-resolution version, consolidating aggregated information from the available pixels at higher resolutions. For easier visualization, the channel dimension has been averaged.

\begin{figure*}[t!]
\centering
\begin{tabular}{cc}
\includegraphics[trim={0 0 0cm 0},clip,width=0.48\linewidth]{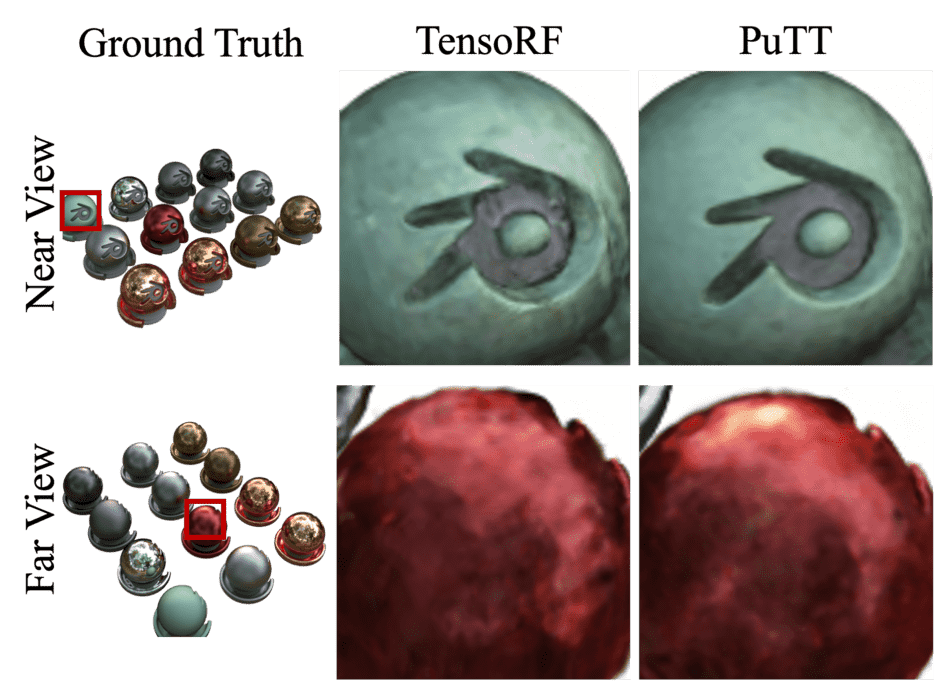} & 
\includegraphics[trim={0 0 0cm 0},clip,width=0.48\linewidth]{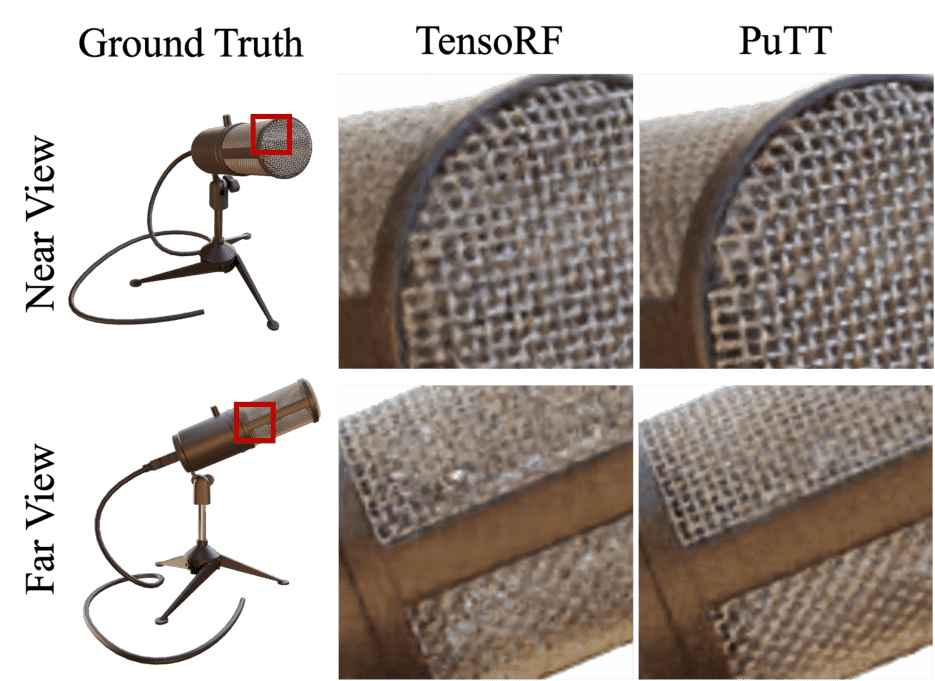} \\
\includegraphics[trim={0 0 0cm 0},clip,width=0.48\linewidth]{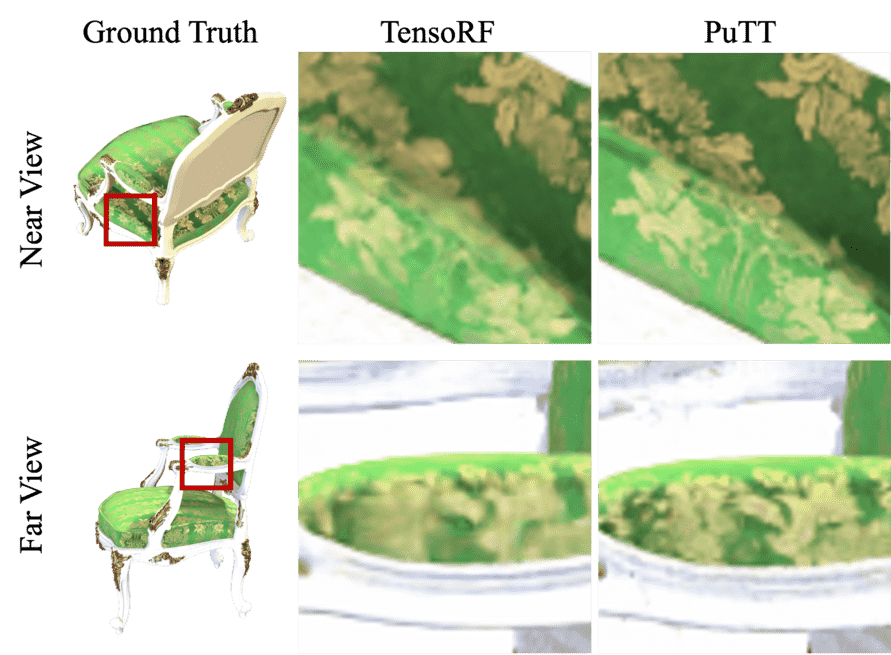} & 
\includegraphics[trim={0 0 0cm 0},clip,width=0.48\linewidth]{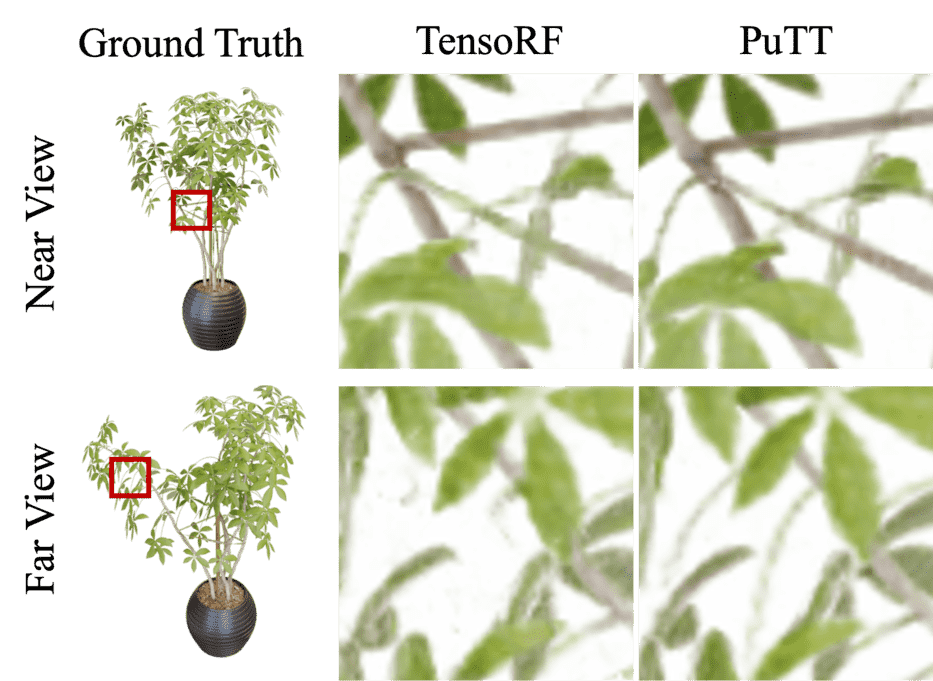} 
%\vspace{-0.2cm}
\end{tabular}
\caption{Visual comparison of TensoRF and PuTT on Near and Far views for ``Materials", ``Mic", ``Chair", ``Ficus" scenes.}
\label{fig:nerf-supp}
\vspace{-0.4cm}
\end{figure*}

\section{Novel View Synthesis}
\label{app:Putt_nerf}

%%%%%%%%%%%%%%%%%%%
\begin{table}[ht]
\centering
\small % Smaller font size
\begin{tabular}{lccc}
\toprule
Model (Size) & PSNR & SSIM & LPIPS \\
\midrule
PuTT (12 MB) & \textbf{31.953} & \textbf{0.957} & \textbf{0.0597} \\
PuTT (12 MB) No PE & 31.880 & 0.9537 & 0.0612 \\
PuTT (12 MB) No Up & 31.466 & 0.9502 & 0.0654 \\
TensoRF (12 MB) & 31.623 & 0.9523 & 0.0626 \\
\midrule 
PuTT (7 MB) & \textbf{31.665} & \textbf{0.9560} & \textbf{0.0602} \\
PuTT (7 MB) No PE & 31.633 & 0.9516 & 0.0650 \\
PuTT (7 MB) No UP & 31.295 & 0.9480 & 0.0710 \\
TensoRF (7 MB) & 31.103 & 0.9476 & 0.0697 \\
\bottomrule
\end{tabular}
\caption{Comparison of novel view synthesis (NeRF) under different model sizes (7MB and 12MB) in comparison to TensoRF. "No Up": No upsampling, "No PE": No positional encoding used for inputs to the Shader (MLP) function.}
\label{tab:nerf_performance_app}
\vspace{-0.4cm}
\end{table}

\label{appendix:nerf_experiments}

We compare our approach with other grid-based novel view synthesis methods, including previous works NSVF~\cite{NSVF} PlenOctrees~\cite{PlenOctrees}, Plenoxels~\cite{plenoxels}, DVGO~\cite{DVGO}, K-planes~\cite{fridovichkeil2023kplanes}, and TensoRF~\cite{tensoRF}. Most of these results were taken directly from the respective papers if possible.
We directly compare the tensor methods used by TensoRF and their efficiency vs our PuTT method's effeciency, by aligning the model sizes of the two. 
In our comparison with TensoRF, we omitted its "shrinkage" feature, which involves using an alpha mask to dynamically reshape and resample tensor factors during training. However, due to its current incompatibility with our QTT structure, integrating this feature into PuTT requires additional development. Instead, in our experiments, we employed a progressively updated alpha mask, utilized exclusively for ray-filtering purposes. This adjustment was critical in maintaining a fair and relevant comparison between the methods under test.

In our novel view synthesis experiments, models were trained for a total of $80,000$ steps. The learning rate (LR) schedule followed the same structure as outlined in Sec.~\ref{subsec:compression_quality} of the main text, but for PuTT, we used a base LR of $1e-3$. In the case of TensoRF, we adhered to its standard configuration but adjusted the factor counts in the VM decomposition to align with our 7MB and 12MB model sizes. 
For the 7MB TensoRF models, the rank factors were set to $[4,4,4]$ for density and $[5,5,5]$ for appearance. In contrast, for the larger 12MB models, we increased these ranks to $[8,8,8]$ for both density and appearance factors, ensuring a consistent comparison in terms of memory footprint. For PuTT 7MB is equal to a rank of $140$ for the density grid and $200$ for the appearance grid, and for the 12MB model, this is $200$ and $280$ respectively.

Regarding batch sizes, our experiments utilized $4096$ rays per batch. The number of uniform samples per ray was capped at $884$. Both models, PuTT and TensoRF, demonstrated similar running times, ranging between 4 and 5 hours for both 7MB and 12MB on an NVIDIA Tesla V100 GPU. The increased running time observed with TensoRF compared to those reported in their paper can be attributed to not employing its shrinkage feature. While this feature typically enhances efficiency in TensoRF's processing, its exclusion was necessary to ensure a more direct comparison with the PuTT model under equivalent conditions.

We assessed our method under different conditions: with and without Positional Encoding (PE), and with and without Upsampling, applied to both the 7MB and 12MB model sizes. Additionally, these configurations were contrasted with TensoRF, both with and without its Upsampling feature. Each experiment was replicated over three seeds to ensure robustness, with the outcomes detailed in Tab.~\ref{tab:nerf_performance_app}.

\subsection{Omitting Positional Encoding}
\label{sec_no_pos_encoding}
One of the key innovations in the original NeRF~\cite{nerf} was the use of Positional Encoding (PE) to transform 5D input coordinates into a higher-dimensional space, enhancing the representation of high-frequency scene content. This technique resulted in a significant PSNR improvement for NeRF~\cite{nerf}. Our findings, as shown in Tab.~\ref{tab:nerf_performance_app}, suggest that the quantized TT structure effectively encodes inputs for the MLP responsible for color value computations based on viewing directions. With the quantized TT structure, we observed only a minor drop in performance when PE was omitted. For the 12MB model, we only see a drop of $0.07$ PSNR when not using Positional Encoding as opposed to the drop reported by Mildenhall et al. in NeRF \cite{nerf}, which was a difference of $2.24$ PSNR. This highlights the effectiveness of the quantized tensor train encoding.

\subsection{Testing from Near to Far Views }

\begin{table*}[]
\small
\resizebox{\textwidth}{!}{
\begin{tabular}{cccccccccccccccccccc}
\toprule
\multirow{2}{*}{Metric} & \multirow{2}{*}{Method} & \multicolumn{2}{c}{Chair} & \multicolumn{2}{c}{Drums} & \multicolumn{2}{c}{Ficus} & \multicolumn{2}{c}{Hotdog} & \multicolumn{2}{c}{Lego} & \multicolumn{2}{c}{Materials} & \multicolumn{2}{c}{Mic} & \multicolumn{2}{c}{Ship} & \multicolumn{2}{c}{Average} \\ 
                        &                         & TensoRF      & PuTT       & TensoRF      & PuTT       & TensoRF      & PuTT       & TensoRF      & PuTT        & TensoRF     & PuTT       & TensoRF        & PuTT         & TensoRF     & PuTT      & TensoRF     & PuTT       & TensoRF       & PuTT        \\ \midrule
\multirow{3}{*}{PSNR}   & Near                    & 35.953       & \textbf{37.450}      & 27.024       & \textbf{27.672}     & 32.653       & \textbf{34.254}     & 37.176       & \textbf{38.276}      & 34.396      & \textbf{35.579}     & 28.915         & \textbf{29.309}       & 33.510       & \textbf{34.886}    & 30.463      & \textbf{30.990}      & 32.511      & \textbf{33.552}      \\
                        & Far                     & 31.798       & \textbf{33.262}     & 24.444       & \textbf{24.951}     & 30.457       & \textbf{31.822}     & \textbf{31.579}       & 31.331      & 31.434      & \textbf{31.523}     & 25.986         & \textbf{26.177}       & 33.515      & \textbf{34.382}    & \textbf{27.287}      & 27.264     & 29.563       & \textbf{30.089}      \\
                        & All                     & 32.616       & \textbf{34.144}     & 25.004       & \textbf{25.444}     & 30.986       & \textbf{32.253}     & 36.034       & \textbf{36.590}       & 33.264      & \textbf{34.009}     & 28.953         & \textbf{29.331}       & 32.520       & \textbf{33.596}    & 29.128      & \textbf{29.559}     & 31.063     & \textbf{31.866}   \\ \bottomrule    
\end{tabular}}
\caption{Novel View Synthesis comparison to TensoRF (PSNR) for all scenes in the Synthetic-NeRF~\cite{nerf} dataset. We test on views that are near training views (Near), far from training views (Far), and all test views (All). The models being compared have a size og 7MB.
}
\vspace{-0.3cm}
\label{tab:nerf_NearFar_ssim_lpips_appendix}
\end{table*}

To study the fine-grained representation ability of PuTT, we experiment with testing views captured from near to far camera positions compared with source views.
Specifically, we compute the average variance in view angles between each testing view and its nearest twenty source views. After sorting the average view angle differences for all testing views, we identify the twenty testing views with the smallest differences as near views and the twenty testing views with the largest differences as far views.

In addition to Sec.~\ref{subsec:compression_quality} of the main text, Tab.~\ref{tab:nerf_NearFar_ssim_lpips_appendix} shows that PuTT outperforms TensorRF from near to far testing views on both SSIM and LPIPS for all scenes in the dataset. It should be noticed that the advantage of PuTT over TesorRF on near views is more significant than that on far views. It indicates that PuTT, benefiting from the coarse-to-fine learning method, learns a more compact and expressive representation of the fine-grained details, as shown in Fig.~\ref{fig:nerf-supp}.

\subsection{Per-scene results novel view synthesis}
\label{app:nerf_results_per_scene}
%%%%%%%%%%%%%%%%%%%%%%%%%%%%%%%%%%%%%%%%%%%%%%%%%%%%%%%%%%%%%%%
%%%%%%%%%%%%%%%%%% Blender Dataset %%%%%%%%%%%%%%%%%%%%%%%%%%%
%%%%%%%%%%%%%%%%%%%%%%%%%%%%%%%%%%%%%%%%%%%%%%%%%%%%%%%%%%%%%%%
\begin{table*}[htpb]
    \vspace{2em}
    \centering
    \begin{tabular}{l|c|cccccccc}
    \hline
    Methods & Avg. & {\it Chair} & {\it Drums} & {\it Ficus} & {\it Hotdog} & {\it Lego} & {\it Materials} & {\it Mic} & {\it Ship} \\
    \hline\hline
    \multicolumn{9}{@{}l}{\rule{0pt}{3ex}\bf PSNR$\uparrow$} \\
    \hline
    PlenOctrees-L*~\cite{PlenOctrees} & 31.71 & 34.66 & 25.31 & 30.79 & 36.79 & 32.95 & 29.76 & 33.97 & 29.42 \\
    Plenoxels-L*~\cite{plenoxels} & 31.71 & 33.98 & 25.35 & 31.83 & 36.43 & 34.10 & 29.14 & 33.26 & 29.62 \\
    DVGO-L*~\cite{DVGO} & 31.95 & 34.09 & 25.44 & 32.78 & 36.74 & 34.64 & 29.57 & 33.20 & 29.13 \\
    kplanes-L*~\cite{fridovichkeil2023kplanes} & 32.36  & 34.99  & 25.66  & 31.41  & 36.78  & 35.75  & 29.48  & 34.05  & 30.74  \\
    Instant-NGP*~\cite{instantNerf}\textbf-{L} & 33.18 & 35.00 & 26.02 & 33.51 & 37.40 & 36.39 & 29.78 & 36.22 & 31.10 \\
    TensoRF VM-192-L*~\cite{tensoRF} & 33.14 & 35.76 & 26.01 & 33.99 & 37.41 & 36.46 & 30.12 & 34.61 & 30.77 \\
    Our PuTT-200-L & 32.79 & 35.29 & 25.82 & 33.39 & 37.29 & 35.93 & 29.93 & 34.21 & 30.39  \\
    \hline
    NeRF-S*~\cite{nerf} & 31.01 & 33.00 & 25.01 & 30.13 & 36.18 & 32.54 & 29.62 & 32.91 & 28.65 \\
    TensoRF VM-30-S (no shrinking)  & 31.10 & 32.62 & 25.00 & 30.99 & 36.03 & 33.26  & 28.95  & 32.52 & 29.13  \\
    Our PuTT-200-S & 31.87 & 34.14 & 25.44 & 32.25 & 36.59 & 34.01 & 29.33 & 33.60 & 29.56  \\
    \hline
    Instant-NGP-M*~\cite{instantNerf} & 33.17 & 35.00 & 26.02 & 33.51 & 37.40 & 36.39 & 29.78 & 36.22 & 31.10 \\
    TensoRF VM-48-M (no shrinking)  & 31.51 & 33.29 & 25.22 & 31.62 & 36.24 & 33.79 & 29.34 & 32.97 & 29.59  \\
    Our PuTT-200-M & 31.98 & 34.23 & 25.67 & 32.32 & 37.07 & 34.02 & 29.58 & 33.56 & 29.44  \\
    \hline
    \multicolumn{9}{@{}l}{\rule{0pt}{3ex}\bf SSIM$\uparrow$} \\
    \hline
    PlenOctrees-L*~\cite{PlenOctrees} & 0.958 & 0.981 & 0.933 & 0.970 & 0.982 & 0.971 & 0.955 & 0.987 & 0.884 \\
    Plenoxels-L*~\cite{plenoxels} & 0.958 & 0.977 & 0.933 & 0.976 & 0.980 & 0.976 & 0.949 & 0.985 & 0.890 \\
    DVGO-L*~\cite{DVGO} & 0.957 & 0.977 & 0.930 & 0.978 & 0.980 & 0.976 & 0.951 & 0.983 & 0.879 \\
    kplanes-L*~\cite{fridovichkeil2023kplanes} & 0.962 & 0.983 & 0.938  & 0.975  & 0.982 & 0.982  & 0.950  & 0.988  & 0.897   \\
    TensoRF VM-192-L*~\cite{tensoRF}   & 0.963 & 0.985 & 0.937 & 0.982 & 0.982 & 0.983 & 0.952 & 0.988 & 0.895 \\
    Our PuTT-600-L  & 0.958 & 0.977 & 0.933 & 0.979 & 0.981 & 0.972 & 0.947 & 0.986 & 0.888  \\
    \hline
    NeRF-S*~\cite{nerf} & 0.947 & 0.967 & 0.925 & 0.964 & 0.974 & 0.961 & 0.949 & 0.980 & 0.856 \\
    TensoRF VM-30-S (no shrinking)  & 0.948 & 0.965 & 0.920 & 0.966 & 0.974 & 0.966 & 0.938 & 0.979 & 0.873  \\
    Our PuTT-200-S  & 0.956 & 0.976 & 0.928 & 0.975 & 0.979 & 0.974 & 0.945 & 0.987 & 0.876 \\
    \hline
    TensoRF VM-48-M (no shrinking) & 0.952 & 0.971 & 0.923 & 0.969 & 0.975 & 0.974 & 0.944 & 0.982 & 0.879  \\
    Our PuTT-280-M  & 0.957 & 0.976 & 0.929 & 0.975 & 0.979 & 0.973 & 0.946 & 0.989 & 0.878  \\
    \hline

    \multicolumn{9}{@{}l}{\rule{0pt}{3ex}\bf LPIPS$_{Vgg}\downarrow$ } \\
    \hline
    DVGO-L*~\cite{DVGO}  & 0.035 & 0.016 & 0.061 & 0.015 & 0.017 & 0.014 & 0.026 & 0.014 & 0.118 \\
    TensoRF VM-192-L*~\cite{tensoRF}  & 0.027 & 0.010 & 0.051 & 0.012 & 0.013 & 0.007 & 0.026 & 0.009 & 0.085 \\
    Our PuTT-600-L & 0.045 & 0.021 & 0.081 & 0.025 & 0.019 & 0.027 & 0.045   & 0.018 & 0.124  \\
    \hline
    NeRF-S*~\cite{nerf} & 0.081 & 0.046 & 0.091 & 0.044 & 0.121 & 0.050 & 0.063 & 0.028 & 0.206 \\
    TensoRF VM-30-S (no shrinking)  & 0.067 & 0.042 & 0.104 & 0.034 & 0.042 & 0.039 & 0.074 & 0.025 & 0.177 \\
    Our PuTT-200-S & 0.061 & 0.037 & 0.085 & 0.029 & 0.038 & 0.039 & 0.069 & 0.024 & 0.161 \\
    \hline
    TensoRF VM-48-M (no shrinking) & 0.062 & 0.041 & 0.100 & 0.032 & 0.0412 & 0.037 & 0.072 & 0.022 & 0.155\\
    Our PuTT-280-M & 0.059 & 0.034 & 0.084 & 0.029 & 0.037 & 0.039 & 0.069   &0.022 & 0.158  \\
    \hline
    
    \hline
    \end{tabular}
    \caption{Quantitative results on each scene from the {\bf Synthetic-NeRF}~\cite{nerf} dataset. (Results denoted with an asterisk (*) were sourced from the data presented in the referenced paper.
    }
    \label{tab:supp_breakdown_nerf}
    \vspace{2em}
\end{table*}

\newpage

%%%%%%%%%%%%%%%%%%%%%%%%%%%%%%%%%%%%%%%%%%%%%%%%%%%%%%%%%%%%%%%
%%%%%%%%%%%%%%%%%% NSVF Dataset %%%%%%%%%%%%%%%%%%%%%%%%%%%
%%%%%%%%%%%%%%%%%%%%%%%%%%%%%%%%%%%%%%%%%%%%%%%%%%%%%%%%%%%%%%%
\begin{table*}[htpb]
    \centering
    \begin{tabular}{l|c|cccccccc}
    \hline
    Methods & Avg. & {\it \tiny{Wineholder}} & {\it \tiny{Steamtrain}} & {\it \tiny{Toad}} & {\it \tiny{Robot}} & {\it \tiny{Bike}} & {\it \tiny{Palace}} & {\it \tiny{Spaceship}} & {\it \tiny{Lifestyle}} \\
    \hline\hline
    \multicolumn{9}{@{}l}{\rule{0pt}{3ex}\bf PSNR$\uparrow$} \\
    \hline
    DVGO-L*~\cite{DVGO} &35.08 &30.26 &36.56 &33.10 &36.36 &38.33 &34.49 &37.71& 33.79\\
    TensoRF VM-192-L*~\cite{tensoRF} & 36.52 & 31.32 & 37.87 & 34.85 & 38.26 & 39.23 & 37.56 & 38.60 & 34.51 \\
    Our PuTT-600-L & 36.57 & 31.33 & 37.21 & 34.87& 38.22 & 39.32 & 37.47 & 39.07 & 35.05  \\
    \hline
    NeRF-S*~\cite{nerf} & 30.81 & 28.23 & 30.84 & 29.42 & 28.69 & 31.77 & 31.76 & 34.66 & 31.08 \\
    TensoRF VM-30-S (no shrinking)  & 34.58 & 29.62 & 35.51 & 32.21 & 35.42 & 37.85 & 35.49 & 37.05 & 33.48  \\
    Our PuTT-200-S & 35.59 & 30.55 & 36.22 & 33.73 & 36.44 & 38.59 & 36.02 & 38.72 & 34.43  \\
    \hline
    TensoRF VM-48-M (no shrinking) & 35.41 &  30.35 & 36.54 & 32.84 & 36.85 & 38.33 & 36.54 & 37.51 & 34.34  \\
    Our PuTT-280-M & 36.04 & 30.89 & 36.41 & 34.48 & 37.02 & 38.89 & 36.78 & 38.95 & 34.86 \\
    \hline
    
    \multicolumn{9}{@{}l}{\rule{0pt}{3ex}\bf SSIM$\uparrow$} \\
    \hline
    DVGO-L*~\cite{DVGO} &0.975& 0.949& 0.989& 0.966 &0.992& 0.991& 0.962& 0.988& 0.965 \\
    TensoRF VM-192-L*~\cite{tensoRF} & 0.982 & 0.961 & 0.991 & 0.978 & 0.994 & 0.993 & 0.979 & 0.989 & 0.968 \\
    Our PuTT-600-L  & 0.982 & 0.960 & 0.989 & 0.979 & 0.994 & 0.993 & 0.977 & 0.990 & 0.970 \\
    \hline
    NeRF-S*~\cite{nerf} & 0.952 & 0.920 & 0.966 & 0.920 & 0.960 & 0.970 & 0.950 & 0.980 & 0.946 \\
    TensoRF VM-30-S (no shrinking) & 0.974 & 0.949 & 0.985 & 0.962 & 0.992 & 0.991 & 0.970 & 0.982 & 0.962  \\
    Our PuTT-200-S  & 0.976 & 0.954 & 0.985 & 0.971 & 0.990 & 0.989 & 0.968 & 0.989 & 0.965  \\
    \hline
    TensoRF VM-48-M (no shrinking) & 0.975 & 0.949 & 0.986 & 0.963 & 0.992 & 0.991 & 0.973 & 0.985 & 0.964  \\
    Our PuTT-280-M  & 0.977 & 0.956 & 0.983 & 0.974 & 0.990 & 0.990 & 0.975 & 0.984 & 0.967 \\
    \hline

    \multicolumn{9}{@{}l}{\rule{0pt}{3ex}\bf LPIPS$_{VGG}\downarrow$} \\
    \hline
    DVGO-L*~\cite{DVGO} & 0.019 &0.038 &0.010 &0.030& 0.005 &0.004 &0.027 &0.009 &0.027\\
    TensoRF-VM-192-L*~\cite{tensoRF}        & 0.012 & 0.024 & 0.006 & 0.016 & 0.003 & 0.003 & 0.011 & 0.009 & 0.021 \\
    Our PuTT-600-L  & 0.017 & 0.032 & 0.017 & 0.023 & 0.008 & 0.004 & 0.021 & 0.011 & 0.023  \\
    \hline
    NeRF-S*~\cite{nerf} & 0.043 & 0.096 & 0.031 & 0.069 & 0.038 & 0.019 & 0.031 & 0.016 & 0.047 \\
    TensoRF VM-30-S (no shrinking)  & 0.036 & 0.059 & 0.032 & 0.046 & 0.014 & 0.017 & 0.033 & 0.027 & 0.057  \\
    Our PuTT-200-S  & 0.032 & 0.059 & 0.029 & 0.039 & 0.013 & 0.014 & 0.032 & 0.018 & 0.055 \\
    \hline
    TensoRF VM-48-M (no shrinking) & 0.034 & 0.065 & 0.028 & 0.043 & 0.013 & 0.014 & 0.029 & 0.025 & 0.056  \\
    Our PuTT-280-M  & 0.027 & 0.053 & 0.022 & 0.031& 0.011 & 0.011 & 0.024 & 0.017 & 0.046  \\
    \hline
    
    \hline
    \end{tabular}
    \caption{Quantitative results on each scene from the {\bf Synthetic-NSVF}~\cite{NSVF} dataset. Results denoted with an asterisk (*) were sourced from the data presented in the referenced paper. }
    \label{tab:supp_breakdown_nsvf}
\end{table*}

%%%%%%%%%%%%%%%%%%%%%%%%%%%%%%%%%%%%%%%%%%%%%%%%%%%%%%%%%%%%%%%
%%%%%%%%%%%%%%%%%% Tanks and Temples Dataset %%%%%%%%%%%%%%%%%%%%%%%%%%%
%%%%%%%%%%%%%%%%%%%%%%%%%%%%%%%%%%%%%%%%%%%%%%%%%%%%%%%%%%%%%%%

\begin{table*}[htpb]
    \centering
    \renewcommand\tabcolsep{5.0pt}
    \begin{tabular}{l|c|ccccc}
    \hline
    Methods & Avg. & {\it Ignatius} & {\it Truck} & {\it Barn} & {\it Caterpillar} & {\it Family} \\
    \hline\hline
    \multicolumn{7}{@{}l}{\rule{0pt}{3ex}\bf PSNR$\uparrow$} \\
    \hline
    PlenOctrees-L*~\cite{PlenOctrees} & 27.99 & 28.19 & 26.83 & 26.80 & 25.29 & 32.85 \\
    Plenoxels-L*~\cite{plenoxels} & 27.43 & 27.51 & 26.59 & 26.07 & 24.64 & 32.33  \\
    DVGO-L*~\cite{DVGO} & 28.41 & 28.16 & 27.15 & 27.01 & 26.00 & 33.75 \\
    TensoRF VM-192-L~\cite{tensoRF}  & 28.56 & 28.34 & 27.14 & 27.22 & 26.19 & 33.92 \\
    Our PuTT-600-L & 28.37 & 28.08 & 27.09 & 27.22 & 25.93 & 33.56  \\
    \hline
    NeRF-S*~\cite{nerf} & 25.78 & 25.43 & 25.36 & 24.05 & 23.75 & 30.29 \\
    TensoRF VM-30-S (no shrinking) & 27.67 & 27.58 & 26.42 & 26.67 & 25.04 & 32.62  \\
    Our PuTT-200-S & 27.99  &  27.98 & 26.47 & 27.08 &25.47 &32.96 \\
    \hline
    TensoRF VM-48-M (no shrinking)  & 28.03 & 28.04 & 26.70 & 27.06 & 25.28 & 33.06  \\
    Our PuTT-280-M & 28.15 & 28.01 & 26.91 & 27.01 & 25.61 & 33.23 \\
    \hline 
    \hline    

    \multicolumn{7}{@{}l}{\rule{0pt}{3ex}\bf SSIM$\uparrow$} \\
    \hline
    PlenOctrees-L*~\cite{PlenOctrees} & 0.917 & 0.948 & 0.914 & 0.856 & 0.907 & 0.962 \\
    Plenoxels-L*~\cite{plenoxels} & 0.906 & 0.943 & 0.901 & 0.829 & 0.902 & 0.956 \\
    DVGO-L*~\cite{DVGO} &0.911 & 0.944 & 0.906 & 0.838 & 0.906 & 0.962\\
    TensoRF VM-192-L*~\cite{tensoRF}  & 0.920 & 0.948 & 0.914 & 0.864 & 0.912 & 0.965 \\
    Our PuTT-600-L  & 0.917 & 0.943 & 0.908 & 0.865 & 0.907 & 0.960  \\
    \hline
    NeRF-S*~\cite{nerf} & 0.864 & 0.920 & 0.860 & 0.750 & 0.860 & 0.932 \\
    TensoRF VM-30-S (no shrinking) & 0.899 & 0.924 & 0.900 & 0.840 & 0.880 & 0.951  \\
    Our PuTT-200-S  & 0.901 & 0.931 & 0.901 & 0.824 &0.901 & 0.950  \\
    \hline
    TensoRF VM-48-M (no shrinking) & 0.901 & 0.928 & 0.900 & 0.843 & 0.882 & 0.951  \\
    Our PuTT-280-M  & 0.907 & 0.934 &0.909 & 0.835 & 0.903 & 0.953  \\
    \hline
    \hline    

    \multicolumn{7}{@{}l}{\rule{0pt}{3ex}\bf LPIP$_{VGG} \downarrow$} \\
    \hline
    DVGO-L*~\cite{DVGO} & 0.148 & 0.090 & 0.145 & 0.290 & 0.152 & 0.064 \\
    TensoRF VM-192-L*~\cite{tensoRF}   & 0.140 & 0.078 & 0.145 & 0.252 & 0.159 & 0.064 \\
    Our PuTT-600-L & 0.142 & 0.082 & 0.153 & 0.241 & 0.163 & 0.070  \\
    \hline
    NeRF-S*~\cite{nerf} & 0.198 & 0.111 & 0.192 & 0.395 & 0.196 & 0.098 \\
    TensoRF VM-30-S (no shrinking)  & 0.185 & 0.099 & 0.218 & 0.297 & 0.227 & 0.083  \\
    Our PuTT-200-S & 0.197 & 0.094 & 0.326 & 0.272 & 0.220 & 0.072  \\
    \hline
    TensoRF VM-48-M (no shrinking)  & 0.156 & 0.092 & 0.173 & 0.257 & 0.187 & 0.074 \\
    Our PuTT-280-M &0.150 & 0.086 & 0.162 & 0.259 &0.170 & 0.073  \\
    \hline
    \end{tabular}
    \caption{Quantitative results on each scene from the {\bf Tanks\&Temples} \cite{tanks_and_temples} dataset. Results denoted with an asterisk (*) were sourced from the data presented in the referenced paper. }
    \label{tab:supp_breakdown_tanksandtemples}
\end{table*}

\section{Compatibility of TensoRF's "Shrinkage" with QTT}
\label{app:no_shrinkage}
\textbf{Shrinkage:} TensoRF’s “shrinkage” process removes parameters corresponding to empty regions in the 3D space. For instance, in TensoRF’s VM (Vector-Matrix) decomposition, parameters for indices $x, y, z < 10$ are pruned if the region is empty, reallocating them to denser regions to enhance resolution.

\textbf{Incompatibility with QTT:} Unlike TensoRF, QTT uses a hierarchical structure with mode quantization, compressing dimensions into powers of two. This structure mixes parameters, interlinking those representing a coordinate $(x,y,z)$ with many others. Consequently, QTT’s parameter sharing across the hierarchy complicates direct pruning as in TensoRF’s “shrinkage.” Figure \ref{fig:coarse_to_fine_struc} illustrates this interlinked structure.
While QTT’s interdependency complicates “shrinkage,” it allows capturing dimensional correlations, ensures logarithmic parameter growth, and improves robustness against noise and incomplete data. Updating parameters for a specific coordinate refines those across interconnected indices, leveraging correlations for resilient data representation. This complexity necessitates a coarse-to-fine training approach, as proposed in our PuTT method, fostering the hierarchical structure modeled by QTT.

\section{Exploring Inpainting Capabilities with PuTT}
\label{app:impainting}

Inpainting tasks benefit from a rich prior of the true data distribution, often using generative models. Our method, however, focuses on a single example to learn a compact, high-quality representation suitable for tasks like denoising or predicting missing values.

\begin{figure}[h]
\centering
\includegraphics[width=0.8\textwidth]{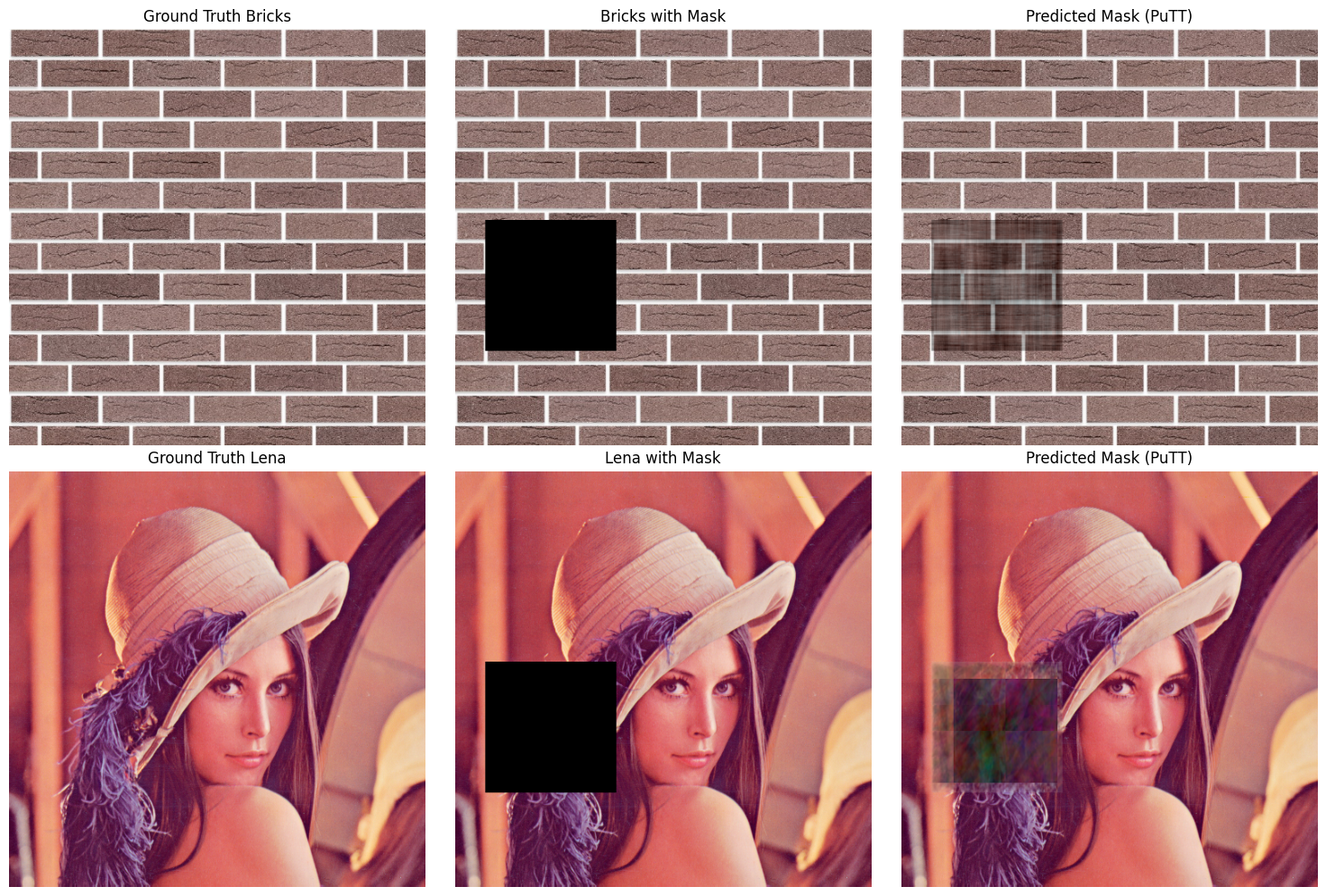}
\caption{Inpainting results on two $512 \times 512 \times 3$ images: a brick pattern (top) and the Lena image (bottom). Columns display the original image, the masked image (10\% mask), and the PuTT reconstruction. Reconstructions were achieved after 1024 iterations using only unmasked regions.}
\label{fig:impainting_exp}
\end{figure}

We conducted an experiment with two images: a brick pattern and the Lena image. Each image had a continuous mask covering 10\% of the area. Using PuTT, we attempted to reconstruct the masked regions. Results are shown in Fig.~\ref{fig:impainting_exp} with columns for the original image, the masked image, and the reconstructed image. The brick image shows PuTT’s effectiveness in structured settings, while the Lena image highlights the challenges of inpainting large areas without sophisticated priors.

\section{Exploring Rank Incrementation with SlimmeRF}
\label{app:slimmerf}
SlimmeRF~\cite{yuan2023slimmerf} enhances the TensoRF VM framework by introducing an adaptive rank mechanism, dynamically adjusting the model’s learning capacity. The model starts with a low-rank representation and incrementally increases the rank based on learning progress, capturing essential features early and building complexity as needed. Our approach similarly increases tensor rank progressively, a strategy used in physics methods like DMRG~\cite{DMRG} and TT-cross~\cite{TT_cross}, aligning with our progressive learning in PuTT.

We explored this with a 2D compression experiment on 4K images, as shown in Table \ref{tab:rank_increment}. We compared a standard QTT model without upsampling to one with progressive rank incrementation ([2, 4, 8, 16] ranks during initial training). These experiments indicate that gradual rank increments can improve PSNR scores without adding complexity.

These results, excluding dimensional upsampling, highlight the potential of rank incrementation. However, more extensive experiments are needed to fully validate these findings.

\begin{table}[htbp]
    \centering
    
    \label{tab:results}
    \begin{tabular}{|c|c|c|}
        \hline
        \textbf{Rank Increment} & \textbf{PSNR (Avg)} & \textbf{Std} \\ \hline
        0 (Baseline)            & 37.444               & 0.821        \\ \hline
        2                       & 38.168               & 0.391        \\ \hline
        4                       & 38.273               & 0.444        \\ \hline
        8                       & 38.331               & 0.464        \\ \hline
        16                      & 38.131               & 0.513        \\ \hline
    \end{tabular}
    \caption{Results of Progressive Rank Incrementation. This table summarizes the effects of rank incrementation on learning efficiency for a 1k resolution image of a 'Girl with a Pearl Earring'. It contrasts a baseline QTT model without upsampling, starting with a rank of $200$, with QTT models employing progressive rank incrementation from a base rank of $100$ to $200$. Rank increments of $2$, $4$, $8$, and $16$ were tested over four intervals: $[100, 1000], [100, 1500], [250, 1500]$, and $[250, 750]$, to determine the start and end points for rank increments.
    The table displays the average PSNR values, alongside standard deviations, across three iterations for each rank increment strategy. These results highlight the improved performance due to strategic rank incrementation, with all strategies surpassing the baseline. For instance, an increment step of $2$ within the $[100, 1000]$ interval increased the rank every $(1000 - 100) / (200 - 100) = 9$ iterations. Although a rank increment of 8 yielded the best average PSNR, differences between increment steps were minimal, suggesting a thoughtful approach to rank incrementation can significantly enhance the model's learning capability.}
    \label{tab:rank_increment}
\end{table}

\newpage

\end{document}